# Hole-robust Wireframe Detection

Naejin Kong, Kiwoong Park, Harshith Goka
Samsung Research AI Center
{naejin.kong, kyoong.park, h9399.goka}@samsung.com

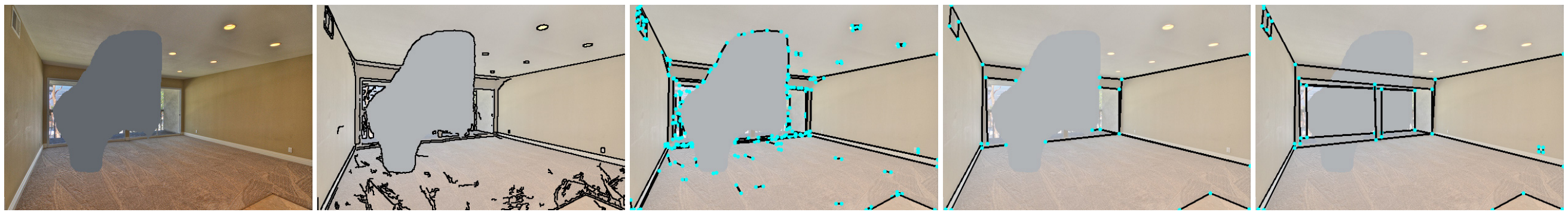

(a) Image with hole (b) Canny Edges [1] (σ=1) (c) Linelet [2] (d) HT-HAWP [16] (e) Ours

Figure 1: **Conventional edges or line segments ((b) and (c)) compared with *"Wireframes"* ((d) and (e)).** (a) Input image occluded by a large hole in dark gray color. Conventional edge and line segment detection such as (b) and (c) does not tell which of those detected are meant for salient structure. Wireframes are far better aligned with salient structural information of the scene as in (d) and (e). However, existing wireframe detection such as [16] in (d) still does not know how to handle hole occlusion, while our approach in (e) handles it very robustly. (black contours: edges/lines, light blue dots: junctions)

## Abstract

*"Wireframe" is a line segment based representation designed to well capture large-scale visual properties of regular, structural shaped man-made scenes surrounding us. Unlike the wireframes, conventional edges or line segments focus on all visible edges and lines without particularly distinguishing which of them are more salient to man-made structural information. Existing wireframe detection models rely on supervising the annotated data but do not explicitly pay attention to understand how to compose the structural shapes of the scene. In addition, we often face that many foreground objects occluding the background scene interfere with proper inference of the full scene structure behind them. To resolve these problems, we first time in the field, propose new conditional data generation and training that help the model understand how to ignore occlusion indicated by holes, such as foreground object regions masked out on the image. In addition, we first time combine GAN in the model to let the model better predict underlying scene structure even beyond large holes. We also introduce pseudo labeling to further enlarge the model capacity to overcome small-scale labeled data. We show qualitatively and quantitatively that our approach significantly outperforms previous works unable to handle holes, as well as improves ordinary detection without holes given.*

## 1. Introduction

More and more people these days live in cities, hang around urban places, or spend most time indoors due to the pandemic. These environments surrounding us are cluttered with man-made architectures and artificial objects that appear to be in regular, structural shapes instead of random, repeated patterns in natural scenery, due to reasons to achieve structural integrity, ease of mass production, physical balance and so on. *"Wireframes"* [11] can represent this visual property effectively and compactly, as a set of line segments and their crossing junctions aligned well with straight and boxy corners salient to the man-made shapes.

Conventional edges or line segments may not be suitable for this task, as they focus on all visible edges and lines without particularly distinguishing which of them are more salient to man-made structural information. For instance, Canny edge detection [1] may produce spurious edges that include all kinds from texture change, object boundaries, or illumination difference (e.g., Figure 1 (b)). Adjusting the global threshold to avoid too many edges may end up with too sparse and discontinuous detection either. Moreover, edge detection is easily distracted by strong noise and illumination in the pixel signals that disturb recognizing the true underlying structure. Conventional line segment detection such as [2] in Figure 1 (c) would still show similar issues except being better aligned with straight structures.

In contrast, wireframes focus on the salient structural shapes that dominantly appear in the scene. This property makes them a far better representation to man-made scene structure, as shown in Figure 1 (d) and (e). However, existing wireframe detection models such as [16] in (d) do not explicitly pay attention to understand the underlying structural composition of the scene, but rely solely on the supervision of labeled wireframes in the dataset. As a result, these methods are unable to predict hidden structure beyond hole occlusion due to insufficient knowledge. On the contrary, our approach is designed to more directly understand the scene composition by learning to regenerate unknown scene contents inside large hidden areas given by holes, thus better fulfills the purpose of wireframes and produces highly hole-robust detection as in Figure 1 (e).

This hole-robustness is important, as we often face that a desired background scene is occluded by many foreground objects, such as furniture and electric gadgets in indoor scenes or passerby and cars in street scenes. Hence, a more practical wireframe detection algorithm should be able to work around the occlusion and infer the correct structure underneath. Note that these occlusions can be indicated by holes masked out on the image, by using existing foreground object segmentation models (ex., Detectron2 [28]) or manual user annotation.

Our hole-robust detection solution is novel in three folds:

1. Proposing new **conditional data generation and training** for hole-robust detection
2. Combining **GAN** [9] with the supervised model to improve its capacity to infer underlying structure
3. Introducing semi-supervised learning with **pseudo labeling** to further enlarge model capacity

Moreover, our approach is flexible and general, applicable to any recent models such as HAWP [31] (HT-HAWP [16]) and L-CNN [35] (HT-LCNN [16]), as well as very recent F-Clip [3] (preprint still under review).

To the best of our knowledge, we are first to raise an issue on this hole-robustness problem for wireframes along with a novel solution. Our approach is not only bound to hole-robust detection but also enhances the ordinary detection performance when holes are not given, thanks to an enhanced scene understanding capability and enlarged model capacity achieved from our novel hole-robust approach.

## 2. Related Works

**Traditional line segment detection.** Line segment detection is one of the classic and core tasks in computer vision. Traditional line segment detectors [27, 2, 19] rely on direct pixel signal features, such as image gradients. Another popular and classic approach is the Hough-transform line detection [6], which globally detects straight lines with an angle-distance based representation of the line segments. However, classical line segment detection methods have a limitation such that there is no way to tell how the lines are connected to each other by junctions, thus produce many uncorrelated scattered lines as shown in Figure 1 (c).

**Deep learning based approaches.** Since the main concept on wireframes [11] emerged, various methods have been proposed to detect the wireframes based on deep learning in three types of mainstream approaches so far.

One mainstream approach first identifies junctions and then finds connectivity in between two sampled junctions as the endpoints of each line segment. [11] proposed a framework that maintains two independent network branches to detect a junction heatmap and a line heatmap in parallel, which are then merged to compose line segments. Inspired by this work, [35] proposed the first end-to-end wireframe detection solution in a two-stage design, called L-CNN. This framework is motivated by typical object detection methods [8, 7, 23, 10], where the first stage predicts both the pixel-wise junction and line heatmaps, and then the second stage samples initial line segment candidates by using the detected junctions and deep feature maps. Then, these line segment proposals are verified through sophisticated classification to leave only true positive line segments. PPGNet [33] predicts a junction heatmap only, and then infers line segment candidates by using the predicted junctions as well as an adjacency matrix formulated from those junctions.

Another mainstream approach pursues direct line segment detection from the deep network without going through intermediate line segment sampling, based on a region-partition based representation of the line segments. This representation interprets a line segment vector as an angle-distance combination with respect to a polar coordinate system. [30] introduced a 6-D Attraction Field Map where every pixel in the AFM map corresponds to only one line segment. [31] refined this Attraction Field Map representation, and proposed a framework called HAWP, that uses a 4-D holistic Attraction Field Map composed of four channels; a distance channel with respect to distance from a pixel to its projected point onto a line segment, an angular orientation channel with respect to global rotation of the line segment, and two more angular channels with respect to the angles of two endpoints. Recently, [16] discovered that a Hough-transform [6] based global line prior combined with the backbone network of L-CNN [35] and that of HAWP [31] further improves their performances.

Inspired by object detection that has evolved from a two-stage scheme [8, 7, 23, 10] to a one-stage scheme [22, 18, 15, 14], there have been very recent attempts to follow a similar one-stage scheme for the wireframe detection. F-Clip [3] (preprint still under review) introduced a center point based representation of a line segment vector, which achieved a high speed while maintaining the ac-curacy on par with HAWP [31] by using a stacked HourGlass [21] backbone commonly shared in recent works. They also showed that slower but higher accuracy can be achieved by replacing the backbone with HRNet [25] that is far more complex and fundamentally different from the stacked HourGlass. TP-LSD [29] also detects line segments in a single stage by introducing a tri-point representation composed of a center point as well as two endpoints, but it underperforms HAWP [31]. LETR [29] introduced Transformers [26] to the wireframe detection task for the first time. It is very slow due to the overhead of the Transformers, and does not show a clear advantage against HAWP [31] on the standard Wireframe test set [11].

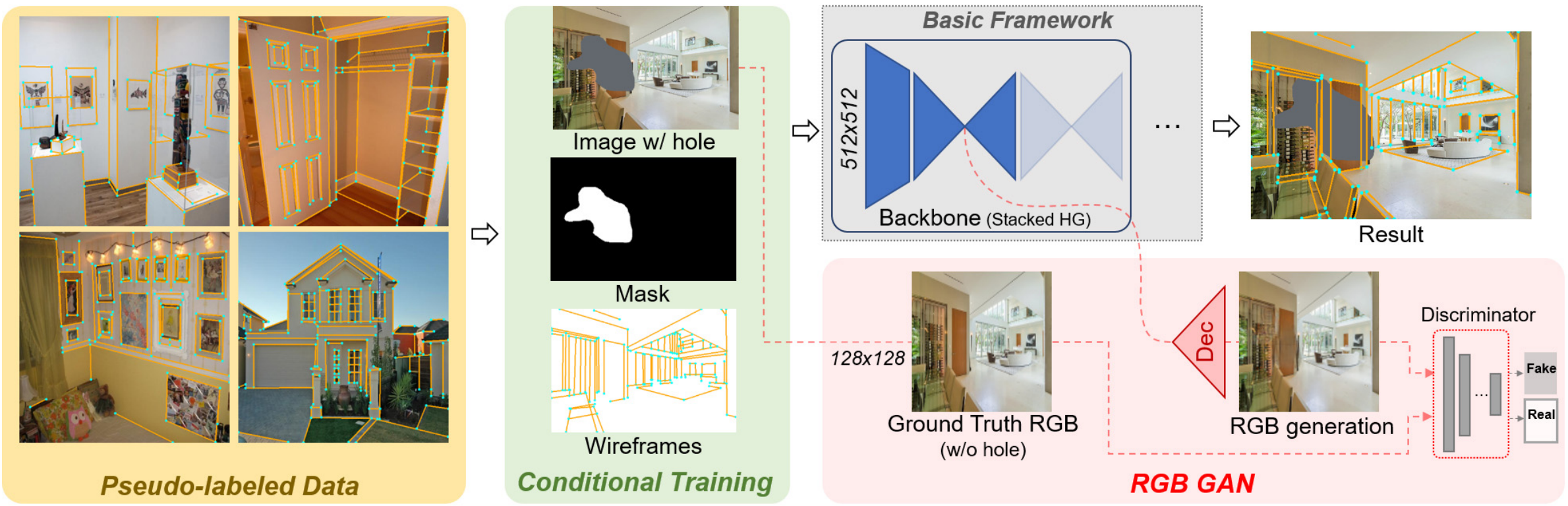

Figure 2: Overview of our approach (see **SuppMat:Sect.2** for details on the RGB GAN architecture).

We select L-CNN [35] (and HT-LCNN [16]), HAWP [31] (and HT-HAWP [16]), and F-Clip [3] as the representative frameworks from the three mainstream approaches, and prove that our approach applied to these frameworks improves hole-robustness as well as ordinary detection.

## 3. Approach

In this section, we choose HT-HAWP [16] as our representative basic framework, because it is the best performing one on conventional wireframe detection among recent models sharing the stacked HourGlass [21] backbone. We will show improvements with our approach applied to other basic frameworks in Section 4. Figure 2 overviews our approach, where the basic framework (gray shaded) can be flexibly chosen from other recent models. The yellow, green and red shaded modules compose our full approach.

### 3.1. Basic Framework: HT-HAWP

We call an HAWP framework [31] with Hough-transform priors [16] as "HT-HAWP", which is adopted as our representative basic framework. The HAWP framework [31] is composed of three steps as follows: 1) stacked HourGlass [21] backbone that predicts a 4-D Attraction Field Map that represents line segment vectors in terms of a 4-channel dense map as well as a junction heatmap that encodes the positions and offsets of the junctions in [-1,1] normalized coordinates, 2) matching raw predicted line segments and junctions, and 3) verifying the matched line segments through classification that outputs a final set of verified line segments coupled with their scores, where a segment with a higher score is more probable to be true positive. In addition, as shown in [16], adding inside the backbone a learnable Hough-transform [6] based mini-network that extracts explicit line-like features from interim network features can further improve the line detection. It can be repeatedly used in between residual modules, the basic building blocks of HourGlass (see **SuppMat:Sect.2** for details).

### 3.2. Conditional Training

We first redesigned the original training for this model to new conditional training with mask holes, as depicted in a green shaded area in Figure 2. For this, we reformed the original Wireframe training dataset [11] of 5,000 ⟨image, wireframe⟩ pairs into new ⟨image with hole, wireframe, mask⟩ pairs by superimposing 0.1-10% size real foreground object silhouettes onto each target image (see Figure 3), where these silhouettes were detected from a Places365 test set [34] using a public segmentation library [28]. Larger then 10% size did not show a clear benefit. We express the pixel value inside the hole using a mean pixel value over all RGB images in the dataset, which performed better than using a peak value such as zero or one, as each input image is normalized by subtracting this mean value during training.

We train on this new conditional dataset by progressively increasing the hole size along with epoch increment, which performed better than choosing a hole in random size. Figure 3 visualizes this. In addition, we place the hole by avoiding isolated components in the scene, such as a window, picture frame, home appliance like TV, or any tiny object that may form a loop of lines. As the hole goes larger, it may become more probable that an isolated line loop is fully contained within the hole, and then known regions

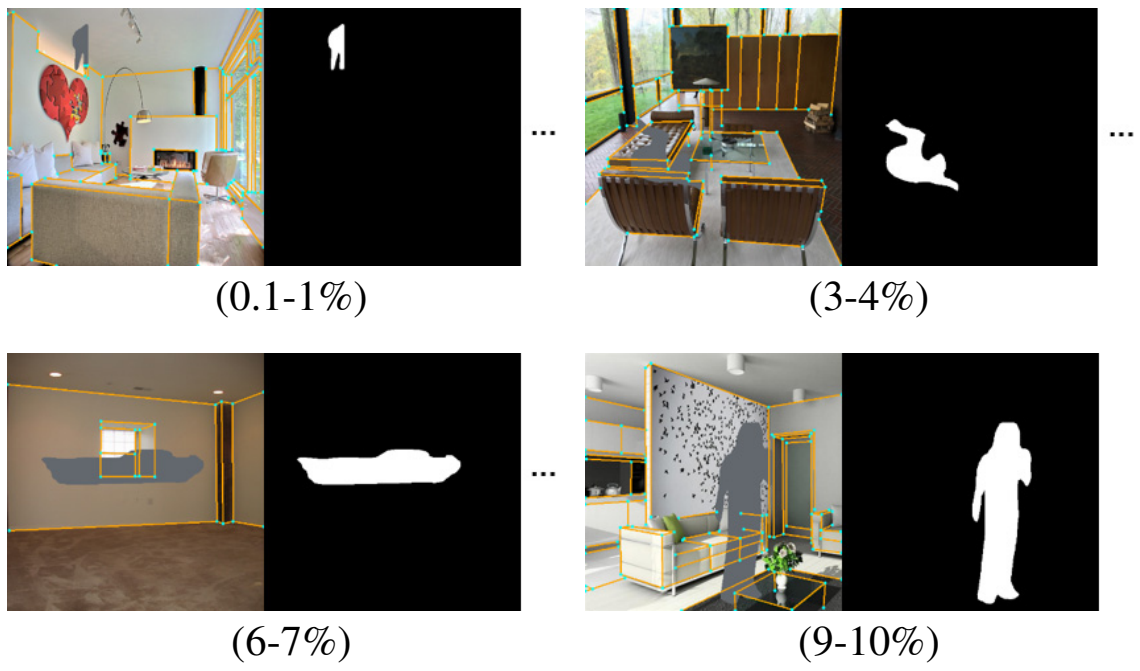

Figure 3: Conditional data with progressive hole size increment.

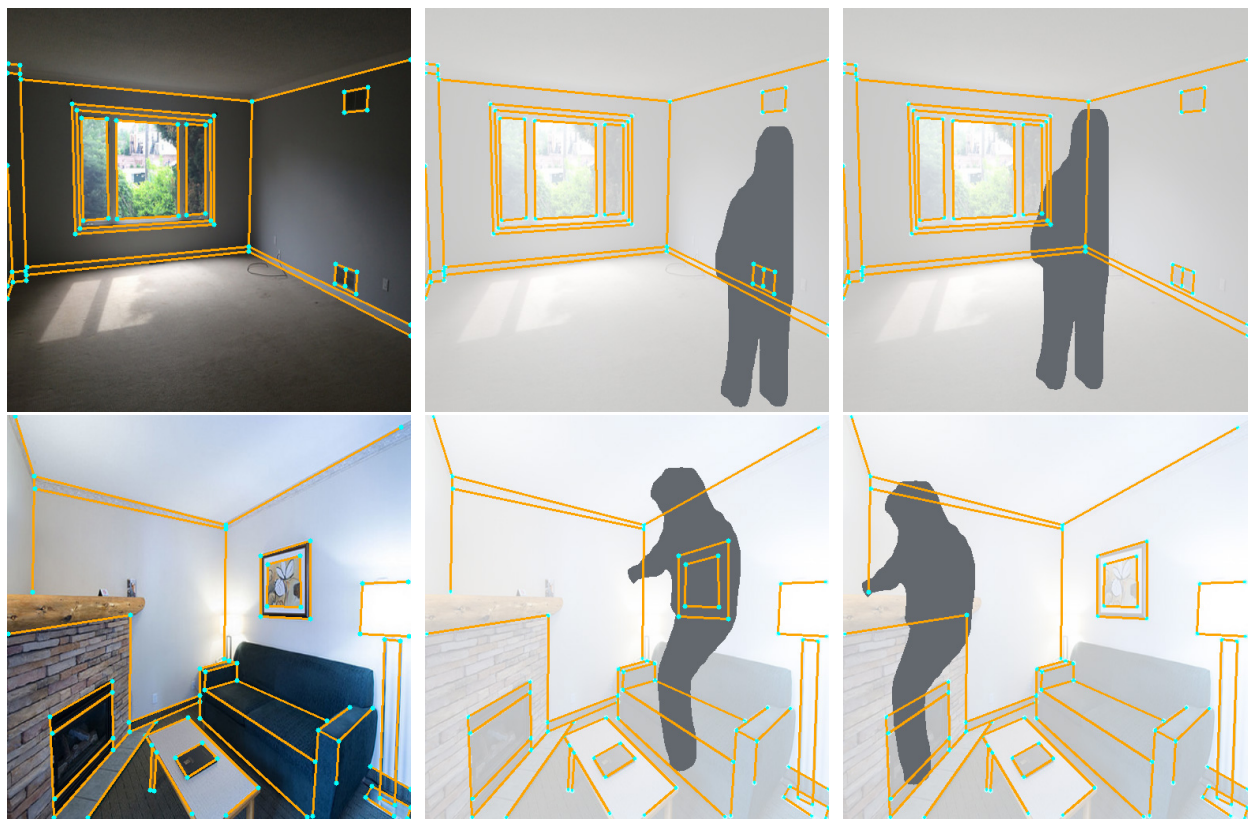

(a) No hole (b) Random (c) Avoid-isolation

Figure 4: **Hole placement schemes**. Random placement as (b) may entirely hide scene components unlike (c) that avoids such isolation. (Hole size: 9.91% above, 9.96% below)

may not provide any clue to infer the entire unknown structure. Figure 4 shows examples without and with the scheme applied, which makes visually convincing why we need to avoid isolation. **SuppMat:Sect.3** describes further details on our conditional data generation and training methods. Ablation studies in **SuppMat:Sect.6.1** justify our choices.

### 3.3. Combining GAN

The conditional training provides a huge improvement on hole-robustness, but as the hole goes larger, it may become harder to detect line segments without predicting the scene composition inside, at least roughly. To further support this case, we transform the original supervised backbone network [21] into an RGB GAN generator by adding a small 128 × 128 decoding branch to the first bottleneck along with a PatchGAN discriminator [12] as shown in the red shaded module of Figure 2. Training the model with our RGB GAN encourages that the backbone's encoder in connection with the RGB decoder learns to better understand the overall scene composition principles by referring to the background contents outside the hole, which helps regenerate invisible scene contents inside the hole as realistically as possible. We also tested GANs directly applied to a junction heatmap, line heatmap, or attraction field map, but found that it was not as effective as the RGB GAN (see **SuppMat:Sect.6.2** for more details).

We derived losses to train the RGB GAN by referring to a basic formulation in [20] for GAN-based *image inpainting*. As to be described below, we adopted some of their unique choices, such as to use an L1 norm instead of a squared L2 or Frobenius norm in the original form of a perceptual or style loss, or to use five layers in the VGG-19 network instead of VGG-16 for the perceptual or style loss. However, note that our RGB GAN does not aim at perfect image inpainting but is purposed to encourage the backbone to better understand overall scene structure composition principles.

Let $G$ is the generator network (i.e., early part of the backbone + RGB decoder) and $D$ is the discriminator network. Then, we define an adversarial loss [9] for an objective $\min_G \max_D \mathcal{L}_{adv}$ as:

$$\mathcal{L}_{adv} = \mathbb{E}_{\mathbf{x}} \left[\log D(\mathbf{x})\right] + \gamma \cdot \mathbb{E}_{\tilde{\mathbf{x}}} \log\left[1 - D(\tilde{\mathbf{x}})\right] \quad (1)$$

where a generated RGB image is $\tilde{\mathbf{x}} = G(\mathbf{x} \odot (1 - \mathbf{m}), \mathbf{m})$, $\mathbf{x}$ is a ground truth image without hole that is resized into $128 \times 128$, $\mathbf{m}$ is a binary mask map (1 inside the hole and 0 outside), $\odot$ is an element-wise multiplication, and $\gamma = 0.1$. We use a non-saturating version of Eq. (1) for the generator loss to avoid vanishing gradients, which minimizes $-\mathbb{E}_{\tilde{\mathbf{x}}} \log\left[D(\tilde{\mathbf{x}})\right]$ instead of $\mathbb{E}_{\tilde{\mathbf{x}}} \log\left[1 - D(\tilde{\mathbf{x}})\right]$.

We add a perceptual loss [13] to encourage perceptually more realistic looking generation:

$$\mathcal{L}_{per} = \sum_{i=1}^{L} \frac{1}{C_i H_i W_i} \left\| \Phi_i(\mathbf{x}) - \Phi_i(\tilde{\mathbf{x}}) \right\|_1 \quad (2)$$

where $\Phi_i$ is a $C_i \times H_i \times W_i$ feature map of the $i$-th layer from a VGG-19 [24] network, pretrained on the ImageNet dataset [4]. In our implementation, we used *relu1_1*, *relu2_1*, *relu3_1*, *relu4_1* and *relu5_1* layers (i.e., $L = 5$).

We also add a style loss [17] on the same pretrained VGG-19 layers adopted in Eq. (2).

$$\mathcal{L}_{sty} = \sum_{i=1}^{L} \left\| G_j^{\Phi}(\mathbf{x}) - G_j^{\Phi}(\tilde{\mathbf{x}}) \right\|_1 \quad (3)$$

where $G_j^{\Phi}(\mathbf{x})$ is a $C_j \times C_j$ *Gram* matrix from a $C_j \times H_j \times W_j$ feature map $\Phi_i$, where each element at $(c, c')$ in the matrix is $\frac{1}{C_j H_j W_j} \sum_{h=1}^{H_j} \sum_{w=1}^{W_j} \Phi_j(\mathbf{x})_{(h,w,c)} \Phi_j(\mathbf{x})_{(h,w,c')}$.

We then provide a reconstruction loss over the known pixels (i.e., outside the mask hole) in a masked L1 norm:

$$\mathcal{L}_{rec} = \frac{1}{N} \left\| (\mathbf{x} - \tilde{\mathbf{x}}) \odot (1 - \mathbf{m}) \right\|_1 \quad (4)$$

where $N$ is the number of non-zero pixels in $(1 - \mathbf{m})$ to get an average. In our experiments, a feature matching loss [32] recently known to be useful to regularize GAN training did not help much but rather degraded the detection quality, so we decided to drop it. Finally, our total GAN loss $\mathcal{L}$ is:

$$\mathcal{L} = \lambda_{adv} \cdot \mathcal{L}_{adv} + \lambda_{per} \cdot \mathcal{L}_{per} + \lambda_{sty} \cdot \mathcal{L}_{sty} + \lambda_{rec} \cdot \mathcal{L}_{rec} \quad (5)$$

where we use $\lambda_{adv} = 1$, $\lambda_{per} = 0.1$, $\lambda_{sty} = 250$, $\lambda_{rec} = 1$.

In **SuppMat:Sect.2.4** we visualize generated RGB images from our trained RGB GAN. It demonstrates that the backbone encoder learned to infer overall scene structure inside the hole, which will be then, propagated over the rest of the model to induce more robust structural estimation.

### 3.4. Further Enhancement with Pseudo Labeling

Existing models rely entirely on the supervision of the Wireframe training dataset [11] composed of 5,000 examples. The scale of this training data would be too small even with strong data augmentation. To alleviate this issue, we further introduce a semi-supervised learning approach through pseudo labeling, which increases the training scale without need to add more manual annotations. Indeed, this additional approach significantly leveled up the performance, as demonstrated by experiments in Section 4.

We created a new pseudo-labeled dataset by using the Places365 Challenge dataset [34], where the pseudo label means estimated wireframe line segments and junctions. First, we only choose about 3.3M images in 157 structural scene categories that we had manually defined (e.g., campus, terminal, hall, store, and so on), out of all 8M images. Next, we retrieve estimated wireframes on these images by applying the pretrained HT-HAWP [16] model. Finally, we filter out supposedly less accurate or less densely annotated examples based on the three criteria:

- Number of lines in an image $> 74.98$
- Total length of all lines in an image $> 6456.57$
- The ratio $\frac{\text{\# of Junctions}}{\text{\# of Lines}}$ in an image $< 1.34$

The threshold values in these criteria were determined by inspecting the distributions of ground truth wireframe labels in each image of the Wireframe training dataset [11], which follow quite ideal normal distributions. Assuming an ideal distribution of the pseudo labels would be similar to that of the ground truth labels, we applied the same thresholds from the ground truth dataset to filter out more possibly well detected and more dense labels from the candidate pseudo-labeled images. **SuppMat:Sect.4** depicts three criteria histograms and justifies an effectiveness of this thresholding.

Once the thresholds are applied, 142k images remain in the final pseudo labeled dataset. This is about 28 times larger than the 5k images in the original Wireframe training set. The yellow shaded area in Figure 2 shows visualization of the pseudo-labeled examples sampled from the final pseudo dataset, demonstrating the great annotation quality. Note that these pseudo-labeled images are still without hole, that is, they have not yet transformed into conditional data.

We transform the Wireframe training set with hole by applying conditional generation in Section 3.2 with avoid-isolation. We also transform the pseudo-labeled data in the similar manner, but with random placement. We avoid additionally generating a huge mask pool for all 142k images but take random masks from the pre-generated mask pool for the Wireframe training set (10 candidates $\times$ 10 hole intervals $\times$ 5,000 images), which was still enough to enhance the performance significantly. We first train on the large-scale pseudo data with hole in order to enlarge the model capacity, and then finally fine-tune on the small-scale ground truth data with hole to further enhance fine-level accuracy.

## 4. Experiments

We provide implementation details in **SuppMat:Sect.5**. In order to test the hole-robust performance, we created new test sets with hole derived from standard test sets: the Wireframe test set [11] (462 examples) and the York Urban dataset [5] (102 examples all used for testing. Created in 2008 thus, its labeling may not be fully consistent with a more recently emerged wireframe concept). We superimposed an object silhouette hole in each image of the test set by following the similar manner described in Section 3.2. Specifically, we placed a hole by avoiding isolated components within the scene in case of 0-10% hole size (starting from 0.1% in practice), and placed it at an entirely random location in case of 10-30% hole size since it is very hard to avoid isolation with such a large hole (see Figure 4 to compare two cases). We also test the ordinary detection performance using the original standard test sets [11, 5] without hole, therefore not taking account of any hole occlusion.

Note that the runtime speed of a model applying our approach is similar to its basic model, as the RGB decoder branch is eliminated without affecting the detection result at inference time. Ours using HT-LCNN or HT-HAWP [16] may be slightly slower than the basic model, as we replace the first residual module with a whole HourGlass unit.

We show the effect our approach on various basic frameworks in Section 4.1. Due to space, we show further analysis and studies in **SuppMat:Sect.6** as follows: **⟦6.1⟧** Ablation on our conditional training made in various settings to show the effectiveness of progressive hole size increment against random size hole selection and the avoid-isolation scheme against random location placement. **⟦6.2⟧** Applying GAN [9] to various types instead of RGB, such as a junction, line, or attraction field map, to show how we had concluded RGB GAN is the best one. **⟦6.3⟧** Testing the sole effect of pseudo labeling without mixing holes on the ordinary wireframe detection performance to show it is still effective but limited to improve hole-robustness significantly with-

| | Wireframe Test Set [11] | | | | | | | | | | | | York Urban Dataset [5] | | | | | | | | | | | |
|---|---|---|---|---|---|---|---|---|---|---|---|---|---|---|---|---|---|---|---|---|---|---|---|---|
| | 10-30% Hole | | | | 0-10% Hole | | | | without Hole | | | | 10-30% Hole | | | | 0-10% Hole | | | | without Hole | | | |
| | $sAP^5$ | $sAP^{10}$ | $mAP^J$ | $AP^H$ | $sAP^5$ | $sAP^{10}$ | $mAP^J$ | $AP^H$ | $sAP^5$ | $sAP^{10}$ | $mAP^J$ | $AP^H$ | $sAP^5$ | $sAP^{10}$ | $mAP^J$ | $AP^H$ | $sAP^5$ | $sAP^{10}$ | $mAP^J$ | $AP^H$ | $sAP^5$ | $sAP^{10}$ | $mAP^J$ | $AP^H$ |
| L-CNN [35] | 32.83 | 35.78 | 40.4 | 61.3 | 47.73 | 51.68 | 53.2 | 75.3 | 58.91 | 62.86 | 59.4 | 80.3 | 14.65 | 16.16 | 21.9 | 44.6 | 20.23 | 22.15 | 27.5 | 54.6 | 24.33 | 26.39 | 30.4 | 58.0 |
| HT-LCNN [16] | 33.56 | 36.48 | 41.4 | 57.1 | 48.89 | 52.79 | 54.2 | 76.6 | 60.33 | 64.22 | 60.6 | 81.6 | 15.84 | 17.48 | 23.5 | 40.4 | 21.47 | 23.62 | 29.3 | 50.1 | 25.72 | 28.02 | 32.5 | 53.0 |
| F-Clip(HG2) [3] | 33.51 | 36.68 | / | 65.3 | 49.46 | 53.87 | / | 79.5 | 61.25 | 65.77 | / | 84.3 | 15.86 | 17.54 | / | 48.6 | 22.02 | 24.07 | / | 58.6 | 27.10 | 29.26 | / | 62.1 |
| F-Clip(HR) [3] | 34.56 | 37.53 | / | 66.3 | 51.44 | 55.60 | / | 80.6 | 64.30 | 68.34 | / | 85.7 | 16.53 | 18.25 | / | 50.4 | 23.53 | 25.76 | / | **61.3** | **28.49** | **30.80** | / | **65.0** |
| LETR [29] | 32.84 | 37.23 | / | 69.5 | 50.24 | 56.22 | / | 82.5 | 59.19 | 65.64 | / | 86.1 | 15.55 | 18.26 | / | 49.1 | 21.40 | 25.33 | / | 59.1 | 25.65 | 29.61 | / | 62.1 |
| HAWP [31] | 34.80 | 37.74 | 40.8 | 64.5 | 50.80 | 54.84 | 54.0 | 79.4 | 62.52 | 66.49 | 60.2 | 85.0 | 15.79 | 17.48 | 22.5 | 46.3 | 21.61 | 23.79 | 28.6 | 57.3 | 26.15 | 28.54 | 31.6 | 61.3 |
| HT-HAWP [16] | 35.64 | 38.54 | 41.8 | 65.5 | 51.58 | 55.50 | 55.1 | 80.1 | 63.26 | 67.12 | 61.3 | 85.7 | 15.57 | 17.18 | 23.2 | 46.8 | 21.45 | 23.63 | 29.0 | 56.6 | 25.31 | 27.65 | 31.9 | 60.7 |
| HAWP + Our full approach | 47.59 | 51.50 | 49.6 | 74.9 | 61.83 | 65.98 | 61.0 | 85.0 | 65.98 | 69.76 | 63.0 | 86.9 | **20.81** | **23.01** | **27.0** | **53.4** | **25.62** | **28.01** | **31.7** | 60.2 | 27.23 | 29.58 | **32.8** | 62.6 |
| HT-HAWP + Our full approach | **48.02** | **51.82** | **49.9** | **76.6** | **62.21** | **66.31** | **61.3** | **85.5** | **66.19** | **69.92** | **63.2** | **87.0** | 19.96 | 21.92 | 25.4 | 52.9 | 24.72 | 26.94 | 29.8 | 57.6 | 26.06 | 28.26 | 30.9 | 59.2 |

Table 1: **Effect of our full approach, evaluated on the Wireframe test set [11] and York Urban dataset [5], *with* and *without* hole.** $sAP^T$ (T: distance threshold) and $mAP^J$ are positional accuracy for line segments and junction locations, respectively, and $AP^H$ is a line heatmap based metric, all widely used in the field. F-Clip [3] does not predict junctions thus $mAP^J$ is unavailable. Note that in gray marked rows, F-Clip(HR) uses an HRNet [25] backbone, and LETR largely relies on Transformer [26] encoder-decoder, which make them both fundamentally very different from the other frameworks sharing the stacked HourGlass [21] backbone.

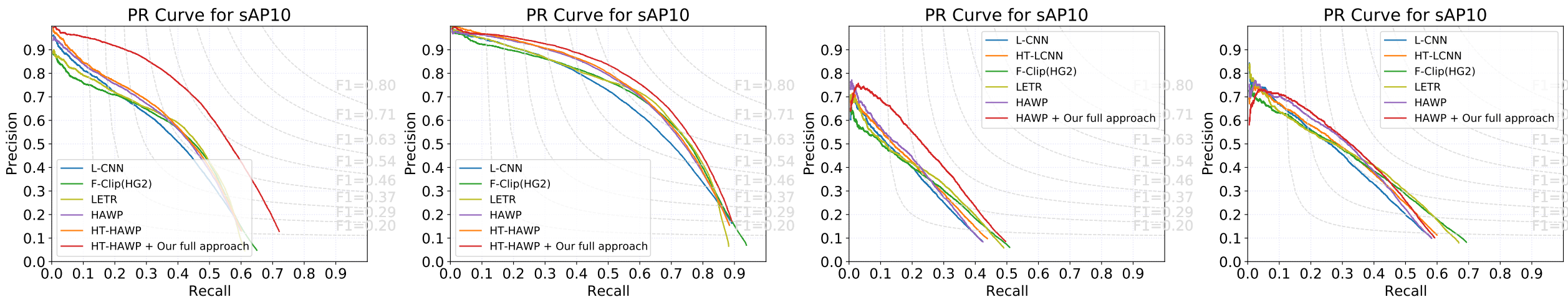

(a) Wireframe [11], **10-30% hole** (b) Wireframe [11], ***without* hole** (c) York Urban [5], **10-30% hole** (d) York Urban [5], ***without* hole** out conditional training and RGB GAN further combined. **SuppMat:Sect.9-11** provide extra studies on the effect of initialization, light conditions, and inpainted input.

Figure 5: PR curves for $sAP^{10}$ on the tests *with* and *without* hole in Table 1. Those on 0-10% hole and more PR curves for $AP^H$ can be found in **SuppMat:Sect.8**.

## 4.1. Effect of Our Approach

Table 1 compares our full models applying all of the conditional training, RGB GAN and pseudo labeling, with recent models that are incapable of handling hole occlusion, by employing widely used accuracy metrics $sAP^T$ (T is a distance threshold): structural average precision with respect to line segment positions where a smaller T makes the metric more challenging, $mAP^J$: mean average precision of junction positions, and $AP^H$: average precision of rasterized line heatmaps. See [35] for more details on these metrics. We show and compare their F-scores in **SuppMat:Sect.7**.

Our full models show the best hole-robust performance at all metrics on both standard test sets (except for a second best $AP^H$ on York Urban 0-10%). Also, our approach contributes to improved ordinary detection without hole as well: on the Wireframe test set [11] without hole, our full models outperform all of the recent models. On the York Urban dataset [5] without hole, our full model applied to the HAWP framework is the second best right after the very recent F-Clip(HR) [3] (preprint work yet unverified by peer reviews) on $sAP^5$ and $AP^H$, and is the best on $mAP^J$. Note that our full model always significantly outperforms its baselines, either HAWP [31] or HT-HAWP [16], with very large margins. In Section 4.2, we show improvements by our approach applied to more various basic frameworks.

In Figure 5, we show Precision-Recall curves for $sAP^{10}$ with respect to models tested in Table 1. The curves demonstrate that models with our full approach significantly outperform existing models when holes are given, and also works better than those when no hole is given. We show PR curves for $sAP^{10}$ on the 0-10% hole and more PR curves for the $AP^H$ metric in **SuppMat:Sect.8**.

Note that F-Clip(HR) [3] and LETR [29] are fundamentally different from the other frameworks thus hard to be compared in parallel. F-Clip(HR) is mostly benefited from using a HRNet [25] backbone in a complex cascade architecture, whose complexity and properties differ fundamentally from stacked HourGlass [21] shared among recent works as well as F-Clip(HG2) [3]. LETR largely relies on Transformer [26] encoder-decoder where a CNN backbone plays only to extract features, which also makes it fundamentally different from the other frameworks.

Once F-Clip(HR) and LETR are excluded from Table 1, our full model, HAWP + Our full approach, presents the best numbers at all metrics on the York Urban dataset [5] regardless with or without hole. Note that the York Urban dataset is not entirely suitable for evaluating wireframe detection: it is not fully annotated as pointed out in [11, 16], and was created based on a Manhattan assumption (scenes were built on a cartesian grid) that largely differs from the wireframe concept.

| Approach | | | | Wireframe Test Set [11] | | | | | | | | | | | | York Urban Dataset [5] | | | | | | | | | | | |
|---|---|---|---|---|---|---|---|---|---|---|---|---|---|---|---|---|---|---|---|---|---|---|---|---|---|---|---|
| | | | | 10-30% Hole | | | | 0-10% Hole | | | | without Hole | | | | 10-30% Hole | | | | 0-10% Hole | | | | without Hole | | | |
| | 3.2 | 3.3 | 3.4 | $sAP^5$ | $sAP^{10}$ | $mAP^J$ | $AP^H$ | $sAP^5$ | $sAP^{10}$ | $mAP^J$ | $AP^H$ | $sAP^5$ | $sAP^{10}$ | $mAP^J$ | $AP^H$ | $sAP^5$ | $sAP^{10}$ | $mAP^J$ | $AP^H$ | $sAP^5$ | $sAP^{10}$ | $mAP^J$ | $AP^H$ | $sAP^5$ | $sAP^{10}$ | $mAP^J$ | $AP^H$ |
| L-CNN [35] | | | | 32.83 | 35.78 | 40.4 | 61.3 | 47.73 | 51.68 | 53.2 | 75.3 | **58.91** | **62.86** | 59.4 | **80.3** | 14.65 | 16.16 | 21.9 | 44.6 | 20.23 | 22.15 | 27.5 | 54.6 | 24.33 | 26.39 | 30.4 | **58.0** |
| | ✓ | | | 39.57 | 43.17 | 45.4 | 67.2 | 52.26 | 56.50 | 56.4 | 76.1 | 57.58 | 61.59 | 59.0 | 79.3 | 18.99 | 20.69 | **25.6** | 42.7 | **23.86** | **26.06** | **30.9** | 49.2 | **25.86** | **28.03** | **32.4** | 51.5 |
| | ✓ | ✓ | | **40.84** | **44.63** | **46.2** | **68.9** | **54.01** | **58.30** | **57.3** | **77.5** | 58.88 | 62.85 | **59.6** | 80.1 | **19.02** | **20.91** | 25.1 | **49.6** | 23.47 | 25.61 | 30.0 | **55.5** | 25.25 | 27.42 | 31.3 | 57.6 |
| HT-LCNN [16] | | | | 33.56 | 36.48 | 41.4 | 57.1 | 48.89 | 52.79 | 54.2 | 76.6 | 60.33 | 64.22 | 60.6 | 81.6 | 15.84 | 17.48 | 23.5 | 40.4 | 21.47 | 23.62 | 29.3 | 50.1 | **25.72** | **28.02** | **32.5** | 53.0 |
| | ✓ | | | 42.08 | 45.69 | 47.6 | 70.8 | 55.21 | 59.29 | 58.5 | 78.7 | 60.20 | 64.11 | 60.9 | 81.2 | 19.12 | 21.17 | **26.1** | 44.7 | 23.36 | 25.63 | **30.7** | 50.0 | 24.87 | 27.22 | 31.9 | 52.0 |
| | ✓ | ✓ | | **43.35** | **47.06** | **48.3** | **72.1** | **57.02** | **61.02** | **59.3** | **80.2** | **61.57** | **65.40** | **61.6** | **82.4** | **19.77** | **21.63** | 25.7 | **51.5** | **23.73** | **25.95** | 29.9 | **56.6** | 25.37 | 27.58 | 31.1 | **58.4** |
| (a): HAWP [31] | | | | 34.80 | 37.74 | 40.8 | 64.5 | 50.80 | 54.84 | 54.0 | 79.4 | 62.52 | 66.49 | 60.2 | 85.0 | 15.79 | 17.48 | 22.5 | 46.3 | 21.61 | 23.79 | 28.6 | 57.3 | 26.15 | 28.54 | 31.6 | 61.3 |
| (b): | ✓ | | | 44.00 | 47.78 | 46.9 | 72.3 | 57.37 | 61.64 | 57.8 | 82.6 | 62.25 | 66.11 | 60.2 | 84.9 | 19.06 | 21.01 | 25.0 | 51.1 | 23.56 | 25.86 | 29.7 | 57.8 | 25.33 | 27.53 | 31.0 | 60.2 |
| (c): | ✓ | ✓ | | 44.17 | 48.11 | 47.1 | 73.1 | 57.99 | 62.35 | 58.2 | 82.9 | 62.60 | 66.60 | 60.5 | 85.1 | 20.30 | 22.34 | 25.8 | 53.0 | 24.90 | 27.35 | 31.2 | **60.6** | 26.69 | 29.08 | 32.4 | 62.3 |
| (d): | ✓ | | ✓ | 47.36 | 51.15 | **49.6** | 74.5 | 61.16 | 65.30 | 60.7 | 84.2 | 65.60 | 69.34 | 62.9 | 86.2 | 20.50 | 22.58 | **27.0** | 52.0 | 25.52 | 27.99 | **31.7** | 59.2 | 27.13 | 29.54 | **32.9** | 61.2 |
| (e): | ✓ | ✓ | ✓ | **47.59** | **51.50** | **49.6** | **74.9** | **61.83** | **65.98** | **61.0** | **85.0** | **65.98** | **69.76** | **63.0** | **86.9** | **20.81** | **23.01** | **27.0** | **53.4** | **25.62** | **28.01** | **31.7** | 60.2 | **27.23** | **29.58** | 32.8 | **62.6** |
| (f): HT-HAWP [16] | | | | 35.64 | 38.54 | 41.8 | 65.5 | 51.58 | 55.50 | 55.1 | 80.1 | 63.26 | 67.12 | 61.3 | 85.7 | 15.57 | 17.18 | 23.2 | 46.8 | 21.45 | 23.63 | 29.0 | 56.6 | 25.31 | 27.65 | **31.9** | 60.7 |
| (g): | ✓ | | | 45.11 | 48.94 | 48.2 | 74.4 | 58.59 | 62.85 | 59.2 | 83.7 | 63.31 | 67.19 | 61.4 | 85.5 | 18.62 | 20.60 | 25.4 | 52.8 | 23.47 | 25.72 | **30.6** | 59.3 | 25.02 | 27.32 | 31.6 | **61.3** |
| (h): | ✓ | ✓ | | 45.72 | 49.56 | 48.0 | 75.0 | 59.33 | 63.59 | 59.1 | 84.0 | 63.56 | 67.49 | 61.2 | 85.6 | 19.44 | 21.39 | 24.6 | 52.8 | 24.35 | 26.62 | 29.6 | **59.6** | 25.91 | 28.16 | 30.5 | 61.0 |
| (i): | ✓ | | ✓ | 46.82 | 50.37 | 49.2 | 75.0 | 60.50 | 64.48 | 60.3 | 84.4 | 64.90 | 68.59 | 62.4 | 86.4 | 18.99 | 20.89 | **25.9** | **53.4** | 23.23 | 25.34 | 30.3 | 59.1 | 24.84 | 26.92 | 31.3 | 61.2 |
| (j): | ✓ | ✓ | ✓ | **48.02** | **51.82** | **49.9** | **76.6** | **62.21** | **66.31** | **61.3** | **85.5** | **66.19** | **69.92** | **63.2** | **87.0** | **19.96** | **21.92** | 25.4 | 52.9 | **24.72** | **26.94** | 29.8 | 57.6 | **26.06** | **28.26** | 30.9 | 59.2 |
| F-Clip(HG2) [3] | | | | 33.51 | 36.68 | / | 65.3 | 49.46 | 53.87 | / | 79.5 | 61.25 | 65.77 | / | 84.3 | 15.86 | 17.54 | / | 48.6 | 22.02 | 24.07 | / | 58.6 | 27.10 | 29.26 | / | 62.1 |
| | ✓ | | | 43.86 | 48.05 | / | 73.6 | 56.74 | 61.57 | / | 82.2 | 61.38 | 65.81 | / | 84.1 | 20.10 | 22.27 | / | 54.4 | 25.39 | 27.84 | / | 60.6 | 27.35 | 29.60 | / | 62.9 |
| | ✓ | ✓ | | **44.92** | **49.14** | / | **74.6** | **58.10** | **63.01** | / | **83.2** | **62.38** | **66.82** | / | **84.8** | **20.76** | **22.91** | / | **54.7** | **25.57** | **28.03** | / | **60.9** | **27.50** | **29.83** | / | **63.2** |
| F-Clip(HR) [3] | | | | 34.56 | 37.53 | / | 66.3 | 51.44 | 55.60 | / | 80.6 | 64.30 | 68.34 | / | 85.7 | 16.53 | 18.25 | / | 50.4 | 23.53 | 25.76 | / | 61.3 | 28.49 | 30.80 | / | **65.0** |
| | ✓ | | | **47.39** | **51.48** | / | **76.0** | **61.36** | **65.78** | / | **84.7** | **65.31** | **69.39** | / | **86.1** | **22.50** | **24.57** | / | **56.4** | **28.27** | **30.88** | / | **62.9** | **29.98** | **32.50** | / | 64.9 |

Table 2: **Applying our approach to various recent frameworks.** Approach in Section 3.2: *'Conditional Training'*, 3.3: *'RGB GAN'*, and 3.4: *'Pseudo Labeling'*. Best metrics within each framework group are marked in **bold** font. Our approach can be applied to various recent frameworks sharing the *stacked HourGlass* [21] backbone, and is consistently effective either with hole or without hole in general. See **SuppMat:Sect.8** to find PR curves on (a),(c),(e) and (f),(h),(j).

### 4.2. Applying to Various Frameworks

In Table 2, our approach was tested with various recent frameworks, L-CNN [35], HT-LCNN [16], HAWP [31], HT-HAWP [16] and F-Clip(HG2) [3], which share the stacked HourGlass [21] backbone. The results show that applying our full or partial approach is consistently effective to improve the performance on both test sets either without or with hole, except baseline L-CNN on the Wireframe test set [11] without hole and baseline HT-LCNN on York Urban [5] without hole. See **SuppMat:Sect.7** for their F-scores.

We also tested on F-Clip(HR) [3] (gray group in Table 2), whose backbone is based on HRNet [25] that fundamentally differs to stacked HourGlass [21]. Applying conditional training enhanced all metrics notably, but we could not turn this backbone into an RGB GAN generator in spite of various attempts. We will explore how to adapt our approach to this heterogeneous framework in the future work.

**SuppMat:Sect.8** shows PR curves on (a),(c),(e) and (f),(h),(j) in Table 2. These curves prove that our approach yet without pseudo labeling is already significantly effective in improving hole-robustness, which is even further improved with pseudo labeling. Also, our approach enhances the ordinary detection performance without hole as well.

### 4.3. Qualitative Comparisons

In this section, we demonstrate the qualitative performance of our approach by taking a few examples from the Wireframe test set [11] and the York Urban [5], with hole and without hole. Figure 6 (a)-(c) show that existing works do not know how to handle hole occlusion, thus fail to detect inside the hole or mistakenly detect lines around the hole boundaries. In contrast, our approach in Figure 6 (d) detects lines and junctions well even across the large holes. Figure 7 shows that our approach still works consistently well on the ordinary detection when no hole is given. See **SuppMat:Sect.12** for more qualitative comparisons, and real-world examples with foreground object occlusions.

## 5. Conclusions

We propose a novel approach to detect wireframes in a hole-robust manner unlike any of the previous works. We derive this goal by introducing conditional training with masked data generation and special training techniques with respect to the hole size and hole placement, combining RGB GAN in the backbone network to let it better understand the scene composition principles while learning to regenerate hidden scene contents, and semi-supervised learning with pseudo labeling to further enlarge the model capacity.

As demonstrated with extensive experiments and analysis in Section 4 and **Suppmat:Sect.6-12**, our approach is highly robust to hole occlusion, which significantly outperforms existing works that are incapable to properly detect across holes, and even improves the ordinary detection performance when no hole is given. In the future work, we will further extend the approach to different backbone architectures, and explore global attention to enable even deeper scene understanding by merging local and global features.

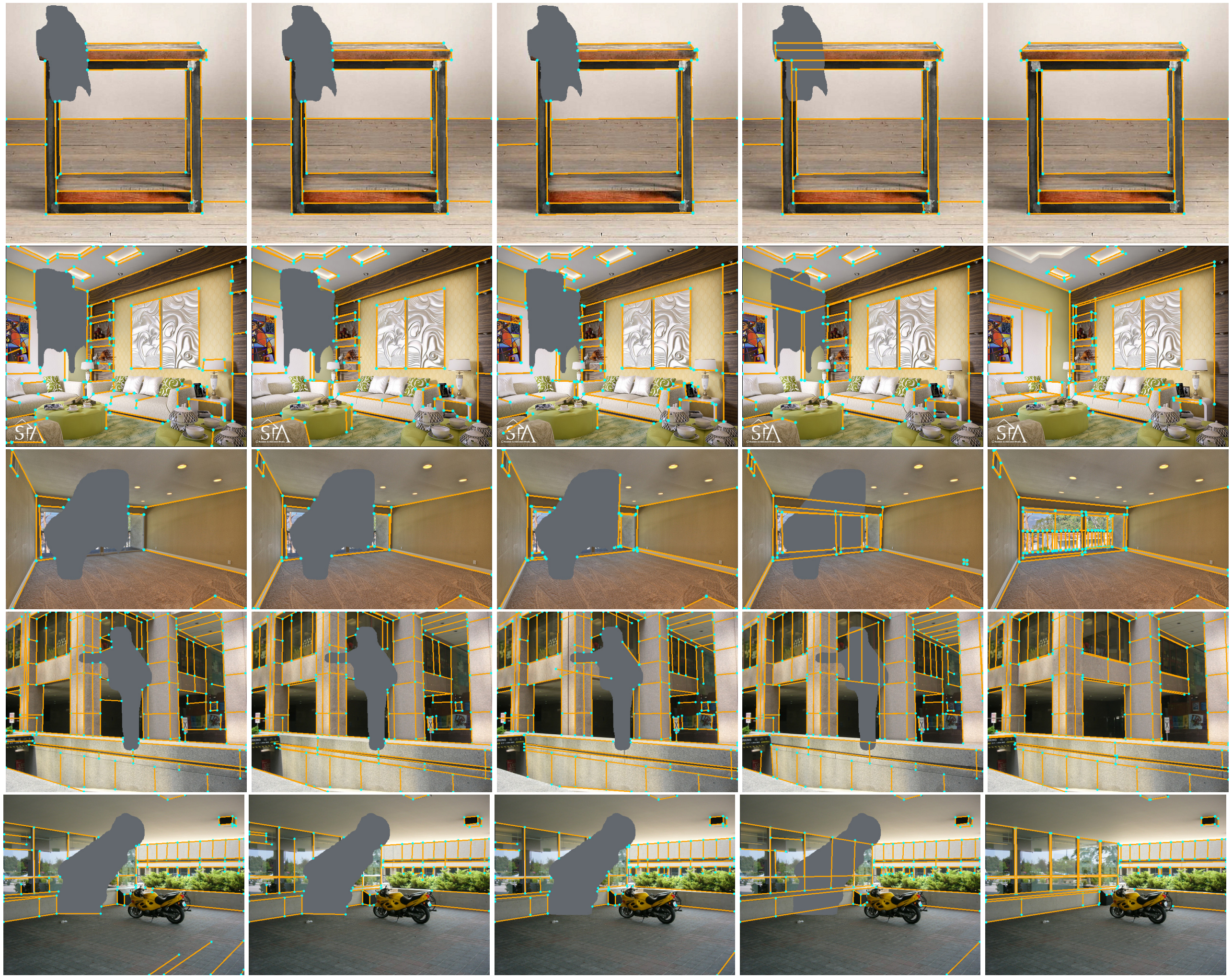

(a) HT-LCNN [16] (b) HT-HAWP [16] (c) F-Clip(HR) [3] (d) Ours (e) Ground Truth

Figure 6: **Results on the Wireframe test set [11] (1st to 3rd rows) and York Urban dataset [5] (4th and 5th row) *with* hole.** (a)-(c) Existing works do not know how to handle holes occluding the scene. (d) Our full approach (using HT-HAWP [16] as the basic framework) robustly detect lines and junctions regardless of large holes given.

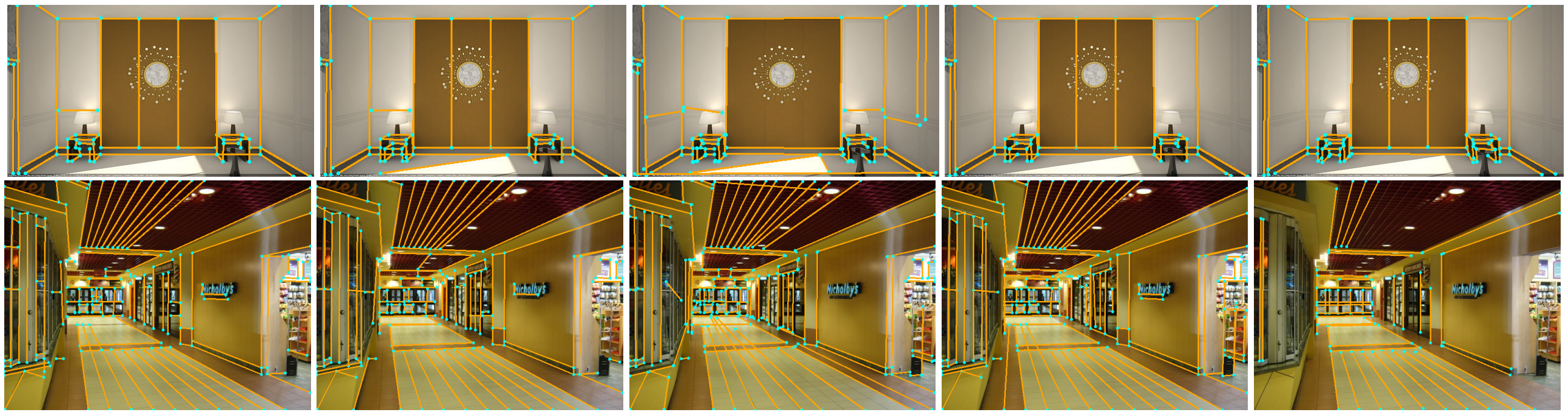

(a) HT-LCNN [16] (b) HT-HAWP [16] (c) F-Clip(HR) [3] (d) Ours (e) Ground Truth

Figure 7: **Results on the Wireframe test set [11] (1st row) and York Urban dataset [5] (2nd row) *without* hole.** Our full approach (using HT-HAWP [16] as the basic framework) still performs well on the ordinary detection without holes given.

# Supplementary Material: Hole-robust Wireframe Detection

Naejin Kong, Kiwoong Park, Harshith Goka
Samsung Research AI Center
{naejin.kong, kyoong.park, h9399.goka}@samsung.com

## 1. Introduction

In this supplementary material, we show more details on the approach and further analysis, experiments and results.

# Contents

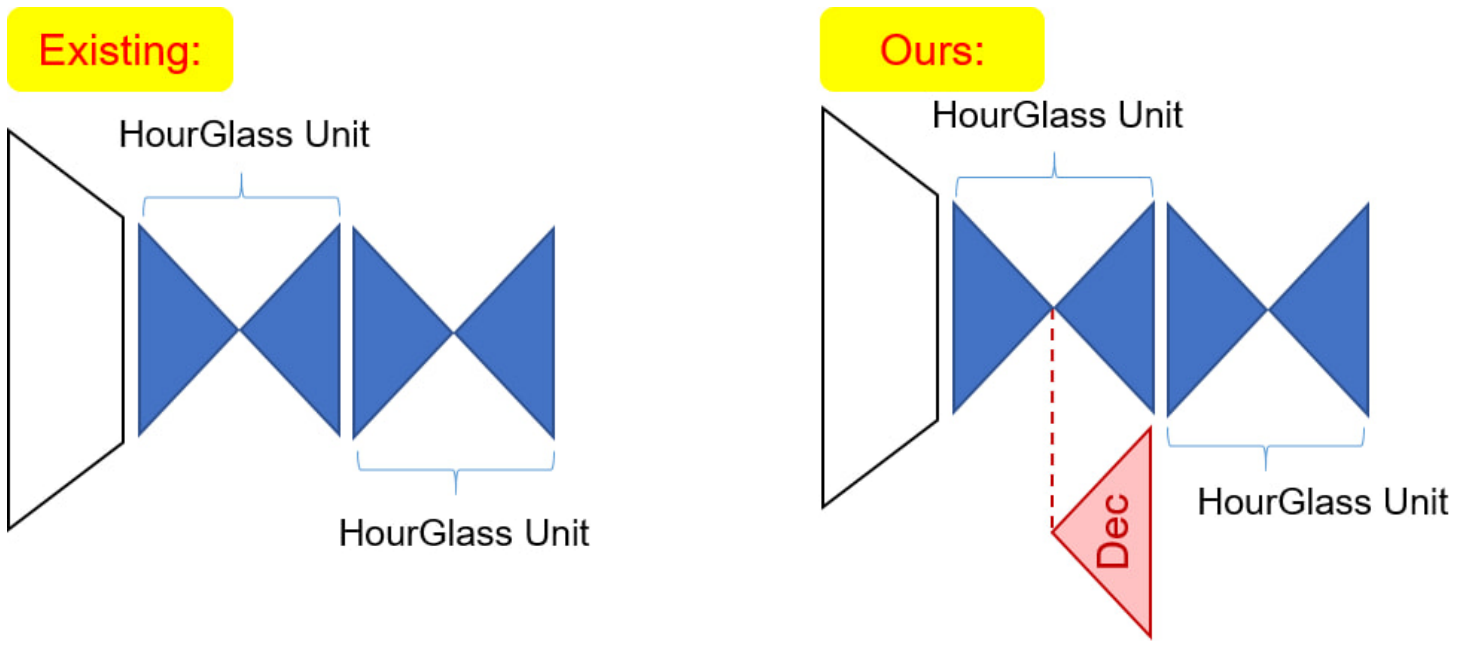

(a) **HG2** Architecture

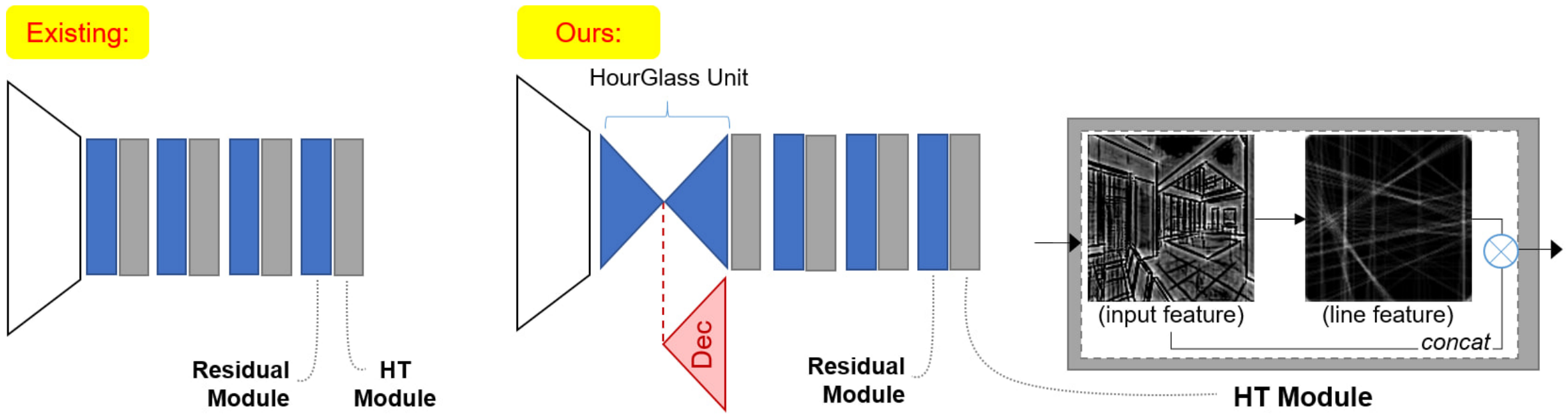

(b) **HT-HG** Architecture

Figure 1: **Backbone architectures**. (a) **HG2** architecture. (Left): HourGlass units are stacked twice. (Right): Our RGB decoder is connected at the first bottleneck. (b) **HT-HG** architecture. (Left): A Hough-transform module is attached right after each Residual Module. See Figure 2 for details of the Residual Module. (Right): We replace the first Residual Module with a full HourGlass unit, and connect an RGB decoder at its bottleneck (see Figure 2 for details).

## 2. Details on RGB GAN

### 2.1. Backbone Network

The backbone network first downsamples a 512×512 input RGB image to 128×128 with 256 channels. Then, it stacks units adopted from HourGlass [9] as follows:

**HG2 Architecture:** A full HourGlass unit is adopted from [9]. In this backbone architecture, the HourGlass units are stacked up twice as shown in Figure 1(a): Existing. L-CNN [15], HAWP [12] and F-Clip(HG2) [1] frameworks use this **HG2** type backbone.

**HT-HG Architecture:** A Residual Module is a basic building block of the HourGlass network [9]. This module takes an input feature $\mathbf{x}$ in any channels and produces a residual output feature in $p * 2$ channels that is added to the input feature, where the original resolution of the input feature is preserved (see Figure 2 for details). In this backbone architecture, the Residual Modules are stacked four times, where a Hough-transfrom module (see Figure 1(b): Existing) is inserted right after each of the Residual Modules. HT-LCNN [15, 7] and HT-HAWP [12, 7] frameworks use this **HT-HG** type backbone.

### 2.2. RGB Decoder for RGB GAN

Our RGB decoder is connected at the bottleneck of the first HourGlass unit. Such a case on the **HG2** architecture is shown in Figure 1(a): Right. In case of the **HT-HG** architecture, we replace the first Residual Module with a full HourGlass unit and connect the RGB decoder at its bottleneck as shown in Figure 1(b): Right. We depict the RGB decoder architecture in detail in Figure 2. The last layer of the RGB decoder is a simple convolutional layer (one 1×1 convolution to compress channels and one batch normalization) that achieves a final 128×128 image in RGB channels. In practice, to better regularize the GAN training, we use spectral normalization [8] for the whole backbone network along with all branches, including the RGB decoder but excluding the Hough-transform modules, as well as a discriminator to be addressed in Section 2.3.

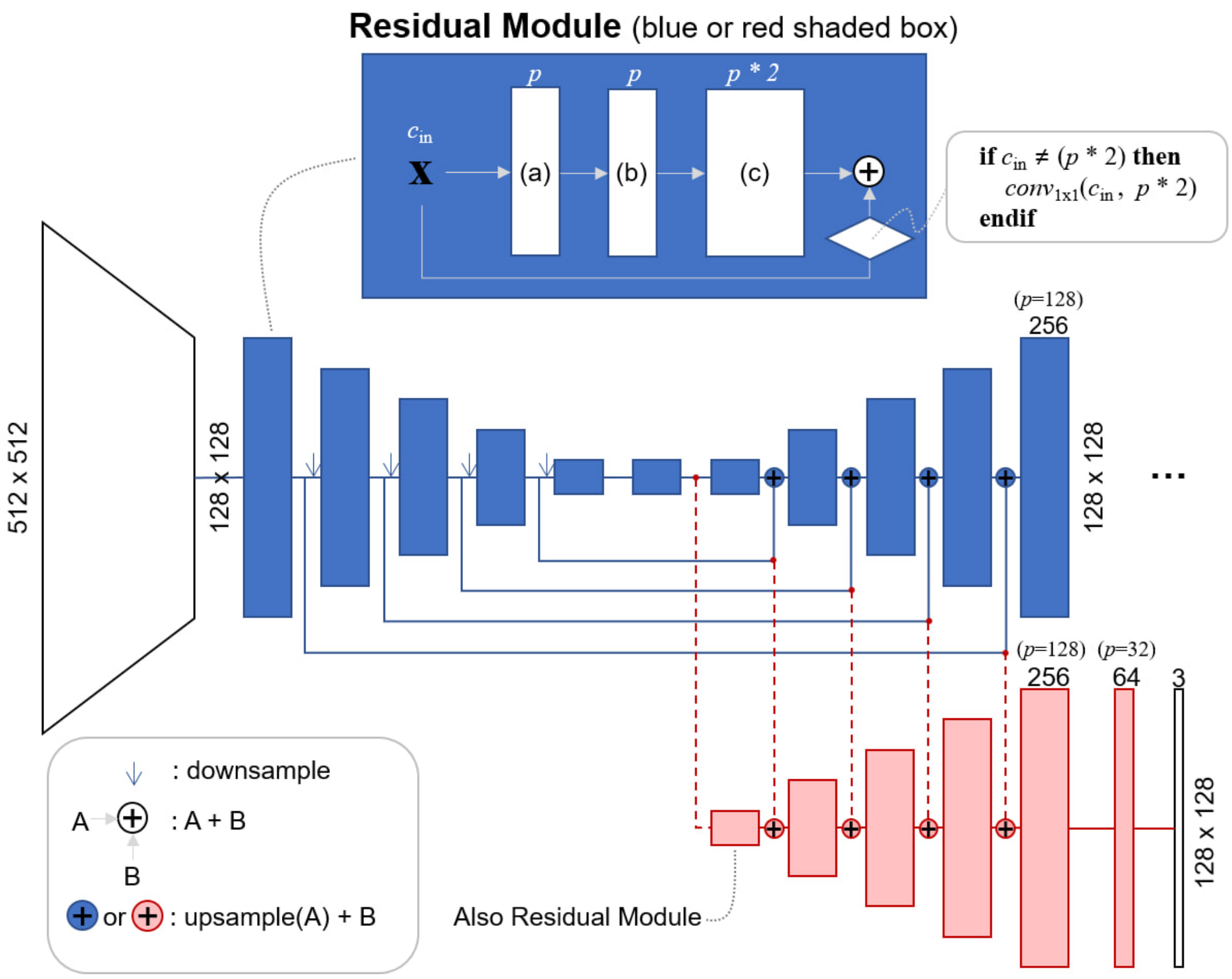

Figure 2: **First HourGlass unit (in blue) and RGB decoder (in red) in our RGB GAN.** The number above the box is the number of output channels. In the Residual Module shown on top, each of convolutional blocks (a), (b), (c) implies $batchnorm + ReLU + conv$, where (a) and (c) use $conv_{1x1}$ and (b) uses $conv_{3x3}$ with padding $1$, hence the original resolution of the input feature is preserved. $c_{\text{in}}$ is the number of channels of the input feature $\mathbf{x}$, and (a), (b), (c) blocks produce $p$, $p$, and $p * 2$ output channels, respectively. If $c_{\text{in}} \neq (p * 2)$, the number of channels of $\mathbf{x}$ is transformed to $p * 2$ before the residual addition. Each of the blue or red shaded boxes indicates a Residual Module shown on top of the figure. The final layer of the RGB decoder is a simple convolutional layer ($conv_{1x1}(64, 3) + batchnorm$) that outputs a $128 \times 128$ image in RGB channels.

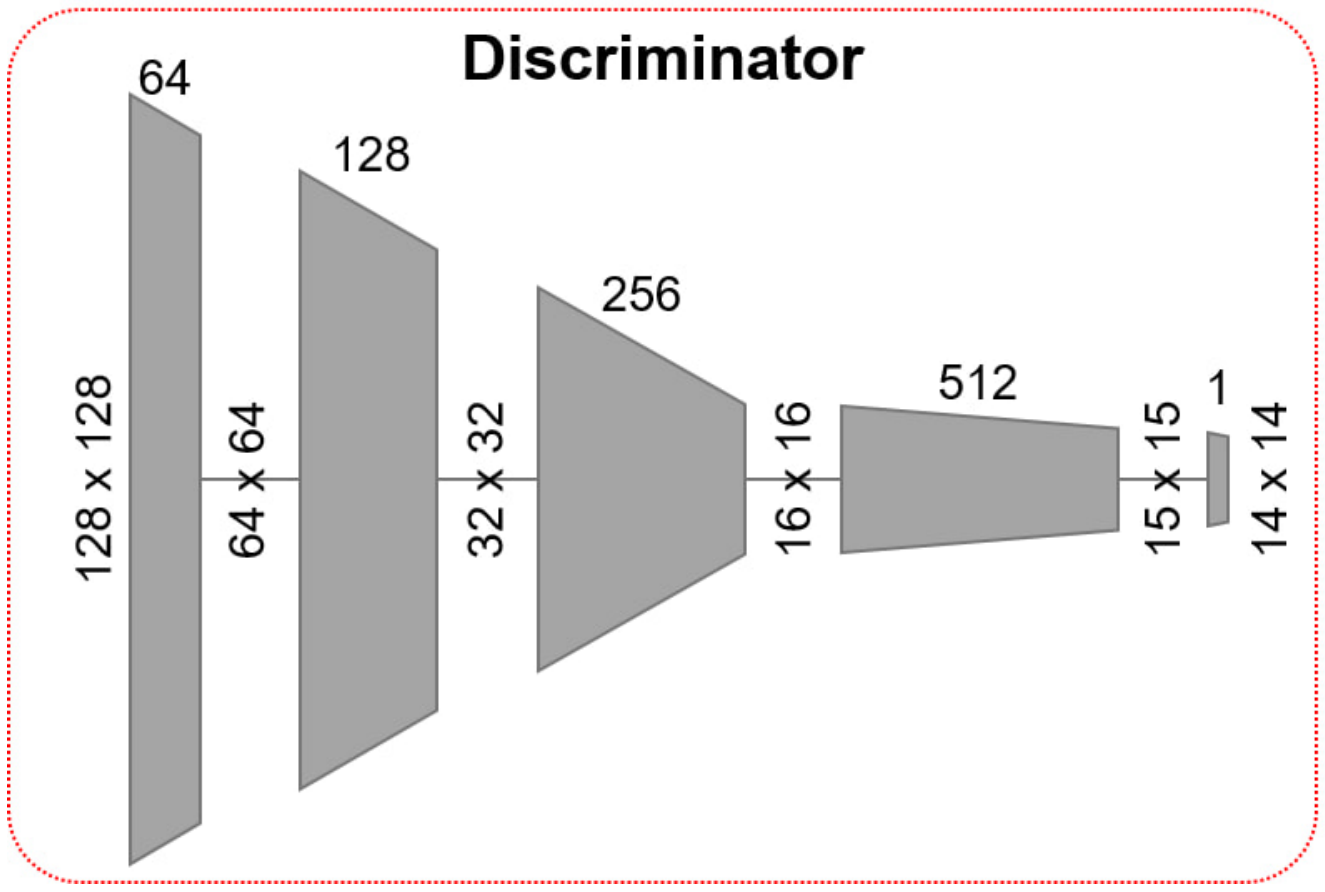

Figure 3: Discriminator architecture in our RGB GAN. The number above the trapezoid is the number of output channels.

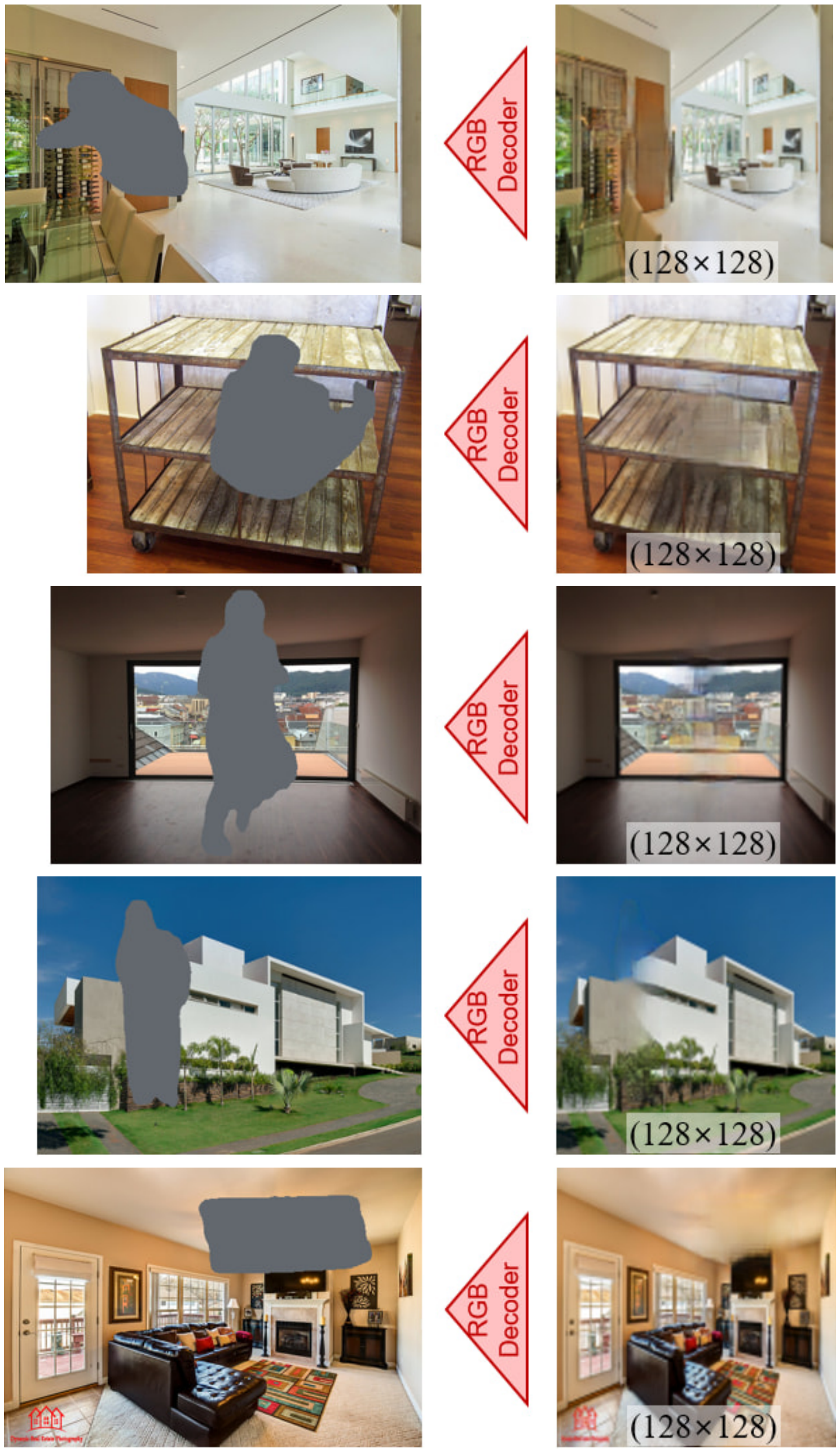

Figure 4: Generated RGB images through our RGB GAN.

### 2.3. Discriminator for RGB GAN

We depict the patch discriminator network architecture in Figure 3. We adapted it from the basic form in PatchGAN [6]. Each pixel in the final $14 \times 14$ output feature corresponds to a $70 \times 70$ patch in the input image. In practice, we use spectral normalization [8] for the discriminator, backbone and RGB decoder to better regularize the GAN training.

### 2.4. Generated RGB Images

We visualize in Figure 4 the generated RGB images in the $128 \times 128$ resolution from our trained RGB GAN. It demonstrates that the backbone network learned to infer the overall structure inside the hole, which will be then, propagated to more robust structural estimation through the other parts of the model.

## 3. Details on Conditional Training

### 3.1. Conditional Data Generation and Training

1. Creating a pool of object silhouettes

   (a) Preparing a pool of 1.3M object silhouette maps by applying Detectron2 [10] (‘*things*’ only from panoptic segmentation with `COCO-PanopticSegmentation/panoptic_fpn_R_50_3x`) on Places365 [14] Test set (328,500 images)

   (b) Leaving 1,107,482 object silhouette maps from the initial pool, where each silhouette map in the pool satisfies both a. and b. such that:
   a. `max(width, height)` of the tight bounding box around the silhouette map is smaller than 512 pixels.
   b. The silhouette map contains less than 78,643.2(= 512 * 512 * 30%) pixels.

   (c) According to the ratio of the hole size (i.e., number of pixels) to 512×512 pixels, grouping the object silhouette maps from the pool into one of 0.1-1%, 1-2%, ..., 29-30% intervals, where each interval has 1% gap

   (d) Randomly selecting 1,500 object silhouette maps from each of the 0.1-1%, ..., 29-30% intervals.

   (e) Finally, 45,000(= 1,500 silhouettes * 30 intervals) object silhouette maps are prepared.

2. Conditional data generation
   - Used both for **Avoid-isolation** and **Random Placement** training

   (a) Generating a pool of $N$ *mask* maps for each input image in the training set (we use $N$ = 10), and for each of the 0.1-1%, 1-2%, ..., 9-10% hole size intervals (i.e., 10 intervals in total)

   (b) A *mask* map within a hole size interval is created as follows:
   a. Randomly select an object silhouette map within the interval size, from the silhouette pool made at Step 1.
   b. Superimpose the chosen silhouette onto a 512×512 region, by avoiding isolated components (see Algorithm 1 below).
   c. A 512×512 *mask* map with the silhouette hole is created.

   (c) Finally, for each image in the training set, $N$ *mask* maps at each interval thus in total [ $N$ * 10 intervals ] *mask* maps are prepared in advance.

3. Conditional training

   (a) Training by choosing a random size hole

      i. **Case1: Avoid-isolation.** For each image, randomly selecting one from all of the [ $N$ * 10 intervals ] *mask* maps corresponding to the image
      **Case2: Random Placement.** For each image, randomly selecting one from all of the [ $N$ * 10 intervals * $T$ images ] *mask* maps, where $T$ = number of all images in the dataset

      ii. Superimposing the hole in the *mask* map onto the 512×512 resized image to create a pair of $\langle$*image_with_hole, mask*$\rangle$

   (b) Training by progressively increasing the hole size along with epochs

      i. **Case1: Avoid-isolation.** For each image and at the designated hole size interval, randomly selecting one from the $N$ *mask* maps corresponding to the image
      **Case2: Random Placement.** For each image and at the designated hole size interval, randomly selecting one from the [ $N$ * $T$ images ] *mask* maps, where $T$ = number of all images in the dataset

      ii. Superimposing the hole in the *mask* map onto the 512×512 resized image to create a pair of $\langle$*image_with_hole, mask*$\rangle$

### 3.2. Avoid-Isolation Algorithm

First, we define the following three hole types:

1. Hole Type 1:
   - The hole overlaps with line segment(s), and also does not contain any junction.
2. Hole Type 2:
   - One or more junction(s) are contained in the hole.
   - The number of line segments whose endpoints are both contained in the hole, is $\leq 1$.
3. Hole Type 3:
   - One or more junction(s) are contained in the hole.
   - Satisfying the following criteria that can be made (sub-)optimal by repeated searching (see Algorithm 1: L10-20):
     - Fewest number of line segments whose endpoints are both contained in the hole
     - Largest total length of line segments that (at least partially) overlap with the hole

Note that Algorithm 1 attempts to create a mask by avoiding isolated components, but there can be exceptional cases that do not allow to strictly avoid isolation after all attempts, and in that case Hole Type 3 is used, which creates a mask that may contain some isolated components.

**Algorithm 1** Avoid-Isolation Algorithm

```
 1: function AVOIDISOLATION(s, K)          ▷ s: a silhouette, K: number of max attempts (we use 500)
 2:     if rand(0, 1) > 0.8 then          ▷ By 20% chance, trying to create Hole Type 1 that contains no junction
 3:         mask ← randomly_place_hole_until_Hole_Type_1_found(s)
 4:         if succeed to find Hole_Type_1 within K attempts then
 5:             return mask
 6:         end if
 7:     end if
 8:     mask ← randomly_place_hole_until_Hole_Type_2_found(s)
 9:     if fail to find Hole_Type_2 after K attempts then
10:         n_prev ← inf
11:         m_prev ← 0
12:         for k ← 1 to K do
13:             n ← number of line segments whose endpoints are both contained in the hole
14:             m ← total length (= number of pixels) of line segments overlapping with the hole
15:             if n ≤ n_prev and m > m_prev then
16:                 n_prev ← n
17:                 m_prev ← m
18:                 mask ← set_mask(s)          ▷ After many attempts, it becomes (sub-)optimal Hole Type 3
19:             end if
20:         end for
21:     end if
22:     return mask
23: end function
```

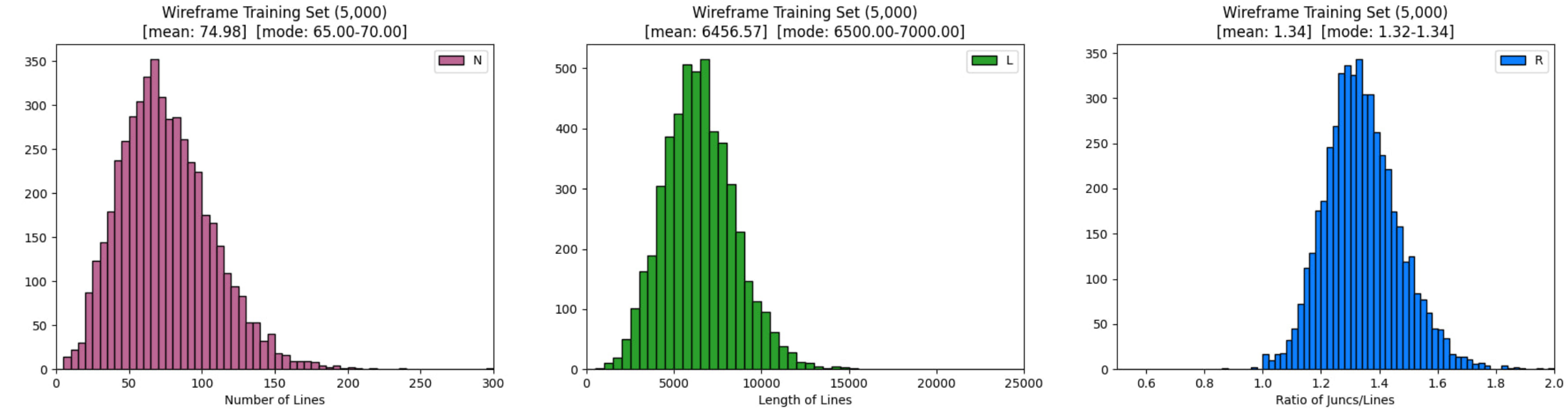

(a) Number of lines (b) Total length of all lines (c) Ratio of # junctions / # lines

Figure 5: **Histograms on three criteria for the ground truth labeled Wireframe training set [5].** Each histogram depicts the frequency of images over the entire Wireframe training set, with respect to the bins corresponding to either (a) number of lines, (b) total length of all lines, or (c) ratio of $\frac{\text{\# of Junctions}}{\text{\# of Lines}}$ in an image.

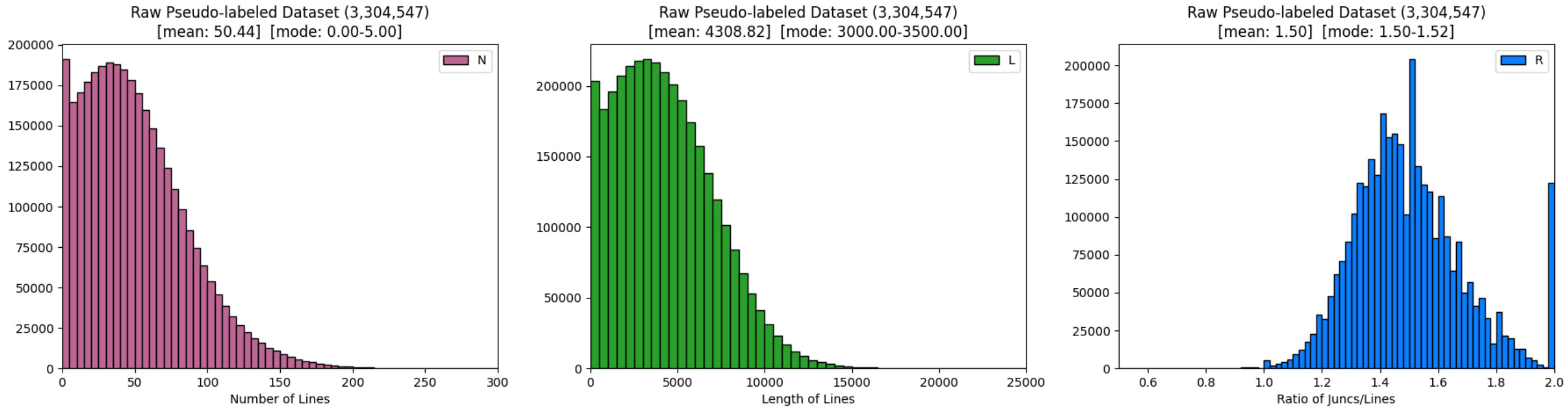

(a) Number of lines (b) Total length of all lines (c) Ratio of # junctions / # lines

Figure 6: **Histograms on three criteria for the *raw* pseudo-labeled dataset *before* applying the thresholds.** The distributions are either skewed as in (a) and (b), or irregular as in (c) due to outliers from inaccurate detection.

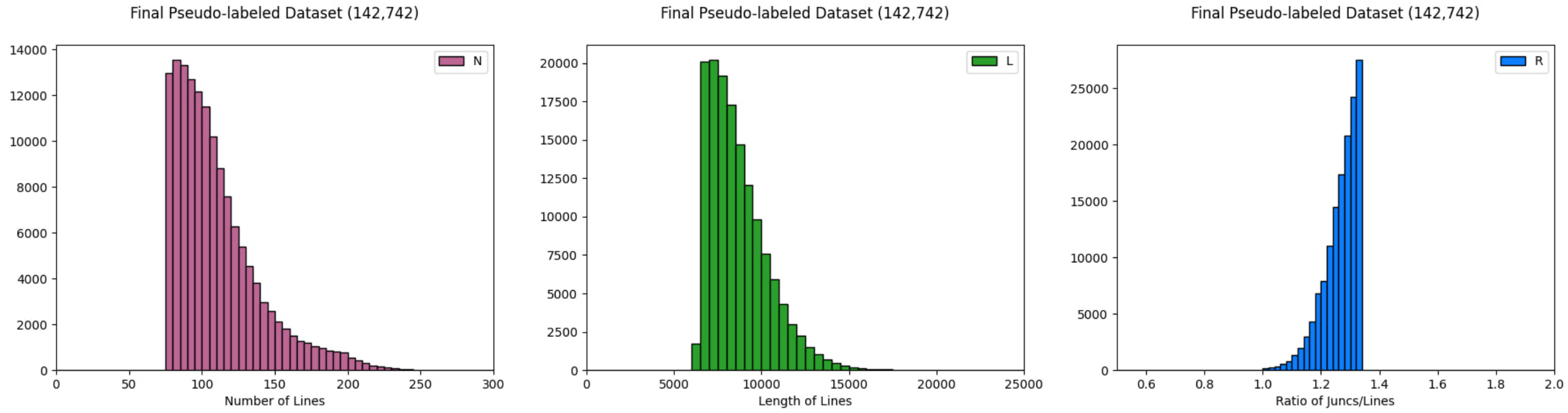

(a) Number of lines (b) Total length of all lines (c) Ratio of # junctions / # lines

Figure 7: **Histograms on three criteria for the *final* pseudo-labeled dataset *after* applying the thresholds.** Once the thresholds are applied, the new distributions look quite alike the ground truth distributions whose other half with respect to the mean value has been cut out.

# 4. Ground Truth and Pseudo Label Distributions on Three Criteria

- Number of lines in an image $> 74.98$
- Total length of all lines in an image $> 6456.57$
- The ratio $\frac{\text{\# of Junctions}}{\text{\# of Lines}}$ in an image $< 1.34$

The threshold values in these three criteria above were determined by inspecting the distributions of ground truth wireframe labels in each image of the original Wireframe training set [5], which follow quite ideal normal distributions. Histograms in Figure 5 depict those distributions on the three criteria.

We extracted the thresholds from the ground truth label distributions instead of pseudo label distributions, because it is hard to obtain accurate distributions from the pseudo labels as they may contain several outliers whose detection had failed badly. As shown in Figure 6, the *raw* pseudo-labeled dataset *before* applying the thresholds appear to be either skewed as in (a) and (b), or irregular as in (c) due to those outliers.

Assuming that ideal distributions of the pseudo labels would be similar to those of the ground truth labels on the three criteria, we applied the thresholds to the candidate pseudo-labeled examples, to pick only a subset of them more likely to have better detected and more dense labels. This effectively removes the outliers from the raw dataset: as shown in Figure 7, the new distributions of the final pseudo-labeled dataset after the thresholds are applied quite resemble the ground truth distributions in Figure 5 whose other half with respect to the mean value, corresponding to each of our threshold values, has been cut out.

## 5. Implementation Details

We implemented our approach using PyTorch, and trained each model on one V100 GPU. We keep the original hyper parameters in the basic frameworks as identical as possible. Here we describe training details of our approach applied to the HT-HAWP [12, 7] framework. When pseudo labeling is not applied, we trained the model on our modified Wireframe dataset 30 epochs in total. We progressively increase the hole size by +1% at every three epochs, such that 0.1-1%, 1-2%, . . . , 9-10%. We kept the default batch size 6 and learning rate 4e-4. When pseudo labeling is enabled, we first trained on the large pseudo-labeled data 10 epochs in total, while incrementing the hole size +1% at every epoch. We used a larger batch size 8, but kept the default learning rate 4e-4. We then fine-tuned on the modified Wireframe dataset 10 epochs in total, again +1% increment of the hole size at every epoch, while using a smaller batch size 6 and smaller learning rate 4e-5.

# 6. Extra Analysis and Studies

## 6.1. Ablation Studies on Conditional Training with Various Settings

| | | | | Wireframe Test Set [5] | | | | | | | | | | | |
|---|---|---|---|---|---|---|---|---|---|---|---|---|---|---|---|
| | | | | 10-30% Hole | | | | 0-10% Hole | | | | without Hole | | | |
| | (a) | (b) | (c) | $sAP^5$ | $sAP^{10}$ | $mAP^J$ | $AP^H$ | $sAP^5$ | $sAP^{10}$ | $mAP^J$ | $AP^H$ | $sAP^5$ | $sAP^{10}$ | $mAP^J$ | $AP^H$ |
| HT-HAWP [12, 7] | | | | 35.64 | 38.54 | 41.8 | 65.5 | 51.58 | 55.50 | 55.1 | 80.1 | 63.26 | 67.12 | 61.3 | **85.7** |
| | ✓ | | | 44.79 | 48.72 | **48.3** | 74.1 | 58.04 | 62.18 | 58.5 | 83.0 | 62.78 | 66.59 | 60.7 | 85.4 |
| | ✓ | ✓ | | 44.99 | 48.92 | 48.2 | 67.8 | 58.27 | 62.55 | 58.7 | 78.8 | 63.00 | 66.90 | 60.9 | 80.9 |
| | ✓ | ✓ | ✓ | **45.11** | **48.94** | 48.2 | **74.4** | **58.59** | **62.85** | **59.2** | **83.7** | **63.31** | **67.19** | **61.4** | 85.5 |

(a): Training with masked data with hole
(b): Progressive hole size increment along with epochs
(c): Placing a hole by avoiding isolated components
(a)+(b)+(c): Conditional training

Table 1: Ablation study on conditional training.

| (a) | (b) | (c) | Wireframe Test Set [5] | | | | | | | | | | | |
|---|---|---|---|---|---|---|---|---|---|---|---|---|---|---|
| | | | 10-30% Hole | | | | 0-10% Hole | | | | without Hole | | | |
| (a) | (b) | (c) | $sAP^5$ | $sAP^{10}$ | $mAP^J$ | $AP^H$ | $sAP^5$ | $sAP^{10}$ | $mAP^J$ | $AP^H$ | $sAP^5$ | $sAP^{10}$ | $mAP^J$ | $AP^H$ |
| (Trained on masked data with *"0.1-10% hole size"*) | | | | | | | | | | | | | | |
| ✓ | | | 44.79 | 48.72 | 48.3 | **74.1** | 58.04 | 62.18 | 58.5 | 83.0 | 62.78 | 66.59 | 60.7 | **85.4** |
| ✓ | ✓ | | 44.99 | **48.92** | 48.2 | 67.8 | **58.27** | **62.55** | **58.7** | 78.8 | **63.00** | **66.90** | **60.9** | 80.9 |
| (Trained on masked data with *"0.1-30% hole size"*) | | | | | | | | | | | | | | |
| ✓ | | | **45.04** | 48.89 | **48.6** | 74.0 | 57.63 | 61.79 | **58.7** | **83.1** | 62.45 | 66.28 | **60.9** | **85.4** |
| ✓ | ✓ | | **45.04** | 48.91 | **48.6** | 70.7 | 57.67 | 62.02 | 58.5 | 80.1 | 62.20 | 66.17 | 60.6 | 82.7 |

(a): Training with masked data with hole
(b): Progressive hole size increment along with epochs

Table 2: Effect of larger hole size around *"0.1-30%"* instead of *"0.1-10%* on training with masked data. The basic framework is HT-HAWP [12, 7].

Table 1 shows an ablation study on our conditional training in various settings. We studied it with the HT-HAWP [12, 7] baseline on the Wireframe test set [5]. The ablations between "applying (a)" and "applying (a) (b)" show that progressively increasing the hole size is better than choosing a random size hole, since this scheme lets the training initially focus on ordinary detection and then, later on, adapt to gradually larger holes thus gives more time before facing more challenging tasks. The ablations between "applying (a) (b)" and "applying (a) (b) (c)" show that placing a hole in the image by carefully avoiding isolated components is more effective than placing it at an entirely random location, because a random placement may result in a scene component entirely hidden if the hole is large. This makes sense, as it would be irrational to expect the network to predict hidden structures without any visible clue left in the input image.

Table 2 studies an effect of larger hole size on training with masked data. Note that above 10%, it is hard to place such a large hole by strictly avoiding isolated components when generating masked data, thus we had to place the hole above 10% at an entirely random location. Since it is not possible to apply the avoid isolation scheme around 10-30% size, we chose to ablate only for (a) and (b) but except (c) avoid-isolation for this study. The numbers in this table show that training with a larger hole size around *"0.1-30%"* is not clearly beneficial at 10-30% hole inference than training with a smaller hole around 0.1-10%, and it works even worse at 0-10% hole inference and without-hole inference in general. Also, according to Table 1, applying the avoid isolation scheme is clearly beneficial. For these reasons, we concluded to use a 0.1-10% hole size range for creating our final conditional training data.

### 6.2. Applying GAN to Other Types of Information

| | Wireframe Test Set [5] | | | | | | | | | | | |
|---|---|---|---|---|---|---|---|---|---|---|---|---|
| | 10-30% Hole | | | | 0-10% Hole | | | | without Hole | | | |
| | $sAP^5$ | $sAP^{10}$ | $mAP^J$ | $AP^H$ | $sAP^5$ | $sAP^{10}$ | $mAP^J$ | $AP^H$ | $sAP^5$ | $sAP^{10}$ | $mAP^J$ | $AP^H$ |
| HT-LCNN [15, 7] + Cond' training | 42.08 | 45.69 | 47.6 | 70.8 | 55.21 | 59.29 | 58.5 | 78.7 | 60.20 | 64.11 | 60.9 | 81.2 |
| + junction heatmap GAN | 29.06 | 31.79 | 31.0 | 48.9 | 37.55 | 41.04 | 37.8 | 59.0 | 43.73 | 47.63 | 40.9 | 65.5 |
| + line heatmap GAN | 40.81 | 44.36 | 46.7 | 69.3 | 53.67 | 57.75 | 57.6 | 77.6 | 58.63 | 62.46 | 60.0 | 80.1 |
| + RGB GAN | **43.35** | **47.06** | **48.3** | **72.1** | **57.02** | **61.02** | **59.3** | **80.2** | **61.57** | **65.40** | **61.6** | **82.4** |
| HT-HAWP [12, 7] + Cond' training | 45.11 | 48.94 | **48.2** | 74.4 | 58.59 | 62.85 | 59.2 | 83.7 | 63.31 | 67.19 | 61.4 | 85.5 |
| + full AFM GAN | 45.41 | 49.02 | 47.9 | 73.2 | 58.96 | 63.03 | 58.9 | 82.6 | 63.52 | 67.24 | 61.1 | 84.7 |
| + distance map GAN | 44.80 | 48.65 | **48.2** | 74.4 | 58.72 | 62.95 | **59.3** | 83.6 | 63.25 | 67.16 | **61.5** | **85.7** |
| + angular maps GAN | 45.10 | 48.72 | 47.7 | 73.6 | 58.94 | 62.90 | 58.7 | 83.1 | 63.44 | 67.11 | 60.9 | 85.1 |
| + RGB GAN | **45.72** | **49.56** | 48.0 | **75.0** | **59.33** | **63.59** | 59.1 | **84.0** | **63.56** | **67.49** | 61.2 | 85.6 |

Table 3: Ablation study on applying GAN to other components.

Table 3 shows how we concluded to apply GAN [3] to RGB generation instead of a junction, line, or attraction field map. The first study was made on HT-LCNN [15, 7] baseline, where combining junction map GAN or line map GAN significantly reduced the performance than without it. We suspect that the GAN is not suitable to estimate accurate sparse image maps for lines and junctions that are used to obtain coordinates information through direct conversion later. The second study was made on HT-HAWP [12, 7] that estimates a 4-D holistic Attraction Field Map (distance map and three more angular orientation maps) representing line segments. We applied GAN [3] to either all components of the AFM or only part of them. This worked quite well since these components of AFM are dense representation maps, but still, RGB GAN showed better numbers on structural metrics. We believe that learning to generate the scene contents through RGB GAN contributes to deeper scene understanding than indirectly going through the generation of structural maps.

### 6.3. Testing the Sole Effect of Pseudo Labeling

| | Wireframe Test Set [5] | | | | | | | | | | | |
|---|---|---|---|---|---|---|---|---|---|---|---|---|
| | 10-30% Hole | | | | 0-10% Hole | | | | without Hole | | | |
| | $sAP^5$ | $sAP^{10}$ | $mAP^J$ | $AP^H$ | $sAP^5$ | $sAP^{10}$ | $mAP^J$ | $AP^H$ | $sAP^5$ | $sAP^{10}$ | $mAP^J$ | $AP^H$ |
| HAWP [12] | 34.80 | 37.74 | 40.8 | 64.5 | 50.80 | 54.84 | 54.0 | 79.4 | 62.52 | 66.49 | 60.2 | 85.0 |
| + Pseudo labeling | **36.51** | **39.39** | **42.6** | **66.3** | **53.29** | **57.07** | **56.5** | **80.9** | **65.45** | **69.13** | **62.8** | **86.5** |
| HT-HAWP [12, 7] | 35.64 | 38.54 | 41.8 | 65.5 | 51.58 | 55.50 | 55.1 | 80.1 | 63.26 | 67.12 | 61.3 | 85.7 |
| + Pseudo labeling | **36.63** | **39.55** | **42.9** | **66.5** | **53.30** | **57.21** | **56.5** | **81.2** | **65.64** | **69.40** | **63.1** | **86.5** |
| | York Urban Dataset [2] | | | | | | | | | | | |
| | 10-30% Hole | | | | 0-10% Hole | | | | without Hole | | | |
| | $sAP^5$ | $sAP^{10}$ | $mAP^J$ | $AP^H$ | $sAP^5$ | $sAP^{10}$ | $mAP^J$ | $AP^H$ | $sAP^5$ | $sAP^{10}$ | $mAP^J$ | $AP^H$ |
| HAWP [12] | 15.79 | 17.48 | 22.5 | 46.3 | 21.61 | 23.79 | 28.6 | 57.3 | 26.15 | 28.54 | 31.6 | 61.3 |
| + Pseudo labeling | **16.37** | **18.22** | **24.2** | **47.6** | **22.43** | **24.77** | **30.0** | **58.0** | **26.75** | **29.18** | **32.8** | **61.6** |
| HT-HAWP [12, 7] | 15.57 | 17.18 | 23.2 | **46.8** | **21.45** | **23.63** | 29.0 | 56.6 | 25.31 | 27.65 | 31.9 | 60.7 |
| + Pseudo labeling | **15.63** | **17.33** | **23.9** | 46.4 | 21.20 | 23.37 | **29.6** | **56.8** | **25.72** | **28.04** | **32.6** | **60.9** |

Table 4: Effect of training on pseudo-labeled data.

In order to test the sole effect of pseudo labeling, we trained HAWP [12] and HT-HAWP [12, 7] baselines on our pseudo-labeled data without conditional masks. We applied 10 epochs on the pseudo data training as well as fine-tuning, same as our full approach works. As shown in Table 4, training on our large pseudo-labeled data helps to outperform the baseline model trained only on small ground truth labeled data in general, but these improvements are not as dramatic as our full approach cases, especially on the hole-robust performance gains. This again proves that the conditional training and RGB GAN in our full approach are important factors to conquer the hole occlusion problem.

## 7. F-Scores on Tables 1 and 2 in the Main Paper

| | Wireframe Test Set [5] | | | | | | | | | York Urban Dataset [2] | | | | | | | | |
|---|---|---|---|---|---|---|---|---|---|---|---|---|---|---|---|---|---|---|
| | 10-30% Hole | | | 0-10% Hole | | | without Hole | | | 10-30% Hole | | | 0-10% Hole | | | without Hole | | |
| | $sF^5$ | $sF^{10}$ | $F^H$ | $sF^5$ | $sF^{10}$ | $F^H$ | $sF^5$ | $sF^{10}$ | $F^H$ | $sF^5$ | $sF^{10}$ | $F^H$ | $sF^5$ | $sF^{10}$ | $F^H$ | $sF^5$ | $sF^{10}$ | $F^H$ |
| L-CNN [15] | 42.71 | 44.87 | 67.1 | 52.19 | 54.61 | 74.4 | 59.12 | 61.31 | 76.9 | 27.00 | 28.36 | 54.6 | 31.99 | 33.53 | 60.1 | 35.42 | 36.92 | 61.9 |
| HT-LCNN [15, 7] | 43.74 | 45.81 | 65.3 | 53.40 | 55.72 | 76.2 | 60.53 | 62.71 | 79.0 | 28.10 | 29.76 | 53.1 | 33.59 | 35.34 | 58.6 | 37.14 | 38.91 | 60.5 |
| F-Clip(HG2) [1] | 44.89 | 47.54 | 69.3 | 54.85 | 57.86 | 77.6 | 62.04 | 64.94 | 80.7 | 28.43 | 30.26 | 56.3 | 33.60 | 35.44 | 62.4 | 37.50 | 39.28 | 64.3 |
| F-Clip(HR) [1] | 45.61 | 48.11 | 69.9 | 55.95 | 58.76 | 78.1 | 63.46 | 66.08 | 81.5 | 29.25 | 31.13 | 57.6 | 35.01 | 37.03 | 64.2 | **38.90** | **40.82** | 66.4 |
| LETR [11] | 45.28 | 48.45 | 71.4 | 57.08 | 60.55 | 80.4 | 62.59 | 66.12 | **83.2** | 28.88 | 31.40 | 57.7 | 34.04 | 37.10 | 64.6 | 37.52 | 40.50 | **66.7** |
| HAWP [12] | 44.94 | 47.04 | 69.1 | 55.29 | 57.71 | 77.4 | 62.63 | 64.87 | 80.6 | 28.03 | 29.67 | 55.6 | 33.86 | 35.56 | 62.7 | 37.88 | 39.61 | 65.3 |
| HT-HAWP [12, 7] | 45.28 | 47.35 | 70.0 | 55.69 | 58.03 | 78.3 | 63.06 | 65.19 | 81.6 | 27.91 | 29.46 | 56.5 | 33.57 | 35.31 | 63.1 | 37.36 | 39.08 | 65.1 |
| HAWP + Our full approach | 53.16 | 55.32 | 73.3 | 62.08 | 64.37 | 80.7 | 64.93 | 67.01 | 82.3 | **32.74** | **34.40** | **59.6** | **37.43** | **39.18** | **64.7** | 38.81 | 40.52 | 66.3 |
| HT-HAWP + Our full approach | **53.48** | **55.63** | **75.1** | **62.57** | **64.90** | **81.8** | **65.44** | **67.58** | 83.0 | 32.35 | 33.81 | 59.2 | 37.30 | 38.89 | 63.9 | 38.68 | 40.33 | 65.1 |

Table 5: **Effect of our full approach, evaluated on the Wireframe test set [5] and York Urban dataset [2], *with* and *without* hole.** $sF^T$ (T: distance threshold) and $F^H$ are F-score measurement for $sAP^T$ and $AP^H$. See text for more details.

| | Approach | | | Wireframe Test Set [5] | | | | | | | | | York Urban Dataset [2] | | | | | | | | |
|---|---|---|---|---|---|---|---|---|---|---|---|---|---|---|---|---|---|---|---|---|---|
| | | | | 10-30% Hole | | | 0-10% Hole | | | without Hole | | | 10-30% Hole | | | 0-10% Hole | | | without Hole | | |
| | 3.2 | 3.3 | 3.4 | $sF^5$ | $sF^{10}$ | $F^H$ | $sF^5$ | $sF^{10}$ | $F^H$ | $sF^5$ | $sF^{10}$ | $F^H$ | $sF^5$ | $sF^{10}$ | $F^H$ | $sF^5$ | $sF^{10}$ | $F^H$ | $sF^5$ | $sF^{10}$ | $F^H$ |
| L-CNN [15] | | | | 42.71 | 44.87 | 67.1 | 52.19 | 54.61 | 74.4 | 59.12 | 61.31 | 76.9 | 27.00 | 28.36 | 54.6 | 31.99 | 33.53 | 60.1 | 35.42 | 36.92 | 61.9 |
| | ✓ | | | 46.39 | 48.54 | 67.7 | 54.38 | 56.76 | 75.2 | 58.28 | 60.50 | **77.4** | 30.48 | 31.81 | 52.6 | **35.43** | **37.15** | 58.1 | **37.24** | **38.80** | 59.6 |
| | ✓ | ✓ | | **47.59** | **49.84** | **69.5** | **55.87** | **58.21** | **75.4** | **59.48** | **61.68** | 77.1 | **30.65** | **32.34** | **56.5** | 34.87 | 36.51 | **60.9** | 36.66 | 38.28 | **62.3** |
| HT-LCNN [15, 7] | | | | 43.74 | 45.81 | 65.3 | 53.40 | 55.72 | 76.2 | 60.53 | 62.71 | 79.0 | 28.10 | 29.76 | 53.1 | 33.59 | 35.34 | 58.6 | 37.14 | 38.91 | 60.5 |
| | ✓ | | | 48.20 | 50.24 | 70.3 | 56.43 | 58.67 | 76.1 | 60.15 | 62.28 | 77.5 | 30.35 | 31.98 | 53.6 | 34.60 | 36.23 | 57.7 | **36.38** | **38.05** | 58.8 |
| | ✓ | ✓ | | **49.23** | **51.41** | **71.7** | **57.95** | **60.18** | **77.8** | **61.35** | **63.51** | **79.2** | **30.62** | **32.08** | **58.4** | **34.68** | **36.43** | **62.0** | 36.29 | 38.02 | **62.9** |
| (a): HAWP [12] | | | | 44.94 | 47.04 | 69.1 | 55.29 | 57.71 | 77.4 | 62.63 | 64.87 | 80.6 | 28.03 | 29.67 | 55.6 | 33.86 | 35.56 | 62.7 | 37.88 | 39.61 | 65.3 |
| (b): | ✓ | | | 50.19 | 52.32 | 71.5 | 58.75 | 61.08 | 78.6 | 62.34 | 64.45 | 80.5 | 30.78 | 32.20 | 57.3 | 35.34 | 37.05 | 62.6 | 37.24 | 38.87 | 64.3 |
| (c): | ✓ | ✓ | | 50.30 | 52.59 | 72.1 | 59.18 | 61.55 | 78.9 | 62.64 | 64.85 | 80.6 | 31.71 | 33.21 | 58.8 | 36.41 | 38.22 | 64.4 | 38.31 | 39.94 | 65.4 |
| (d): | ✓ | | ✓ | 52.87 | 54.83 | 72.7 | 61.63 | 63.87 | 80.0 | **64.95** | 66.88 | 81.7 | 32.52 | 34.11 | 58.1 | 37.37 | **39.31** | 63.9 | **39.00** | **40.76** | 65.1 |
| (e): | ✓ | ✓ | ✓ | **53.16** | **55.32** | **73.3** | **62.08** | **64.37** | **80.7** | 64.93 | **67.01** | **82.3** | **32.74** | **34.40** | **59.6** | **37.43** | 39.18 | **64.7** | 38.81 | 40.52 | **66.3** |
| (f): HT-HAWP [12, 7] | | | | 45.28 | 47.35 | 70.0 | 55.69 | 58.03 | 78.3 | 63.06 | 65.19 | 81.6 | 27.91 | 29.46 | 56.5 | 33.57 | 35.31 | 63.1 | 37.36 | 39.08 | 65.1 |
| (g): | ✓ | | | 50.87 | 53.06 | 72.9 | 59.52 | 61.83 | 79.6 | 63.11 | 65.21 | 81.3 | 30.67 | 32.26 | 59.0 | 35.70 | 37.31 | 63.6 | 37.41 | 39.10 | 65.0 |
| (h): | ✓ | ✓ | | 51.71 | 53.97 | 73.9 | 60.64 | 63.02 | 80.4 | 63.83 | 66.02 | 82.0 | 30.97 | 32.54 | 59.3 | 35.92 | 37.70 | 63.9 | 37.30 | 38.98 | 64.8 |
| (i): | ✓ | | ✓ | 52.37 | 54.27 | 73.5 | 61.15 | 63.41 | 80.7 | 64.25 | 66.33 | 82.1 | 31.07 | 32.52 | **59.6** | 35.74 | 37.34 | **64.0** | 37.46 | 38.98 | **65.2** |
| (j): | ✓ | ✓ | ✓ | **53.48** | **55.63** | **75.1** | **62.57** | **64.90** | **81.8** | **65.44** | **67.58** | **83.0** | **32.35** | **33.81** | 59.2 | **37.30** | **38.89** | 63.9 | **38.68** | **40.33** | 65.1 |
| F-Clip(HG2) [1] | | | | 44.89 | 47.54 | 69.3 | 54.85 | 57.86 | 77.6 | 62.04 | 64.94 | 80.7 | 28.43 | 30.26 | 56.3 | 33.60 | 35.44 | 62.4 | 37.50 | 39.28 | 64.3 |
| | ✓ | | | 50.47 | 53.31 | 72.1 | 58.94 | 62.07 | 79.0 | 62.28 | 65.23 | 80.6 | 31.47 | 33.50 | 58.1 | 36.41 | 38.45 | 63.4 | 38.07 | 39.91 | **64.8** |
| | ✓ | ✓ | | **51.15** | **54.05** | **72.7** | **59.66** | **62.94** | **79.9** | **62.85** | **65.75** | **81.1** | **31.97** | **33.85** | **58.5** | **36.53** | **38.69** | **63.6** | **38.25** | **40.23** | **64.8** |
| F-Clip(HR) [1] | | | | 45.61 | 48.11 | 69.9 | 55.95 | 58.76 | 78.1 | 63.46 | 66.08 | 81.5 | 29.25 | 31.13 | 57.6 | 35.01 | 37.03 | 64.2 | 38.90 | 40.82 | **66.4** |
| | ✓ | | | **52.61** | **55.47** | **73.5** | **61.52** | **64.35** | **80.6** | **64.40** | **66.96** | **82.0** | **33.43** | **35.30** | **59.0** | **38.03** | **40.36** | **64.5** | **39.69** | **41.70** | 65.6 |

Table 6: **Applying our approach to various recent frameworks.** Approach 3.2: *Conditional Training*, Approach 3.3: *RGB GAN*, and Approach 3.4: *Pseudo Labeling*. Best metrics within each framework group are marked in **bold** font. Our approach can be applied to various recent frameworks sharing the *stacked HourGlass* [9] backbone, and is consistently effective both *without* and *with* hole in general.

Table 5 and Table 6 show F-Scores on Tables 1 and 2 in the Main Paper, respectively.

Table 5 shows an almost identical trend as Table 1 in the Main Paper. On the Wireframe test set, our full model (HT-HAWP + Our full approach) shows the best hole-robust performance at all metrics with hole, and the best numbers at $sF^5$ and $sF^{10}$ metrics without hole, and the second best at $F^H$ without hole. On the York Urban dataset, our full model (HAWP + Our full approach) shows the best hole-robust performance at all metrics with hole. Excluding F-Clip(HR) and LETR from Table 5, our full model (HAWP + Our full approach) on the York Urban dataset presents the best numbers at all metrics regardless of with hole or without hole. Also, note that any of our full models consistently improves the ordinary detection performance without hole against the corresponding basic framework.

F-scores in Table 6 show that at least one of the models enabling our approach performs better than its basic model at all cases. It proves the effectiveness of applying at least one of our approaches, 3.2 (Conditional Training), 3.3 (RGB GAN) and 3.4 (Pseudo Labeling). Also, our model enabling all three approaches performs the best in general.

# 8. Extra Precision-Recall Curves

## 8.1. PR Curves for $sAP^{10}$ on Tests in Table 1 of the Main Paper

PR curves for $sAP^{10}$ on the tests with 0-10% hole in Table 1 of the Main Paper are shown in Figure 8. The red curve from our full model shows a significant hole-robust performance improvement for both test sets.

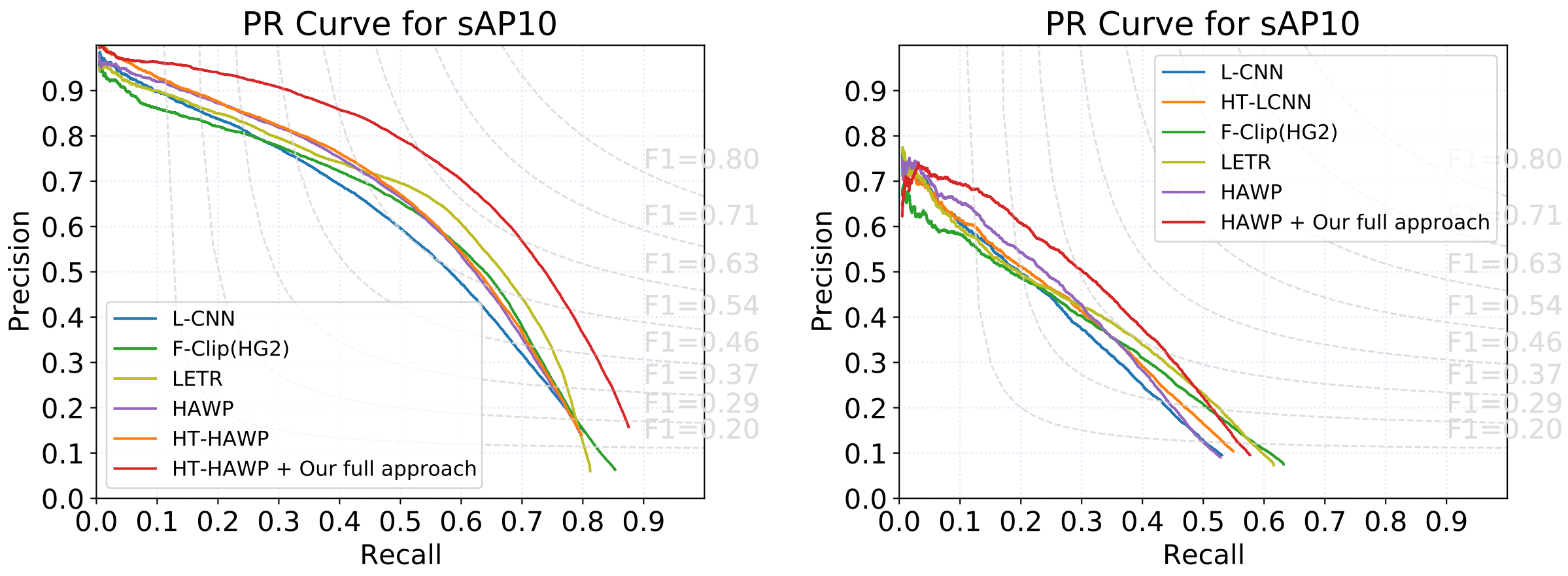

(a) Wireframe test set [5], **0-10% hole**

(b) York Urban [2], **0-10% hole**

Figure 8: PR curves for $sAP^{10}$ on the tests with *0-10%* hole in Table 1 of the Main Paper.

## 8.2. PR Curves for $AP^H$ on Tests in Table 1 of the Main Paper

Figure 9 and Figure 10 show Precision-Recall curves drawn for $AP^H$ on the tests *with* and *without* hole in Table 1 of the Main Paper.

The gap between our full model's curve and existing models' curves is visibly large at 10-30% hole in both test sets as shown in Figure 9 (a) and (c). The gap is still obvious in case of the Wireframe test set with 0-10% hole in Figure 9 (b). In case of the York Urban dataset with 0-10% hole, it is not easy to tell from Figure 9 (d), but the $AP^H$ number is better than all others compared in the graph (see Table 1 in the Main Paper), and also, its corresponding PR curves for $sAP^{10}$ in Figure 8 (b) show a clear superiority of our full model.

In the cases without hole in Figure 10, it is again hard to tell from the graphs intuitively, but the $AP^H$ number is better than all others compared in either of these graphs (see Table 1 in the Main Paper).

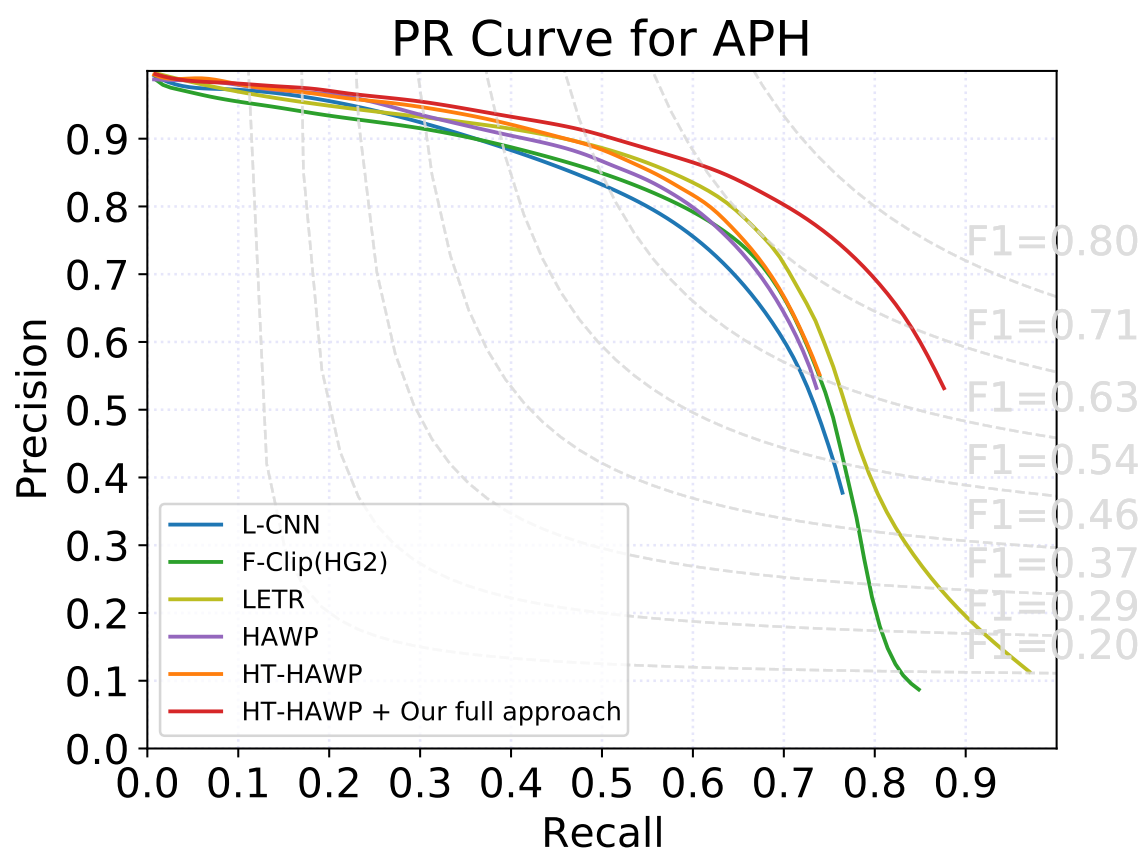

(a) Wireframe test set [5], **10-30% hole**

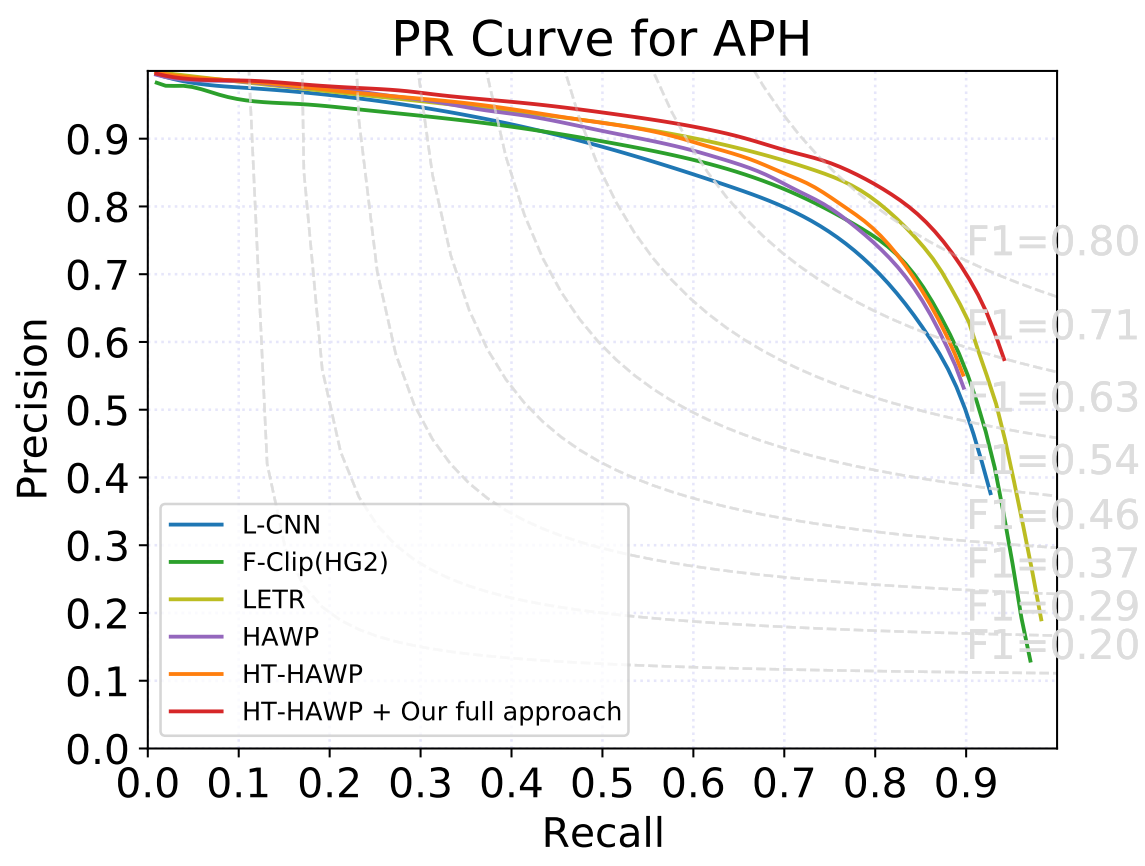

(b) Wireframe test set [5], **0-10% hole**

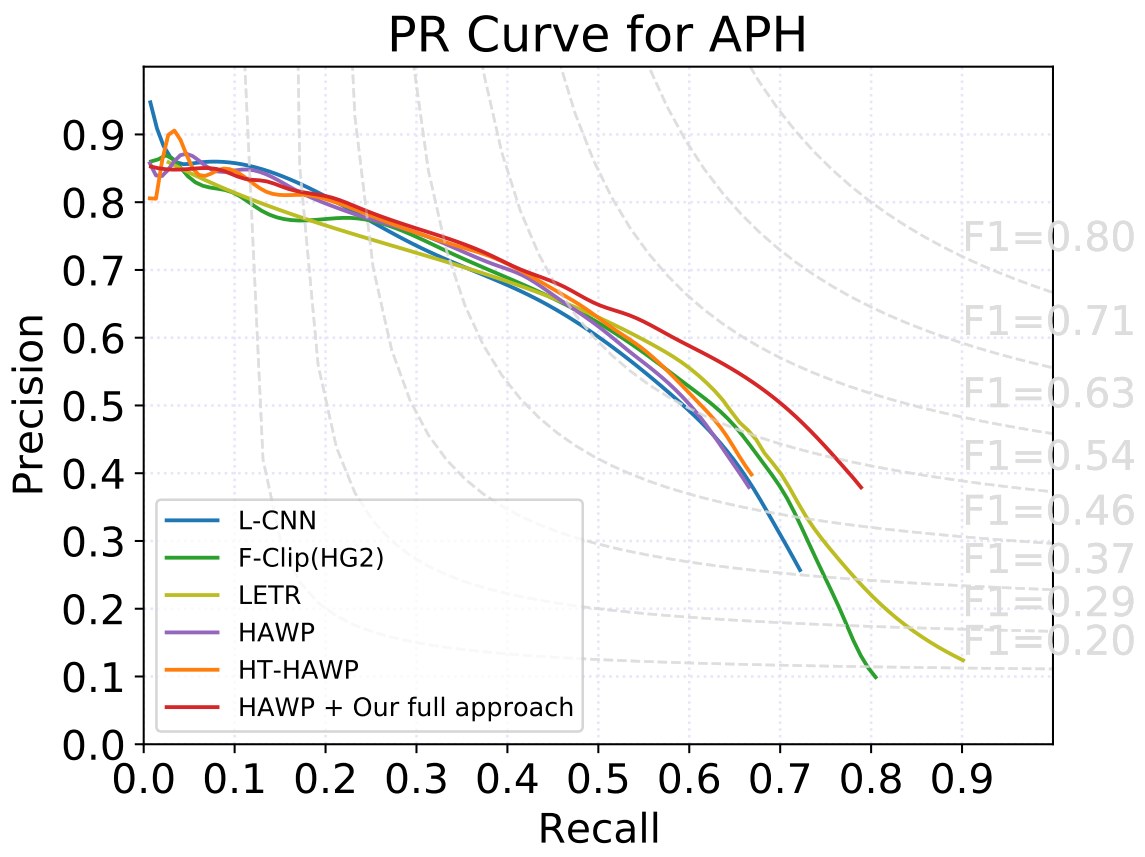

(c) York Urban [2], **10-30% hole**

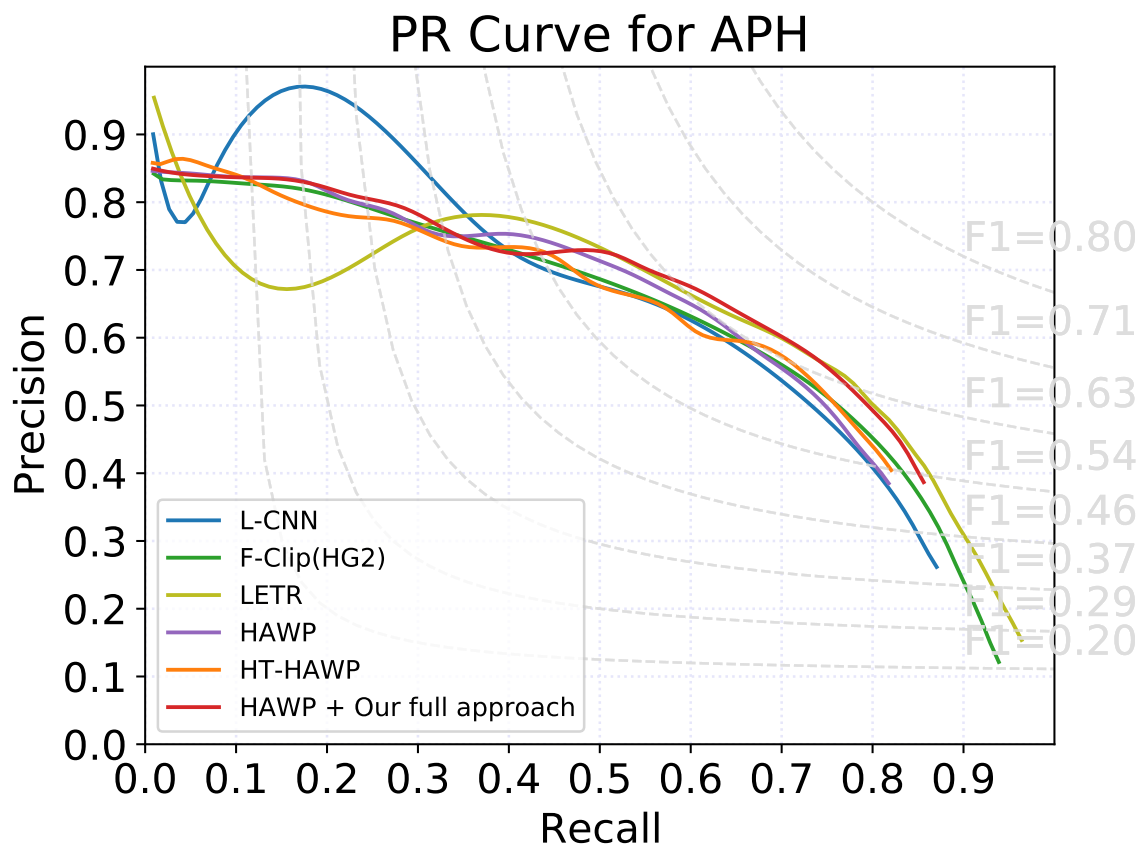

(d) York Urban [2], **0-10% hole**

Figure 9: PR curves for $AP^H$ on the tests *with* hole in Table 1 of the Main Paper.

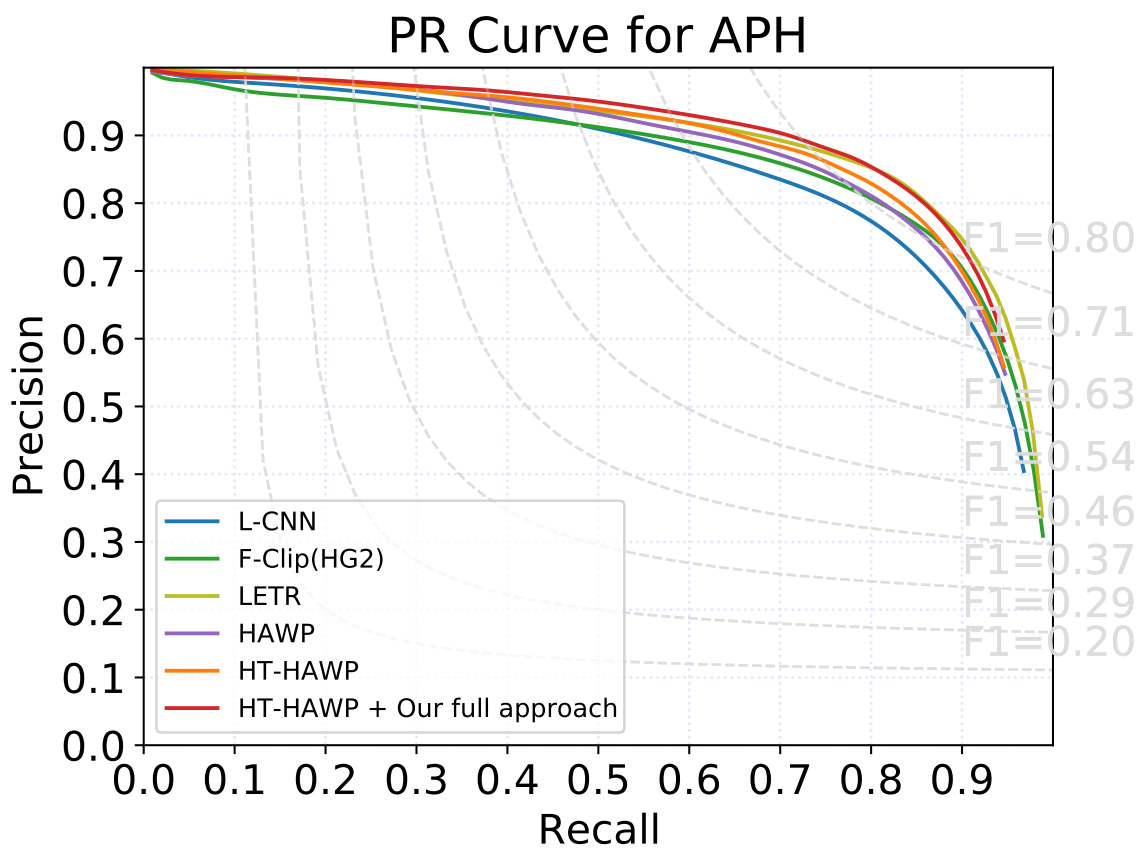

(a) Wireframe test set [5], ***without* hole**

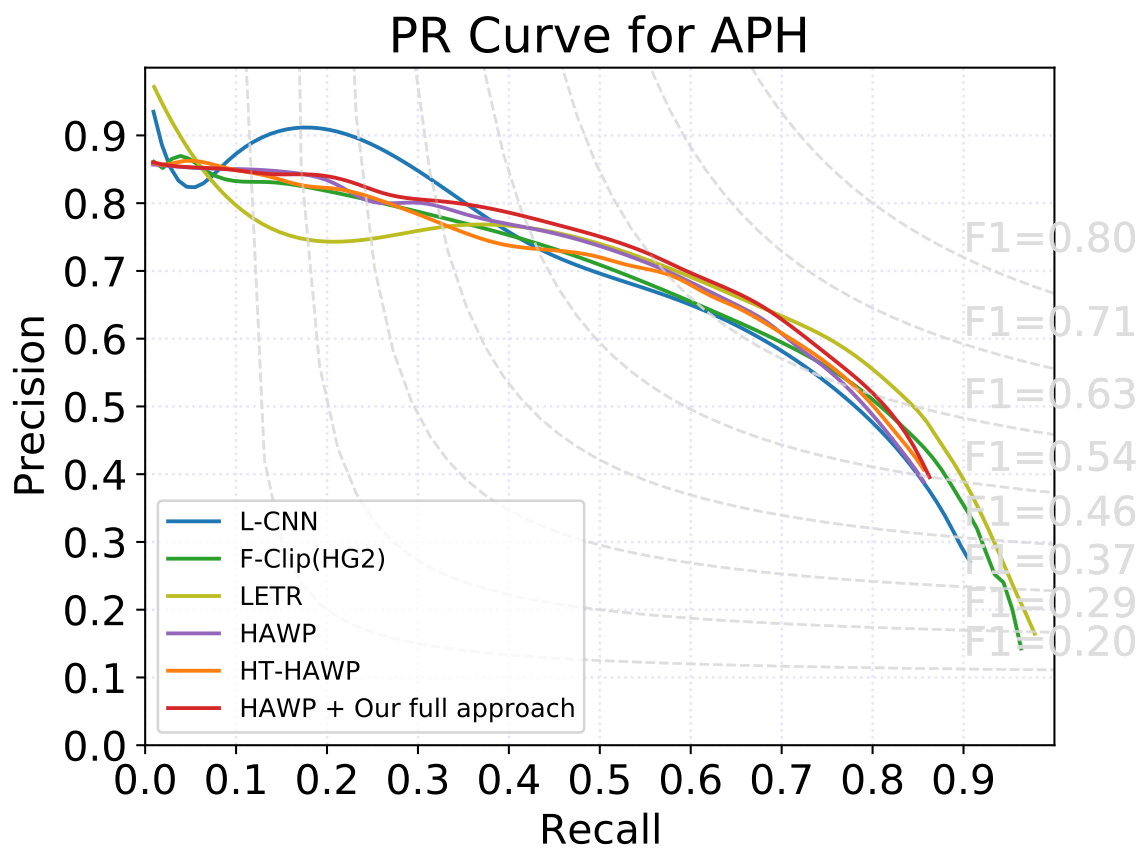

(b) York Urban [2], ***without* hole**

Figure 10: PR curves for $AP^H$ on the tests *without* hole in Table 1 of the Main Paper.

### 8.3. PR Curves for sAP$^{10}$ on Tests in Table 2 of the Main Paper

Figure 11 and Figure 12 show Precision-Recall curves drawn for the structural quality metric sAP$^{10}$ with respect to (a),(c),(e) and (f),(h),(j) cases tested in Table 2 of the Main Paper. Figure 11 proves that compared to baselines (black curves), our approach yet without pseudo labeling (red curves) is already significantly effective in improving hole-robustness, which is even further improved with pseudo labeling (green curves). Figure 12 proves that our approach helps slightly enhance ordinary detection as well, and with pseudo labeling, the gain becomes more distinct.

### 8.4. PR Curves for AP$^{H}$ on Tests in Table 2 of the Main Paper

Figure 13 and Figure 14 show Precision-Recall curves drawn for AP$^{H}$ with respect to (a),(c),(e) and (f),(h),(j) cases tested in Table 2 of the Main Paper. From these curves, we can again conclude as similarly as the PR-curves drawn for sAP$^{10}$ in Section 8.3.

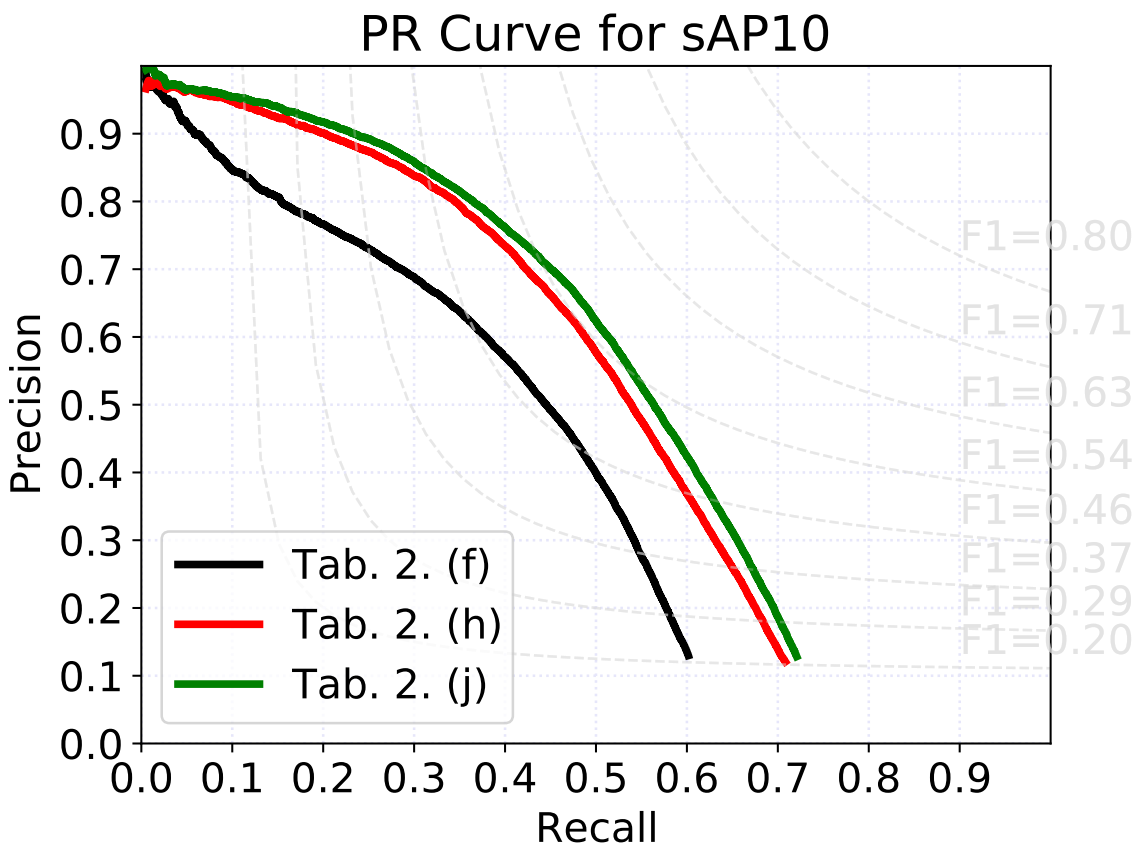

(a) Wireframe test set [5], **10-30% hole**

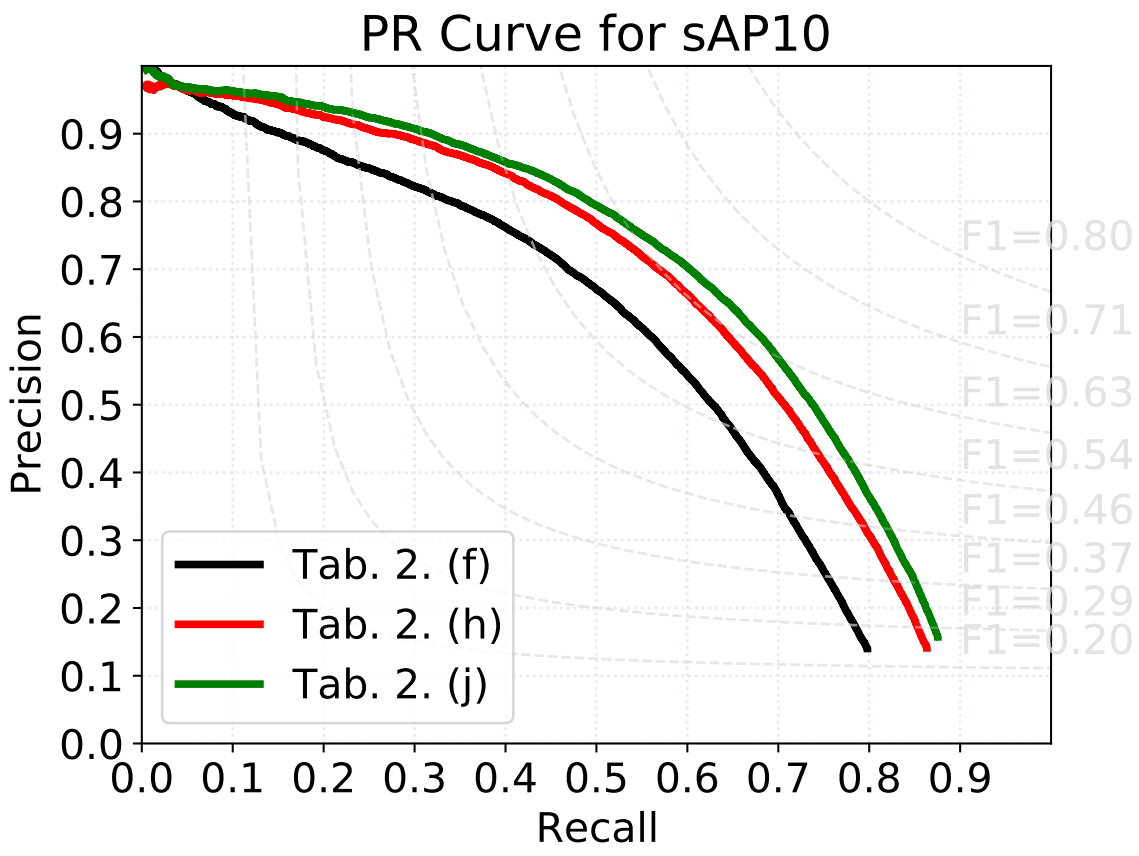

(b) Wireframe test set [5], **0-10% hole**

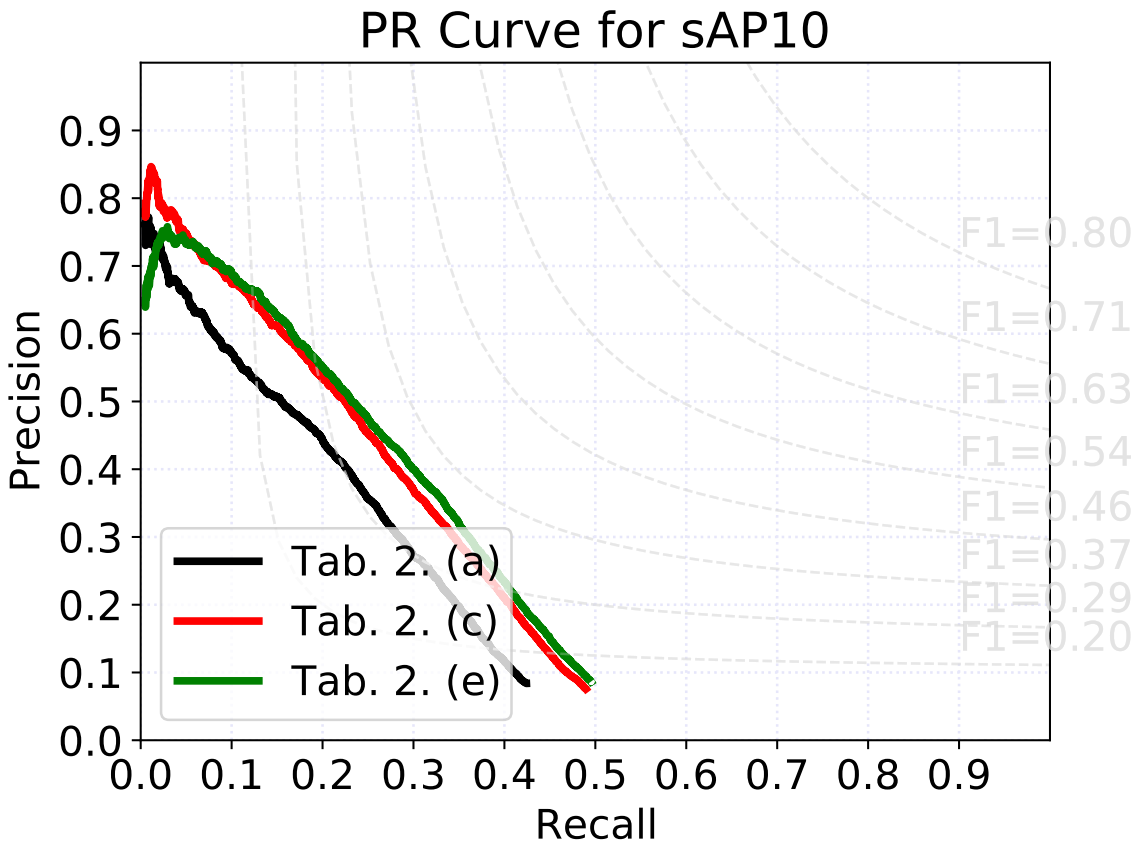

(c) York Urban [2], **10-30% hole**

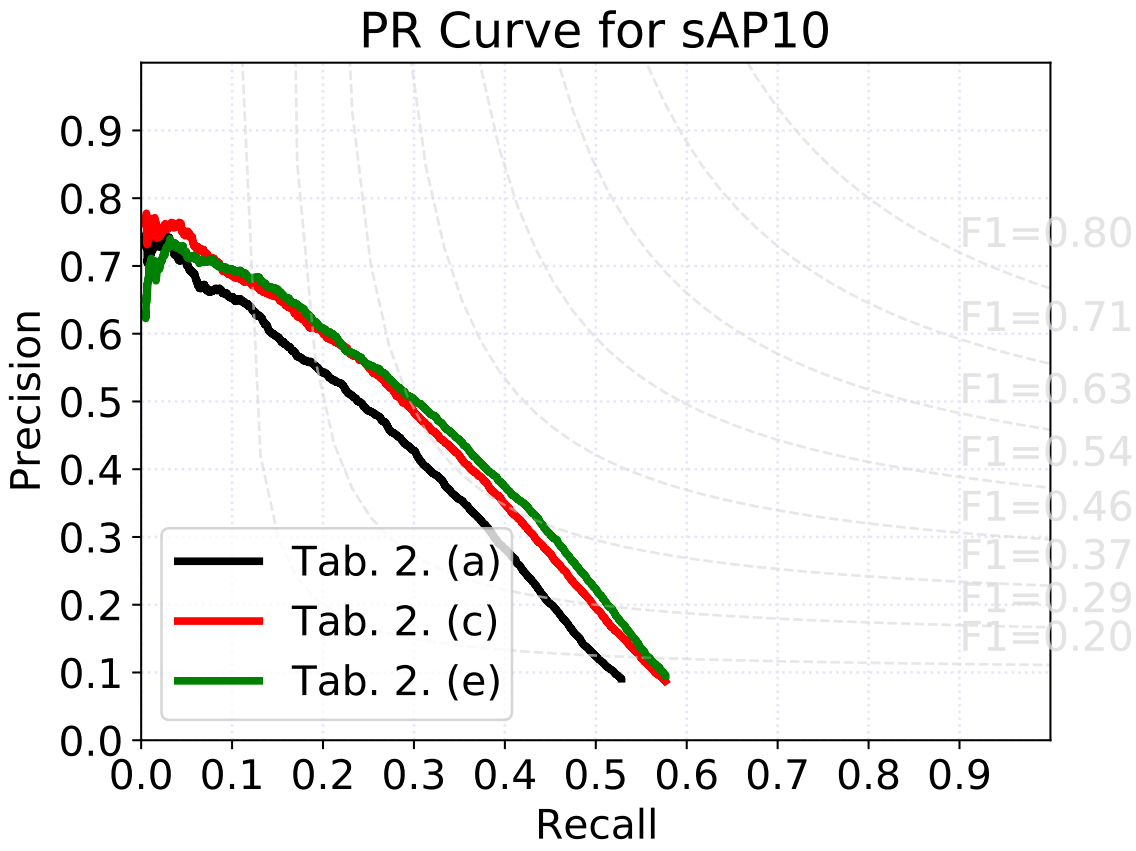

(d) York Urban [2], **0-10% hole**

Figure 11: PR curves for $sAP^{10}$ on the tests *with* hole in Table 2 of the Main Paper.

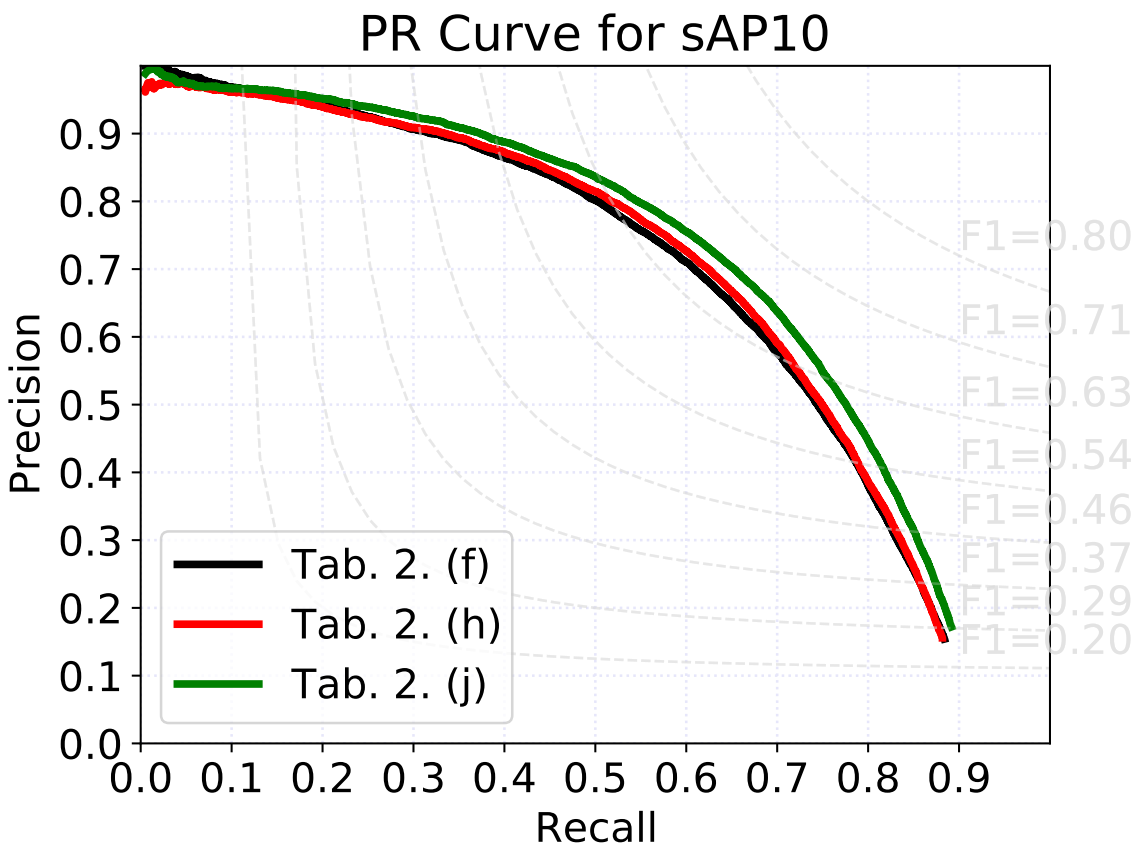

(a) Wireframe test set [5], ***without* hole**

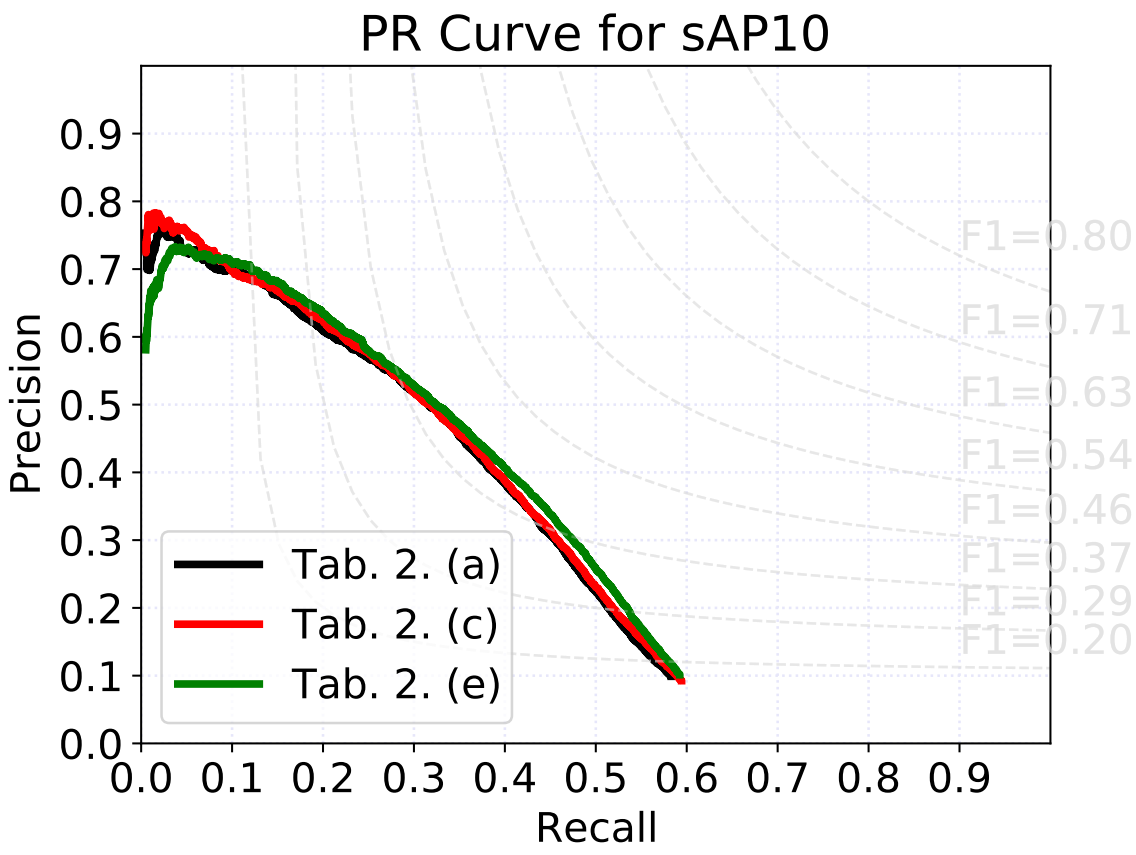

(b) York Urban [2], ***without* hole**

Figure 12: PR curves for $sAP^{10}$ on the tests *without* hole in Table 2 of the Main Paper.

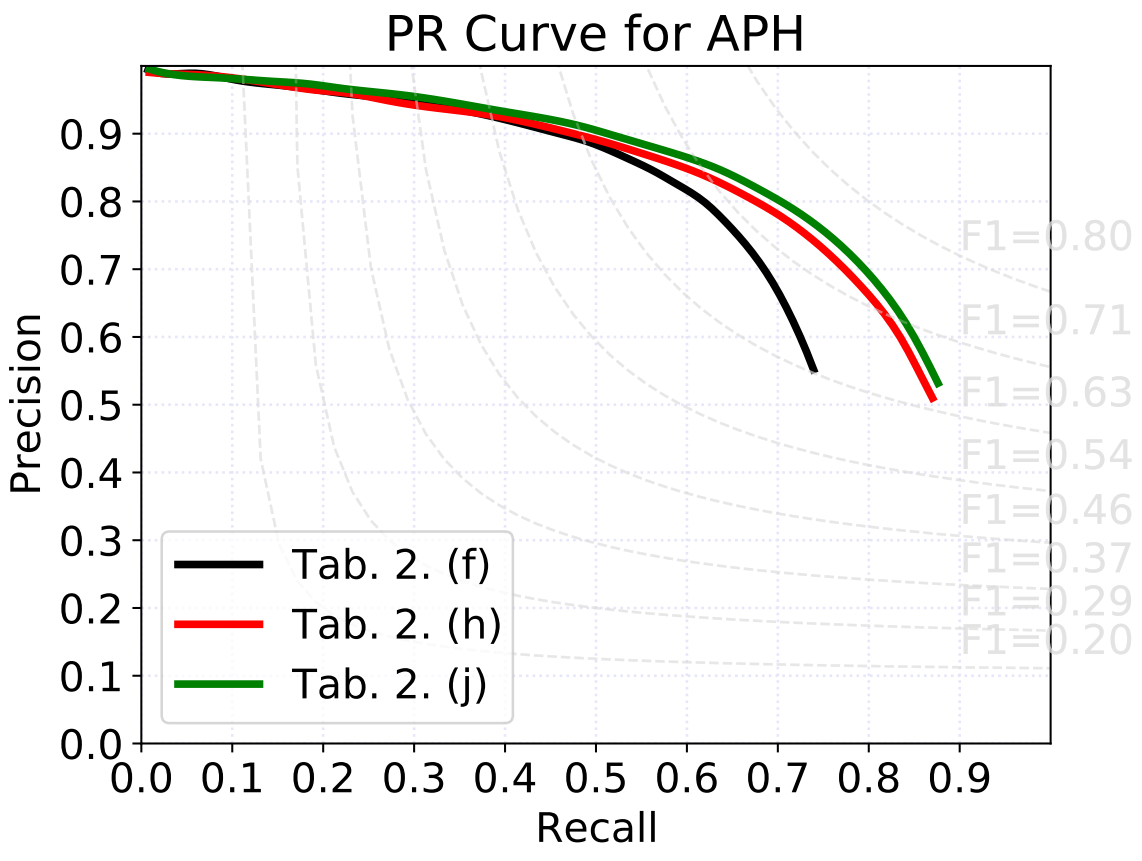

(a) Wireframe test set [5], **10-30% hole**

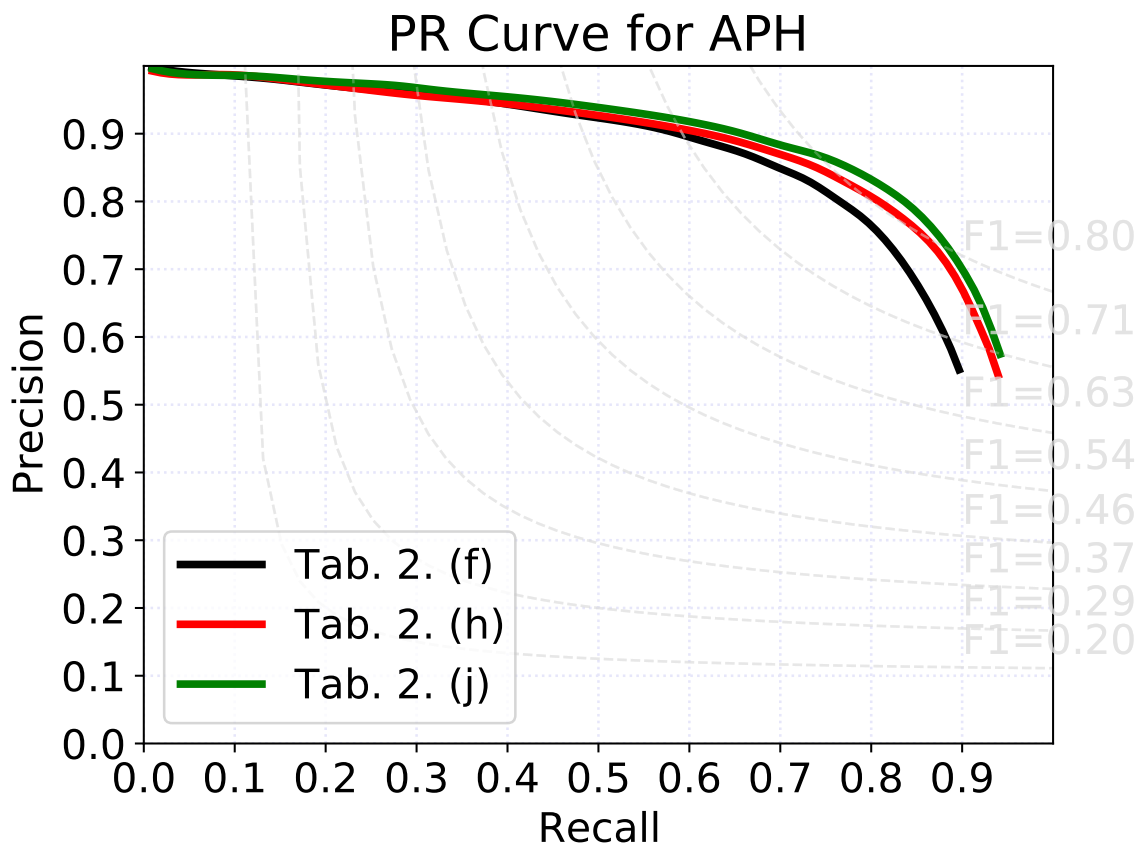

(b) Wireframe test set [5], **0-10% hole**

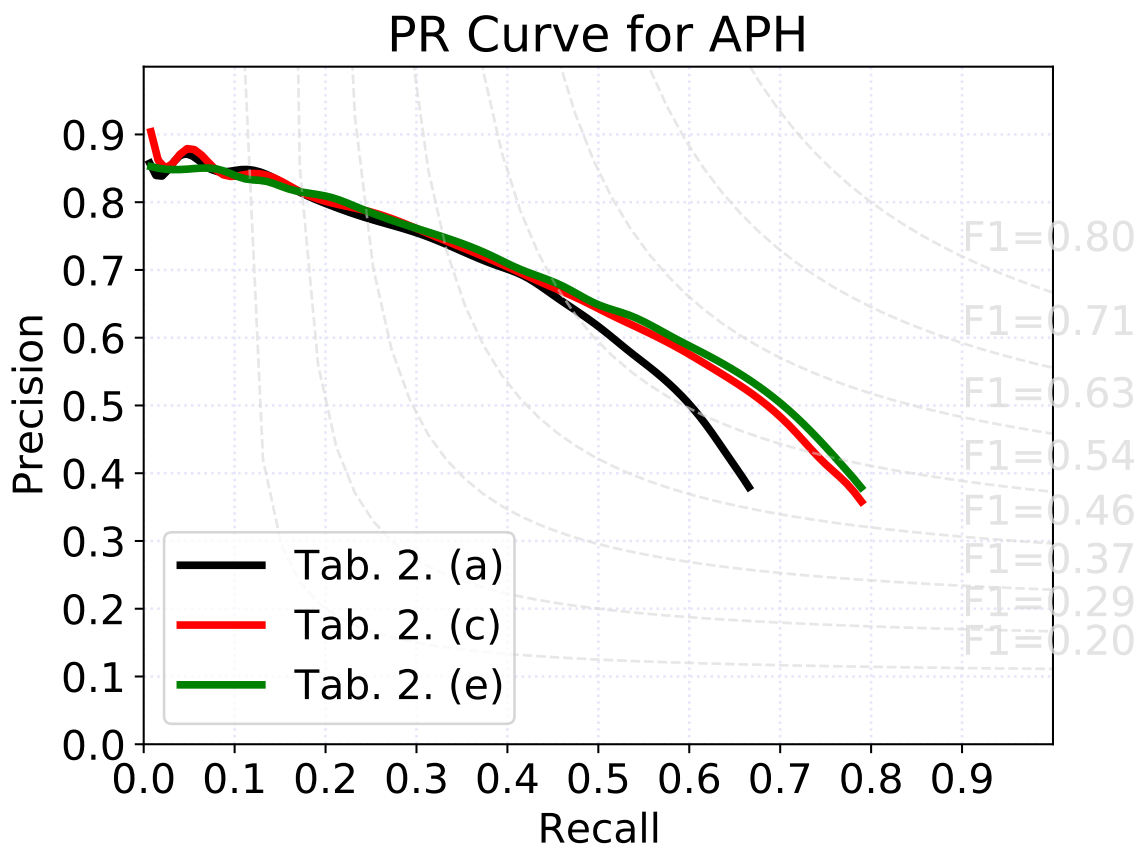

(c) York Urban [2], **10-30% hole**

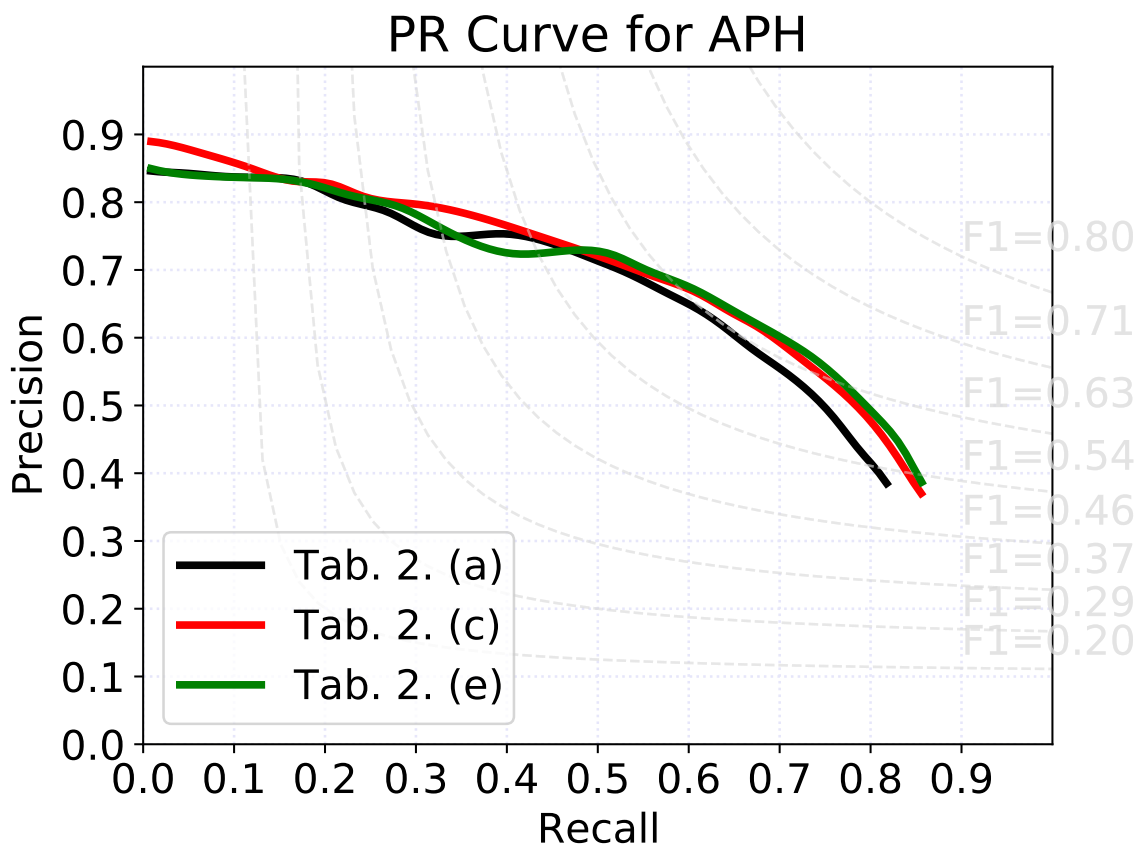

(d) York Urban [2], **0-10% hole**

Figure 13: PR curves for $AP^H$ on the tests *with* hole in Table 2 of the Main Paper.

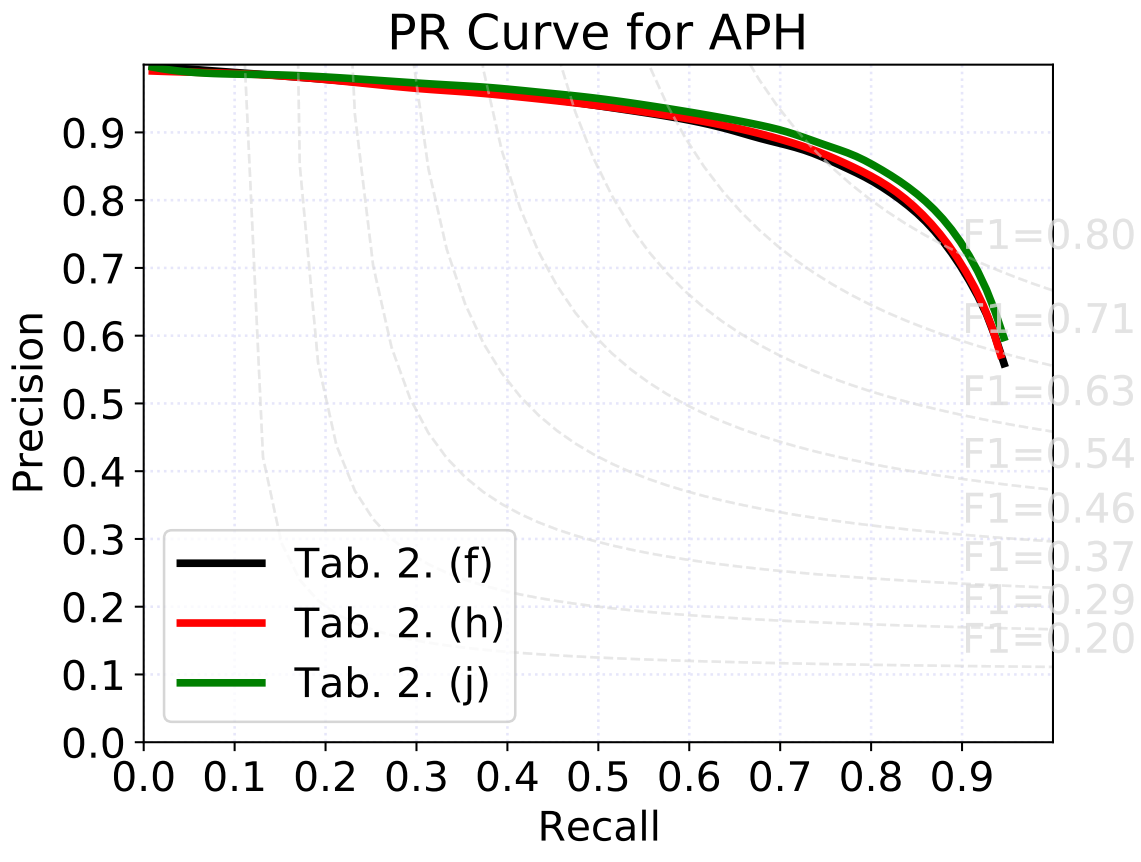

(a) Wireframe test set [5], ***without* hole**

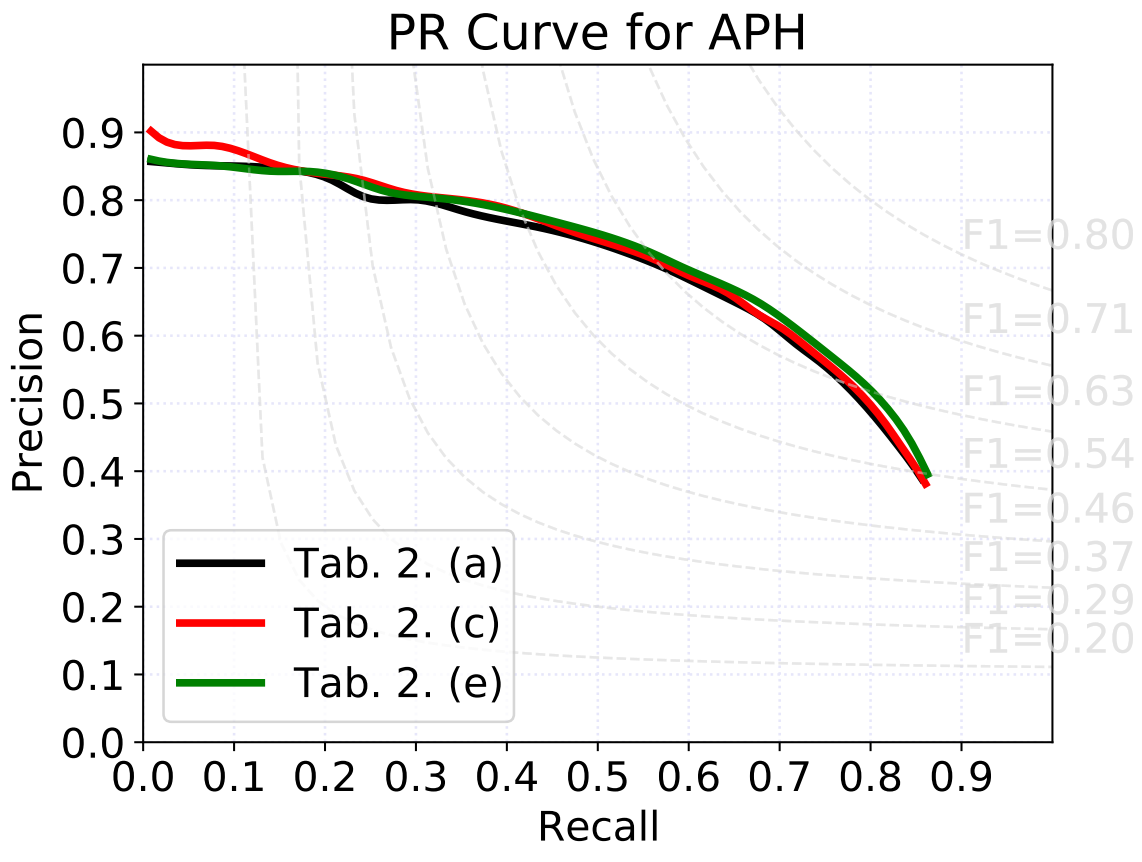

(b) York Urban [2], ***without* hole**

Figure 14: PR curves for $AP^H$ on the tests *without* hole in Table 2 of the Main Paper.

## 9. Initializing the Training with a Pretrained Ordinary Detection Model

| | Wireframe Test Set [5] | | | | | | | | | | | |
|---|---|---|---|---|---|---|---|---|---|---|---|---|
| | 10-30% Hole | | | | 0-10% Hole | | | | without Hole | | | |
| | sAP$^{5}$ | sAP$^{10}$ | mAP$^{J}$ | AP$^{H}$ | sAP$^{5}$ | sAP$^{10}$ | mAP$^{J}$ | AP$^{H}$ | sAP$^{5}$ | sAP$^{10}$ | mAP$^{J}$ | AP$^{H}$ |
| (a): HAWP + 3.2 + 3.3 | 44.17 | 48.11 | **47.1** | **73.1** | 57.99 | **62.35** | **58.2** | **82.9** | 62.60 | 66.60 | **60.5** | **85.1** |
| + Init w/ Pretrained | **44.45** | **48.26** | **47.1** | 72.5 | **58.17** | 62.34 | 58.1 | 82.6 | **62.89** | **66.69** | 60.4 | 84.8 |
| (b): HT-HAWP + 3.2 + 3.3 | 45.72 | 49.56 | **48.0** | **75.0** | 59.33 | 63.59 | **59.1** | **84.0** | 63.56 | 67.49 | **61.2** | **85.6** |
| + Init w/ Pretrained | **46.24** | **49.89** | **48.0** | 74.9 | **60.05** | **64.12** | **59.1** | 83.9 | **64.19** | **67.90** | **61.2** | **85.6** |
| | York Urban Dataset [2] | | | | | | | | | | | |
| | 10-30% Hole | | | | 0-10% Hole | | | | without Hole | | | |
| | sAP$^{5}$ | sAP$^{10}$ | mAP$^{J}$ | AP$^{H}$ | sAP$^{5}$ | sAP$^{10}$ | mAP$^{J}$ | AP$^{H}$ | sAP$^{5}$ | sAP$^{10}$ | mAP$^{J}$ | AP$^{H}$ |
| (c): HAWP + 3.2 + 3.3 | **20.30** | **22.34** | **25.8** | **53.0** | **24.90** | **27.35** | **31.2** | **60.6** | **26.69** | **29.08** | **32.4** | **62.3** |
| + Init w/ Pretrained | 18.47 | 20.32 | 24.7 | 51.2 | 23.18 | 25.34 | 29.3 | 58.7 | 24.62 | 26.76 | 30.5 | 60.4 |
| (d): HT-HAWP + 3.2 + 3.3 | 19.44 | 21.39 | 24.6 | 52.8 | **24.35** | **26.62** | 29.6 | **59.6** | **25.91** | **28.16** | 30.5 | **61.0** |
| + Init w/ Pretrained | **19.91** | **21.92** | **25.6** | **53.0** | 24.27 | 26.60 | **29.8** | 58.3 | 25.74 | 27.99 | **30.7** | 59.1 |

Table 7: Initializing the training with a pretrained ordinary detection model. For the tested models we combined our approach 3.2: *'Conditional Training'* and 3.3: *'RGB GAN'* on top of two basic frameworks HAWP [12] and HT-HAWP [12, 7]. Default training uses random initialization, which worked meaningfully better at (c). We see no clear difference at (a), (b) and (d).

We train a modified backbone network through our approach by initializing the kernel weights with random numbers as stated in [4] (i.e., *kaiming_uniform*). In this section, we test an effect of initializing our modified backbone with pretrained weights from the basic model made for ordinary detection without hole. Since the modified backbone network and the original backbone network are architecturally different in some parts (see Section 2), we copy the pretrained weights only for those matching layers while initializing the rest through default *kaiming_uniform*.

Table 7 shows the results on this experiment. On the Wireframe test set [5], initializing with the pretrained weights was slightly more helpful than full random initialization, but the difference was marginal. On the York Urban dataset [2], training with random initialization was much better in case of (c): HAWP + 3.2 + 3.3, while it is still unclear which initialization method is better in case of (d): HT-HAWP + 3.2 + 3.3. This suggests that initialization with pretrained weights for the baseline model does not have a clear benefit compared to random initialization, or could make it even worse. We suspect that the reason is, 1) our models are not sensitive to initialization methods, and 2) our modified backbone learns different information through the RGB GAN approach, which adds a generative role to the backbone network as part of a scene contents generator, unlike the baseline backbone dedicated only to learn ordinary detection through sole supervision.

## 10. Robustness to a Dim or Over Lit Condition

In this section we test the robustness of our models to a dim or over lit condition that we may rarely but do encounter while capturing a photo in real life. We simulated images captured under the dim or over lit condition by adjusting the original source images in the Wireframe Test Set [5] as follows:

1. Linearizing the camera response (assuming gamma 2.2)
2. Randomly scaling the image value (assuming the original image value is within $[0, 255]$)
   - for dim lit: scale by a random $s \in [\frac{1}{16.0}, \frac{1}{8.0}]$ and truncate below $0$.
   - for over lit: scale by a random $s \in [3.0, 3.3]$ and truncate above $255$.
3. (dim lit only) Adding Poisson noise (shot noise), with peak $\lambda = 8.0$
4. Reapplying the camera response (gamma 2.2)

| | (a) | (b) | (c) | Wireframe Test Set [5] 10-30% Hole sAP$^5$ | sAP$^{10}$ | mAP$^J$ | AP$^H$ | 0-10% Hole sAP$^5$ | sAP$^{10}$ | mAP$^J$ | AP$^H$ | without Hole sAP$^5$ | sAP$^{10}$ | mAP$^J$ | AP$^H$ |
|---|---|---|---|---|---|---|---|---|---|---|---|---|---|---|---|
| Baseline HAWP | ✓ | | | 34.80 | 37.74 | 40.8 | 64.5 | 50.80 | 54.84 | 54.0 | 79.4 | 62.52 | 66.49 | 60.2 | 85.0 |
| | | ✓ | | 29.89 | 33.02 | 35.7 | 59.4 | 44.05 | 48.25 | 47.8 | 74.1 | 54.74 | 59.08 | 54.1 | 79.9 |
| | | | ✓ | 27.96 | 30.78 | 34.0 | 55.5 | 41.15 | 45.05 | 45.0 | 69.0 | 50.80 | 54.82 | 50.4 | 74.1 |
| HAWP + Ours full | ✓ | | | **47.59** | **51.50** | **49.6** | **74.9** | **61.83** | **65.98** | **61.0** | **85.0** | **65.98** | **69.76** | **63.0** | **86.9** |
| | | ✓ | | 41.19 | 45.23 | 44.4 | 69.9 | 54.09 | 58.53 | 54.9 | 80.2 | 58.32 | 62.47 | 57.0 | 82.1 |
| | | | ✓ | 38.43 | 42.14 | 41.0 | 64.5 | 50.14 | 54.30 | 50.5 | 73.5 | 53.74 | 57.59 | 52.3 | 75.5 |
| Baseline HT-HAWP | ✓ | | | 35.64 | 38.54 | 41.8 | 65.5 | 51.58 | 55.50 | 55.1 | 80.1 | 63.26 | 67.12 | 61.3 | 85.7 |
| | | ✓ | | 29.80 | 32.85 | 35.6 | 59.6 | 43.76 | 48.02 | 47.8 | 74.4 | 54.64 | 59.01 | 54.3 | 80.2 |
| | | | ✓ | 28.45 | 31.18 | 34.5 | 56.2 | 41.43 | 45.16 | 45.7 | 69.8 | 51.03 | 54.81 | 51.2 | 74.6 |
| HT-HAWP + Ours full | ✓ | | | **48.02** | **51.82** | **49.9** | **76.6** | **62.21** | **66.31** | **61.3** | **85.5** | **66.19** | **69.92** | **63.2** | **87.0** |
| | | ✓ | | 41.52 | 45.51 | 44.3 | 71.3 | 54.24 | 58.92 | 54.7 | 80.4 | 58.33 | 62.60 | 56.8 | 82.1 |
| | | | ✓ | 38.85 | 42.40 | 41.3 | 66.1 | 50.52 | 54.59 | 50.9 | 74.3 | 53.95 | 57.75 | 52.6 | 75.8 |

(a): normal lit
(b): dim lit
(c): over lit

Table 8: Testing robustness to dim or over lit conditions. We compared the basic frameworks HAWP [12] and HT-HAWP [12, 7] with those modified by applying our full approach.

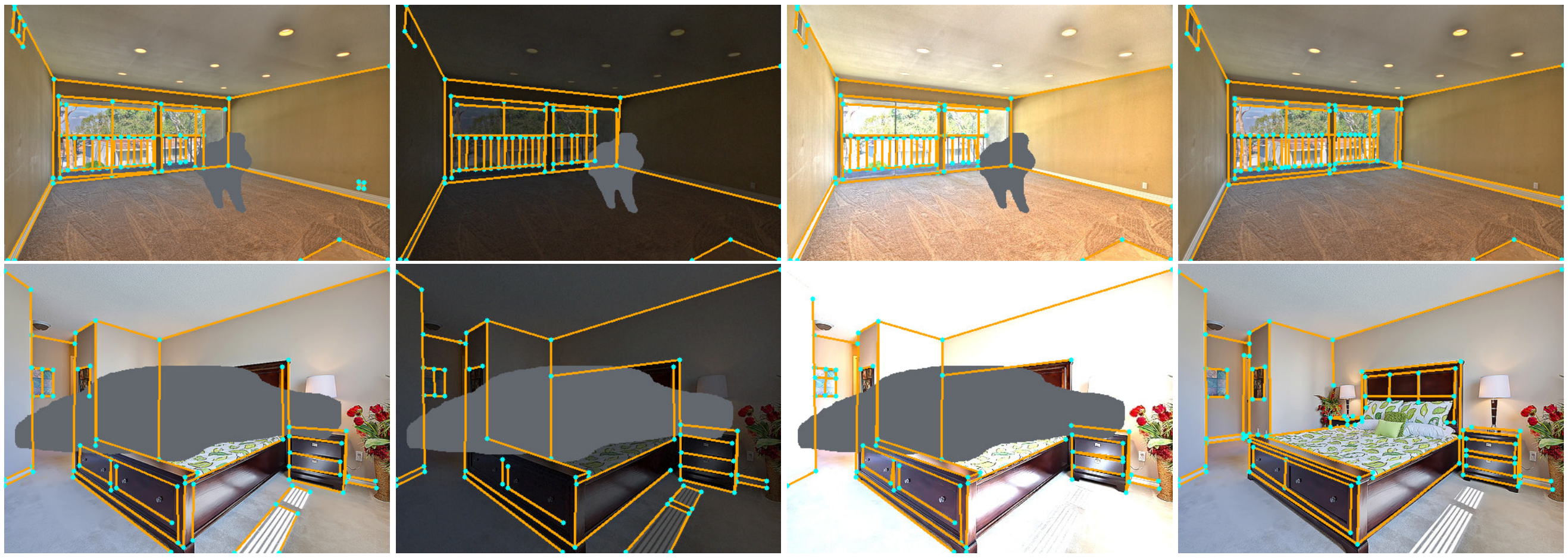

(a) Normal lit (b) Dim lit (c) Over lit (d) Ground Truth

Figure 15: Results on dim and over lit conditions. HT-HAWP [12, 7] with our full approach was applied to detect wireframes visualized in (a), (b) and (c).

Table 8 show results on this experiment. We compared the basic frameworks HAWP [12] and HT-HAWP [12, 7] with those modified by applying our full approach. We found that the performance drop is more prominent in the over lit case than the dim lit case, since the over lit scenario introduced stronger and larger saturated (white) regions than the dim lit one. Here, a model with our full approach always performed far better than its corresponding basic model on the same input condition, either without or with hole. Especially, with 10-30% hole, the model with our full approach applied to challenging *over lit* images performed even better than the basic model applied to *normal* images. We suspect that our large scale training on the pseudo-labeled data derived from diverse images may have helped improve the robustness of the model on unnatural lighting conditions. Figure 15 visualizes simulated images under a dim or over lit condition and detection results on them.

## 11. Testing Ordinary Detection Models on Inpainted Images

| | Wireframe Test Set [5] | | | | | | | | York Urban Dataset [2] | | | | | | | |
|---|---|---|---|---|---|---|---|---|---|---|---|---|---|---|---|---|
| | 10-30% Hole | | | | 0-10% Hole | | | | 10-30% Hole | | | | 0-10% Hole | | | |
| | $sAP^5$ | $sAP^{10}$ | $mAP^J$ | $AP^H$ | $sAP^5$ | $sAP^{10}$ | $mAP^J$ | $AP^H$ | $sAP^5$ | $sAP^{10}$ | $mAP^J$ | $AP^H$ | $sAP^5$ | $sAP^{10}$ | $mAP^J$ | $AP^H$ |
| (Detecting on images with hole) | | | | | | | | | | | | | | | | |
| L-CNN [15] | 32.83 | 35.78 | 40.4 | 61.3 | 47.73 | 51.68 | 53.2 | 75.3 | 14.65 | 16.16 | 21.9 | 44.6 | 20.23 | 22.15 | 27.5 | 54.6 |
| HT-LCNN [15, 7] | 33.56 | 36.48 | 41.4 | 57.1 | 48.89 | 52.79 | 54.2 | 76.6 | 15.84 | 17.48 | 23.5 | 40.4 | 21.47 | 23.62 | 29.3 | 50.1 |
| F-Clip(HG2) [1] | 33.51 | 36.68 | / | 65.3 | 49.46 | 53.87 | / | 79.5 | 15.86 | 17.54 | / | 48.6 | 22.02 | 24.07 | / | 58.6 |
| HAWP [12] | 34.80 | 37.74 | 40.8 | 64.5 | 50.80 | 54.84 | 54.0 | 79.4 | 15.79 | 17.48 | 22.5 | 46.3 | 21.61 | 23.79 | 28.6 | 57.3 |
| HT-HAWP [12, 7] | 35.64 | 38.54 | 41.8 | 65.5 | 51.58 | 55.50 | 55.1 | 80.1 | 15.57 | 17.18 | 23.2 | 46.8 | 21.45 | 23.63 | 29.0 | 56.6 |
| (Detecting on **inpainted** images by applying [13]) | | | | | | | | | | | | | | | | |
| L-CNN [15] | 39.01 | 42.84 | 45.0 | 70.5 | 53.30 | 57.70 | 56.4 | 78.4 | 17.02 | 19.07 | 24.0 | 50.9 | 22.80 | 24.92 | 29.1 | 56.8 |
| HT-LCNN [15, 7] | 40.32 | 44.01 | 46.3 | 71.1 | 54.70 | 59.04 | 57.6 | 79.5 | 18.27 | 20.43 | 25.4 | 46.8 | 24.00 | 26.36 | 31.0 | 51.9 |
| F-Clip(HG2) [1] | 40.87 | 45.04 | / | 75.2 | 55.50 | 60.39 | / | 82.6 | 18.81 | 21.07 | / | **56.9** | 25.05 | 27.43 | / | **60.7** |
| HAWP [12] | 41.82 | 45.67 | 45.7 | 73.7 | 56.61 | 61.10 | 57.1 | 82.8 | 18.53 | 20.61 | 24.5 | 53.7 | 24.32 | 26.70 | 30.2 | 60.1 |
| HT-HAWP [12, 7] | 42.52 | 46.32 | 46.7 | 74.3 | 57.36 | 61.75 | 58.4 | 83.5 | 18.01 | 20.19 | 25.4 | 54.8 | 23.60 | 26.00 | 30.6 | 58.7 |
| (Detecting on images with hole) | | | | | | | | | | | | | | | | |
| HAWP + Our full approach | 47.59 | 51.50 | 49.6 | 74.9 | 61.83 | 65.98 | 61.0 | 85.0 | **20.81** | **23.01** | **27.0** | 53.4 | **25.62** | **28.01** | **31.7** | 60.2 |
| HT-HAWP + Our full approach | **48.02** | **51.82** | **49.9** | **76.6** | **62.21** | **66.31** | **61.3** | **85.5** | 19.96 | 21.92 | 25.4 | 52.9 | 24.72 | 26.94 | 29.8 | 57.6 |

Table 9: Testing existing works on images whose holes had been inpainted by applying a popular method Deepfillv2 [13]. (Note that we excluded F-Clip(HR) [1] and LETR [11] from the comparison as they are fundamentally very different from the other frameworks thus hard to be compared in parallel.)

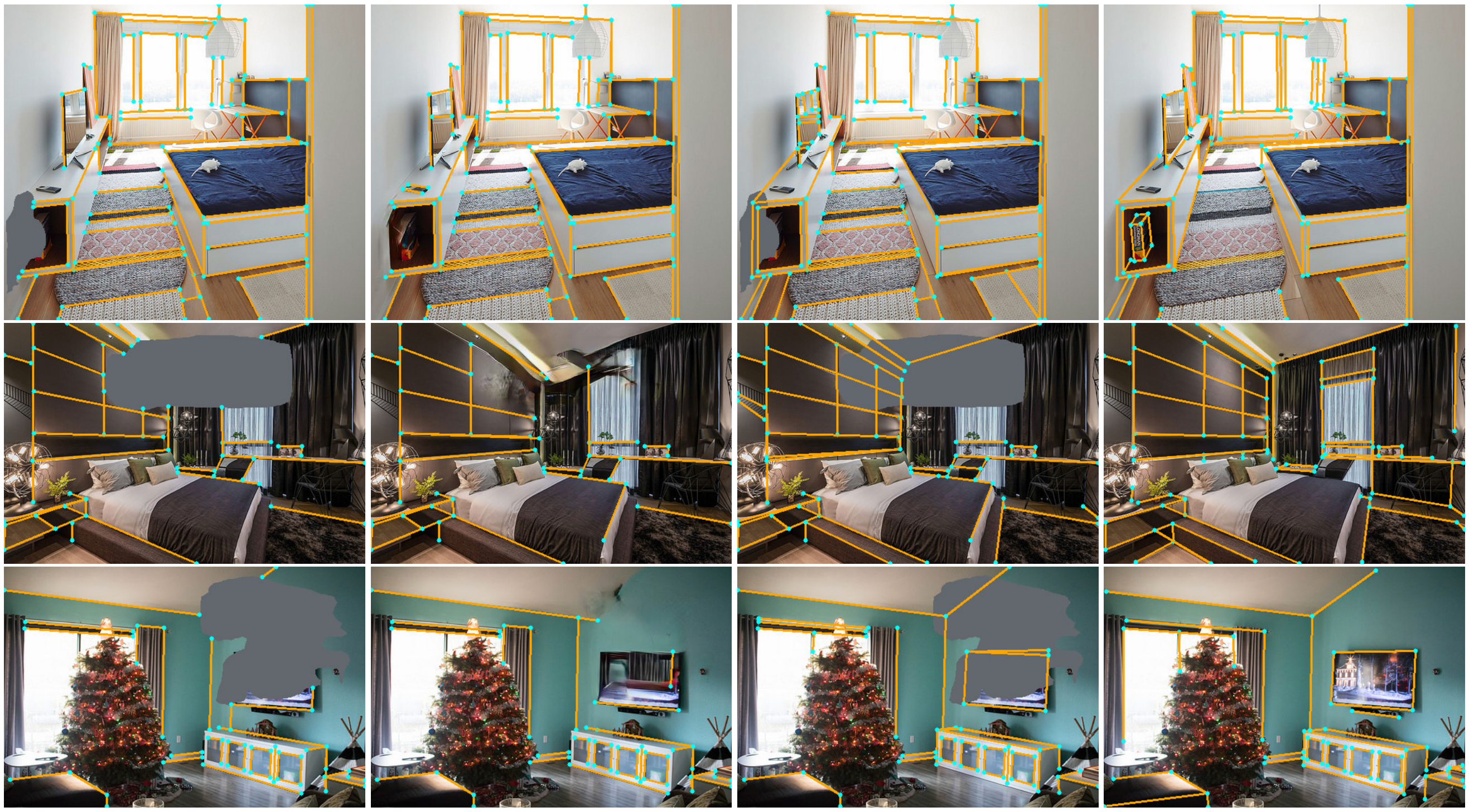

(a) HT-HAWP with hole (b) HT-HAWP on inpainted input (c) Ours with hole (d) Ground Truth

Figure 16: Applying an ordinary detection model, HT-HAWP [12, 7], to images inpainted by using Deepfillv2 [13]. Artifacts in the inpainted regions may still prevent proper inference of line segments as shown in (b). Ours: our full approach using HT-HAWP as the basic framework. Hole size: 0-10% in the first row, 10-30% in the second and third rows.

In Table 9 we tested existing ordinary wireframe detection models on images whose holes had been inpainted by applying a popular inpainting method Deepfillv2 [13]. This does help improve the performance as shown in the second group of the table. However, the inpainting result may not be perfect inside large holes such as those at 10-30%, which suffers from crooked structure, disturbing artifacts or blurry completion. This abnormality may prevent proper inference of line segments across the inpainted regions as shown in Figure 16 (b). We found that in general, our full approach using HAWP or our full approach using HT-HAWP applied to non-inpainted images with hole still performs better than those existing models applied to inpainted images.

## 12. More Qualitative Results

In this section, we show a more number of qualitative results and real-world examples that could not be put in the Main Paper due to space.

### 12.1. Results on Standard Test Sets with and without Hole

We show extra results on the Wireframe test set [5] *with* 10-30% hole in Figure 17, and *with* 0-10% hole in Figure 18. In Figure 19, we show extra results on York Urban dataset [2] *with* 10-30% and 0-10% hole. We show extra results on both test sets *without* hole in Figure 20.

These results are also from the same methods compared in Figures 6 and 7 of the Main Paper.

### 12.2. Real-world Examples with Occlusion by Foreground Objects

In Figures 21, 22, 23 and 24, we demonstrate additional qualitative results on real-world examples with obvious occlusion by foreground objects. The example images were sampled and cropped (512×512) from the Places365 [14] Validation set. We apply existing works HT-LCNN [15, 7], HT-HAWP [12, 7], and F-Clip(HR) [1] to an input image as is. We use Detectron2 [10] panoptic segmentation (‘*things*’ only) to indicate the occluded region by a foreground object in the image, and apply our hole-robust detection full model to it.

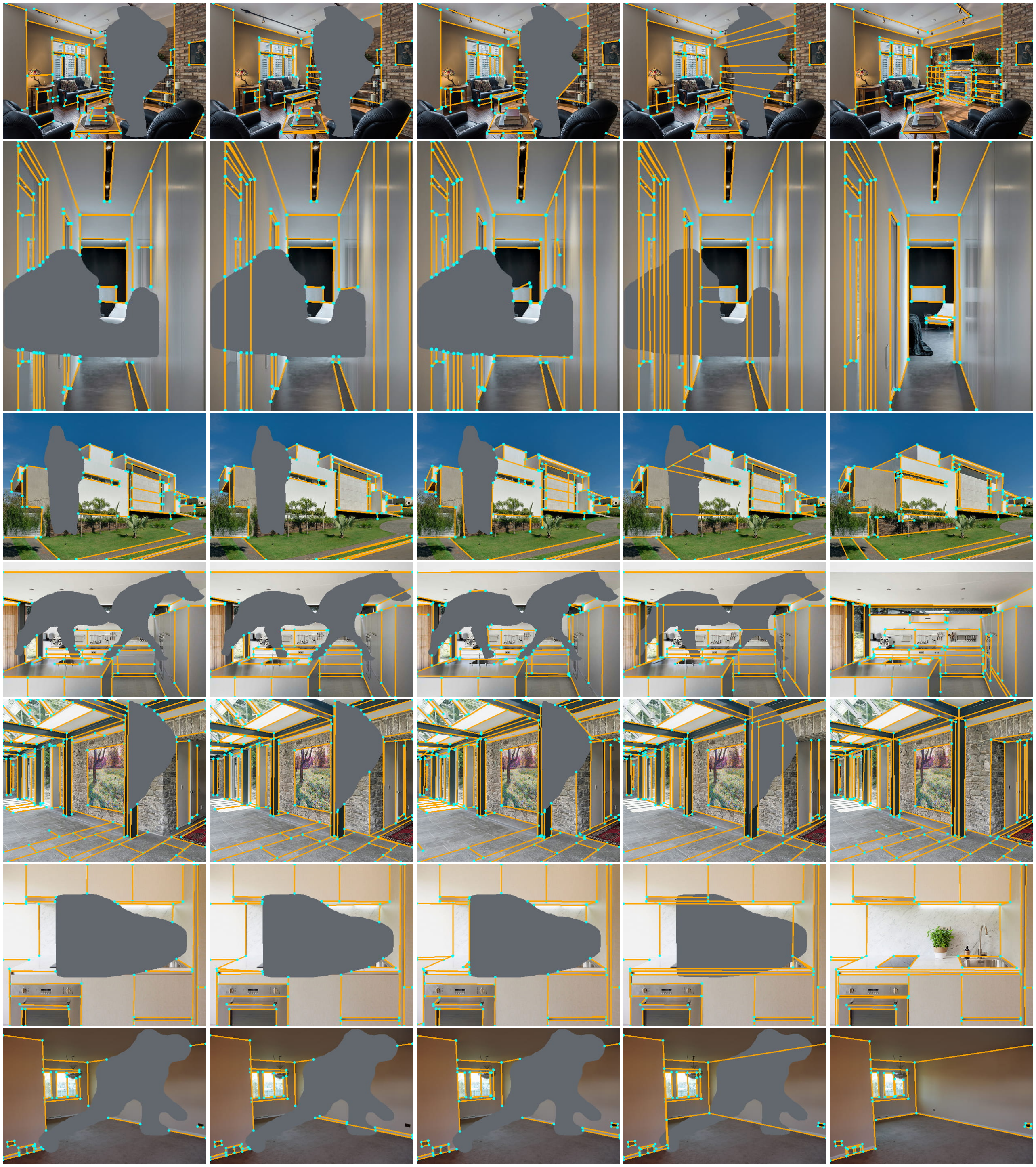

(a) HT-LCNN [15, 7] (b) HT-HAWP [12, 7] (c) F-Clip(HR) [1] (d) Our full approach using HT-HAWP [12, 7] (e) Ground Truth

Figure 17: Results on the **Wireframe test set** [5] ***with*** **10-30% hole.**

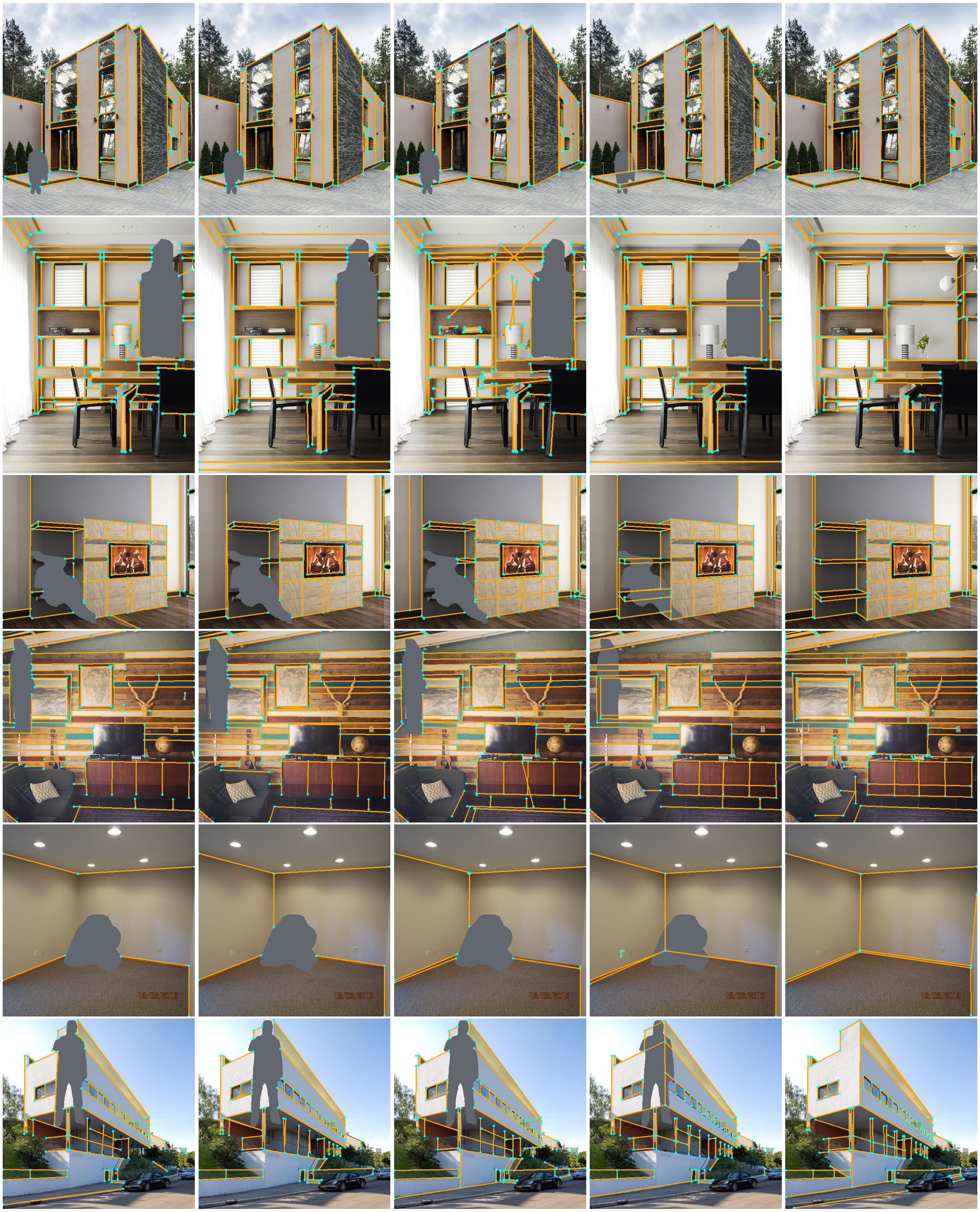

(a) HT-LCNN [15, 7] (b) HT-HAWP [12, 7] (c) F-Clip(HR) [1] (d) Our full approach using HT-HAWP [12, 7] (e) Ground Truth

Figure 18: Results on the **Wireframe test set** [5] ***with* 0-10% hole.**

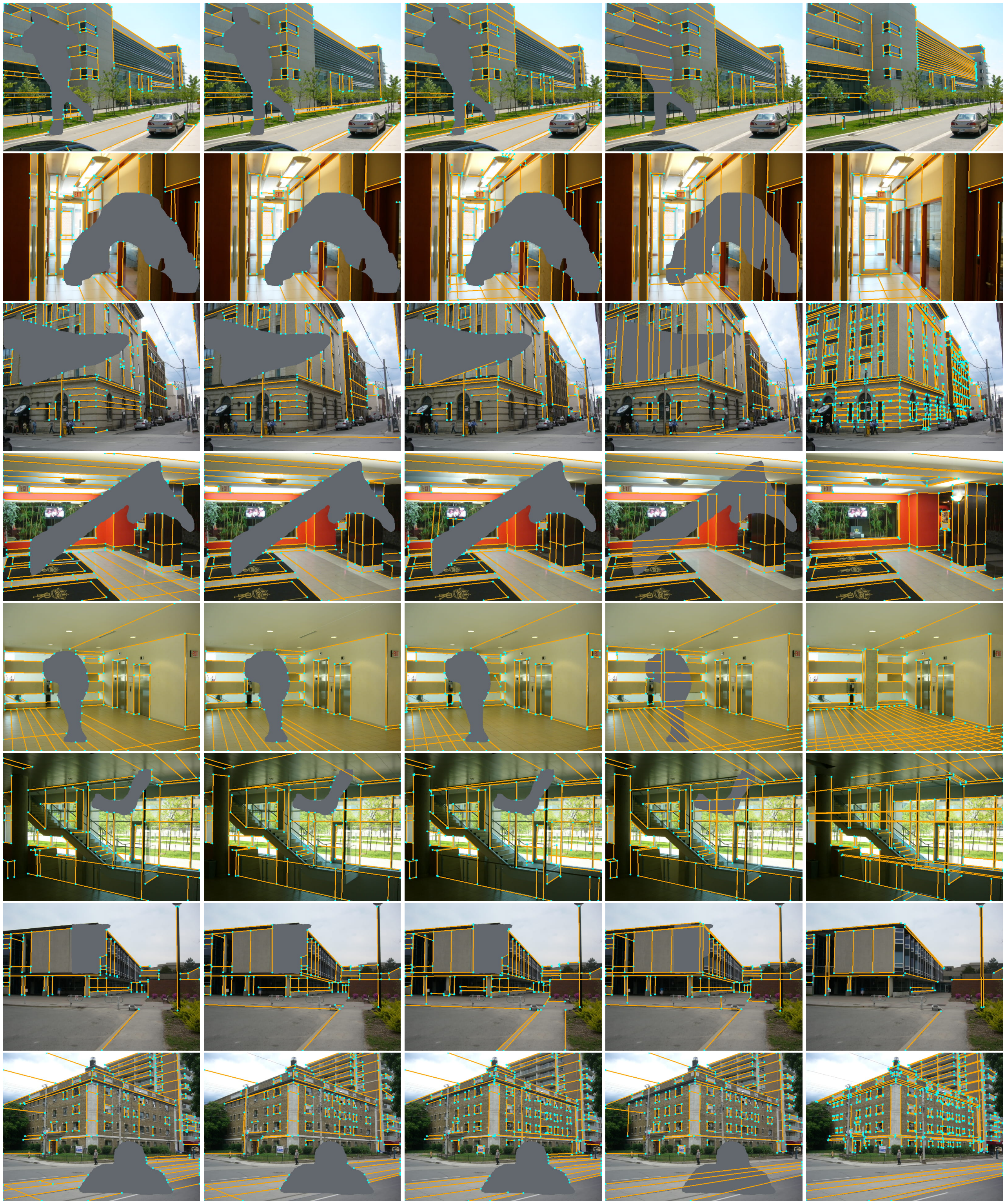

(a) HT-LCNN [15, 7] (b) HT-HAWP [12, 7] (c) F-Clip(HR) [1] (d) Our full approach using HT-HAWP [12, 7] (e) Ground Truth

Figure 19: Results on the **York Urban dataset** [2] ***with* hole.** (**10-30%** from 1st to 4th rows, **0-10%** from 5th to 8th rows)

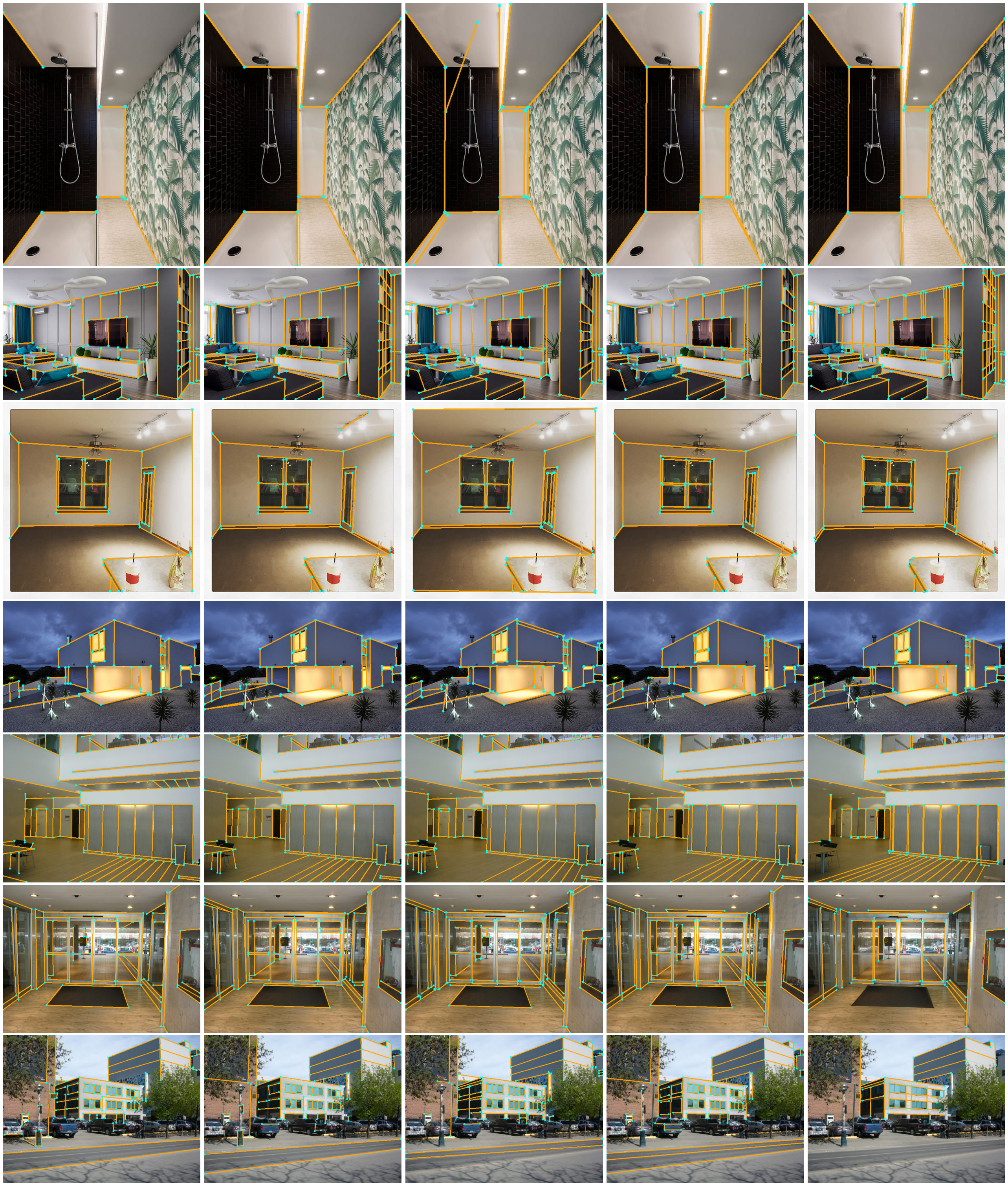

(a) HT-LCNN [15, 7] (b) HT-HAWP [12, 7] (c) F-Clip(HR) [1] (d) Our full approach using HT-HAWP [12, 7] (e) Ground Truth

Figure 20: Results on the **Wireframe test set** [5] (1st to 4th rows) and **York Urban dataset** [2] (5th to 7th rows) ***without* hole.**

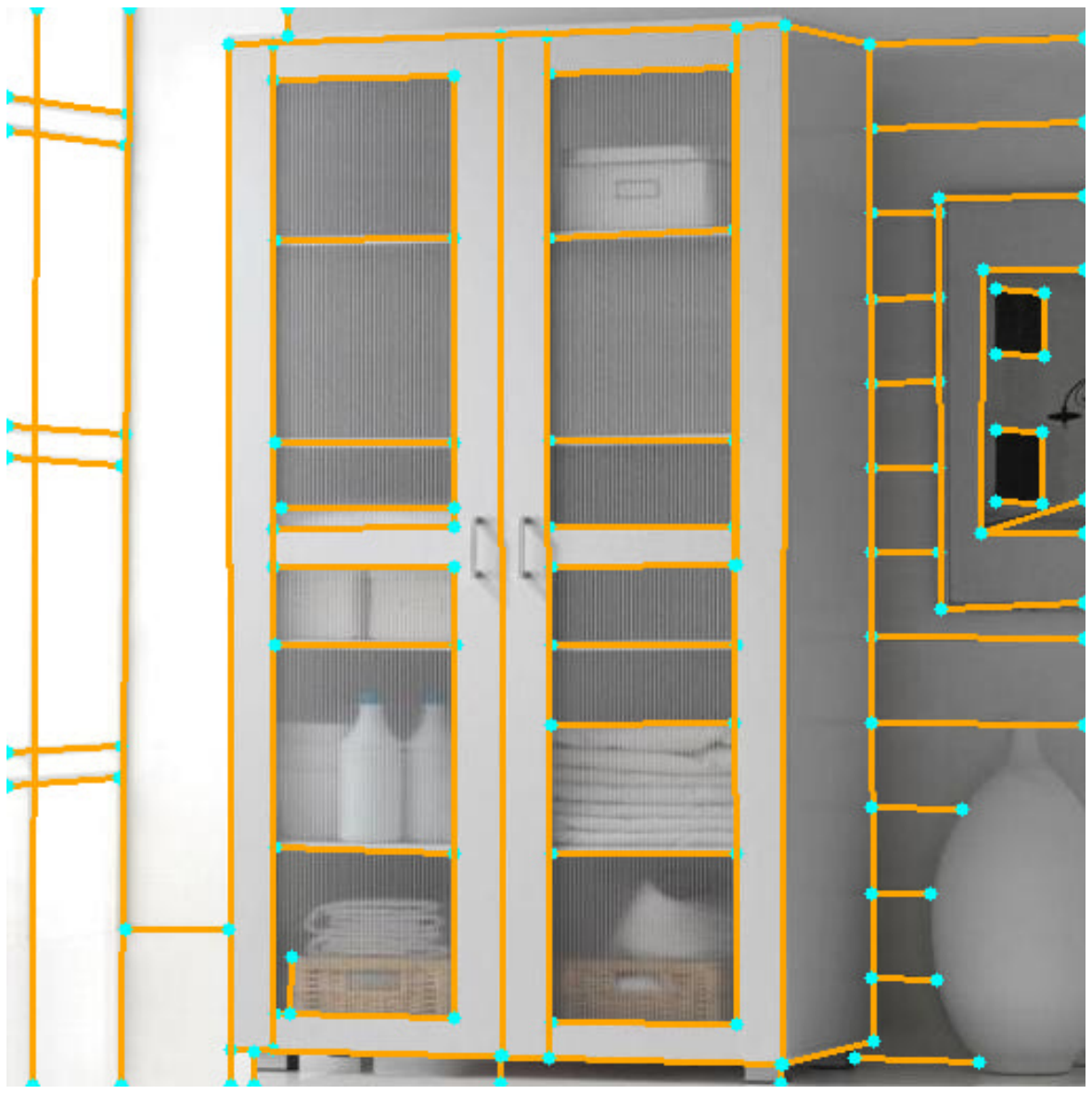

(a) HT-LCNN [15, 7]

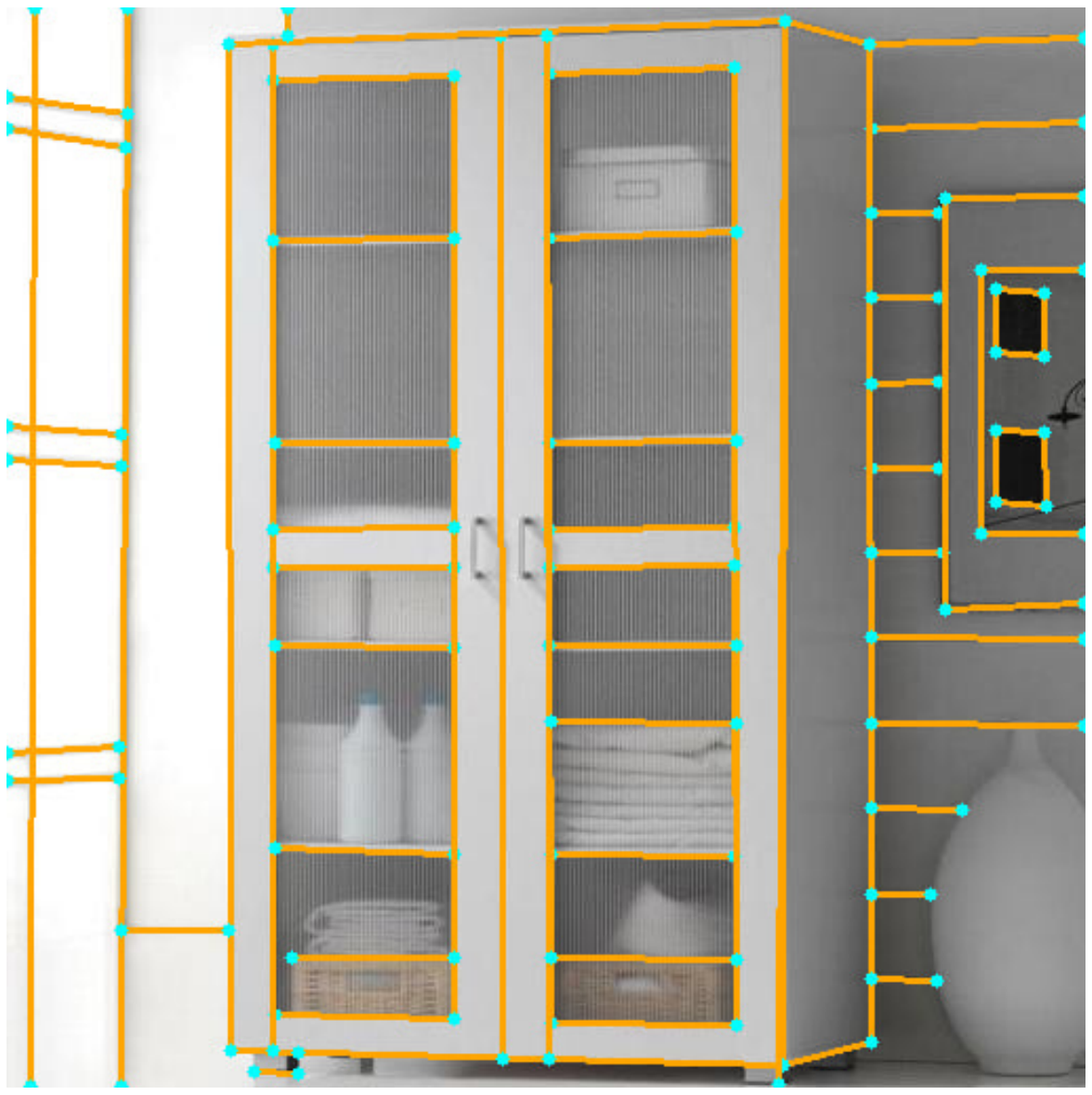

(b) HT-HAWP [12, 7]

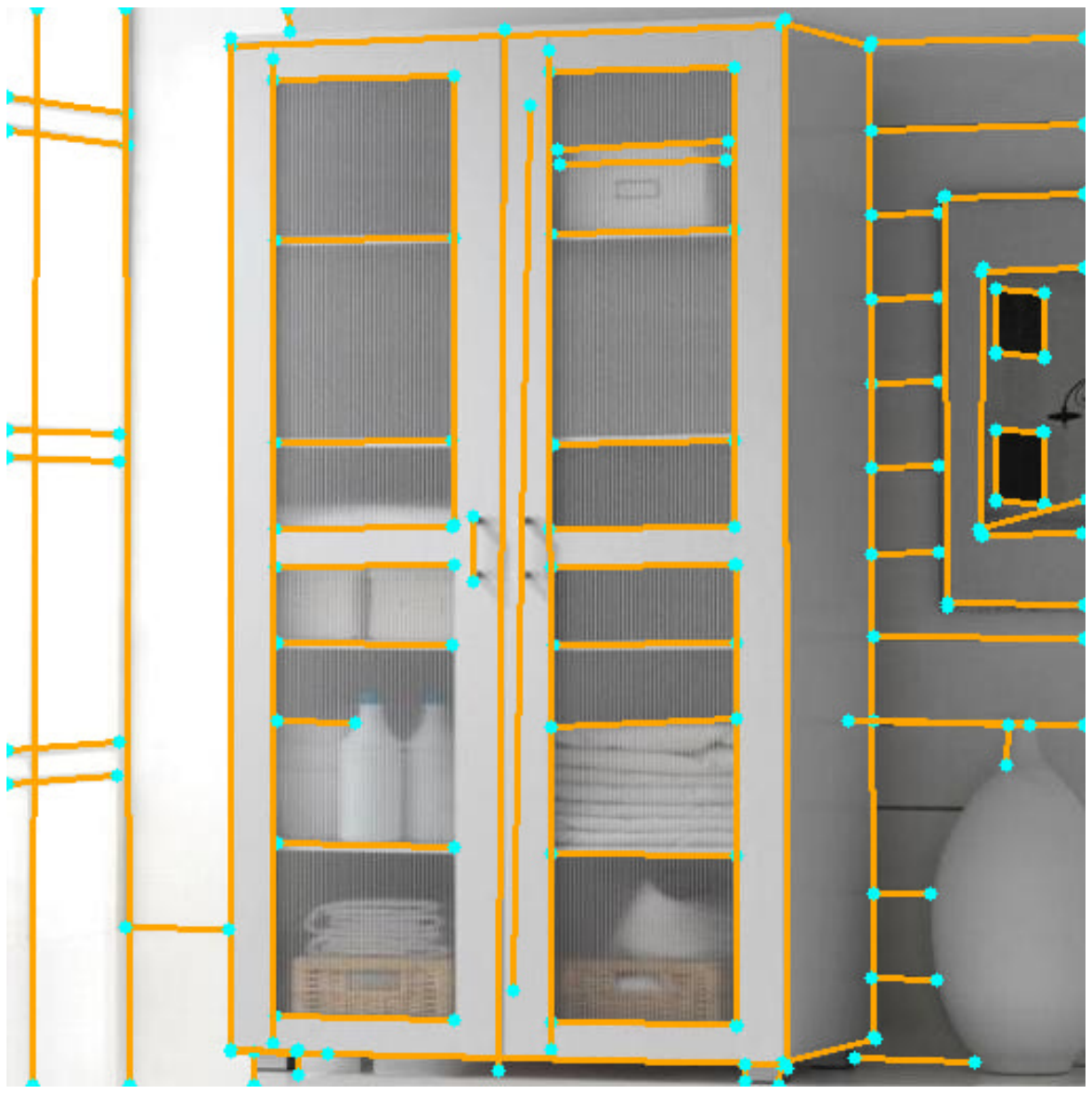

(c) F-Clip(HR) [1]

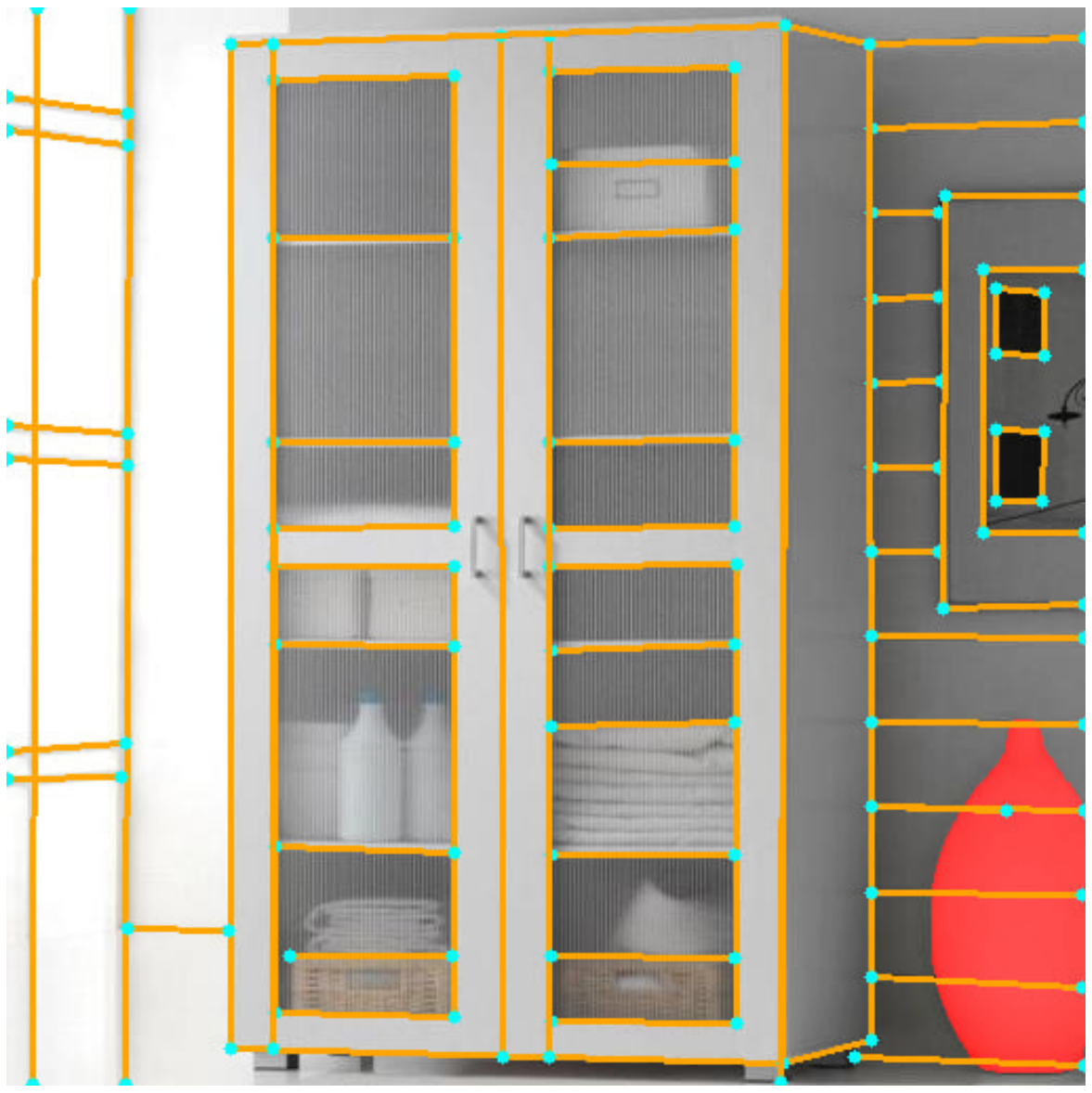

(d) Our full approach using HT-HAWP [12, 7]

Figure 21: Results on real examples with occlusion by a foreground object (white vase on the bottom right corner). (a)-(c) failed to detect wireframe vectors behind the foreground object. (d) The foreground object region from panoptic segmentation [10] is shown in transparent red color. Using this segmented region, we provided a pair of an image with hole and a mask map (not shown above) to our full model. The hidden structures are well detected as shown in the visualized image.

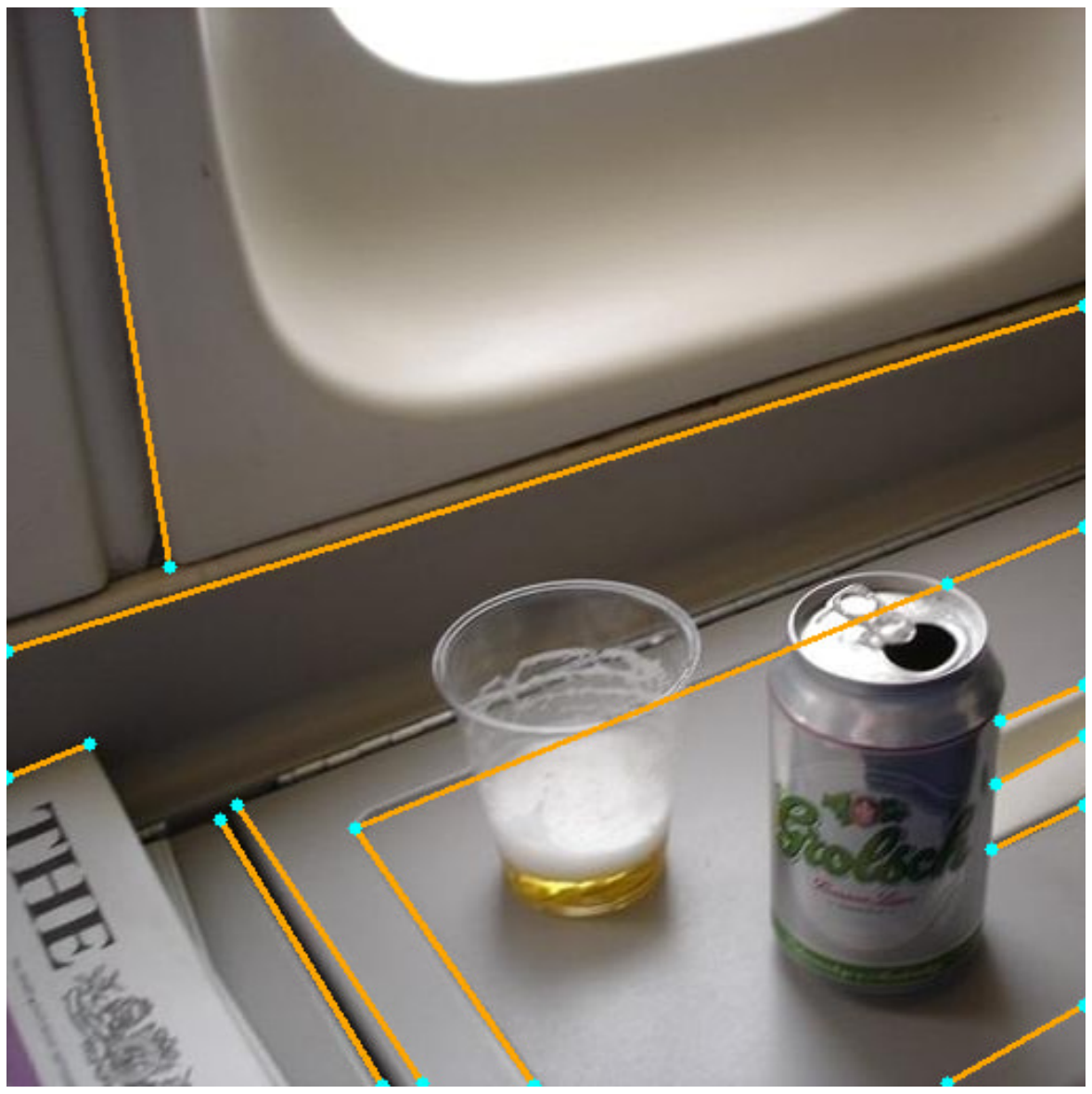

(a) HT-LCNN [15, 7]

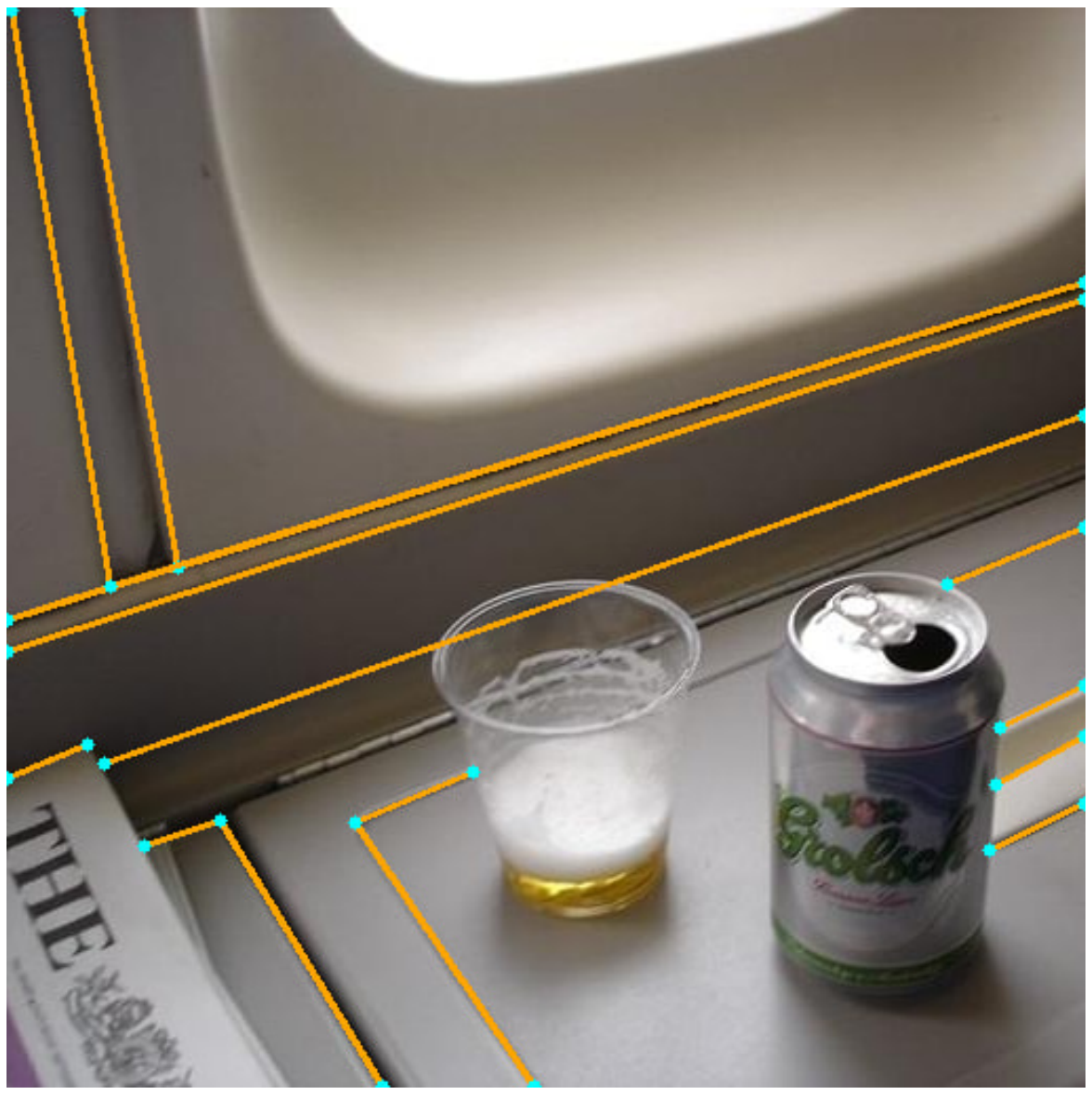

(b) HT-HAWP [12, 7]

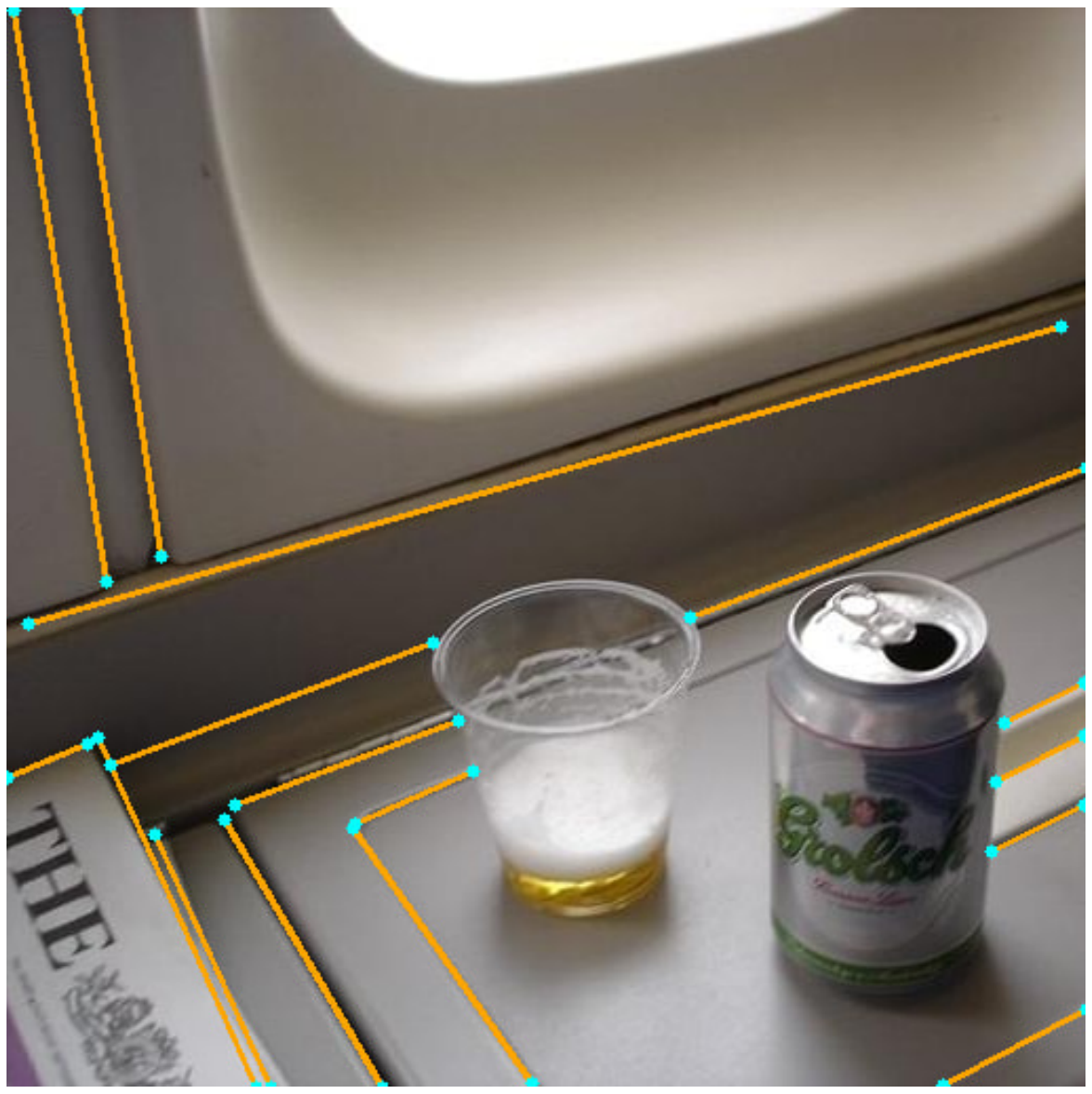

(c) F-Clip(HR) [1]

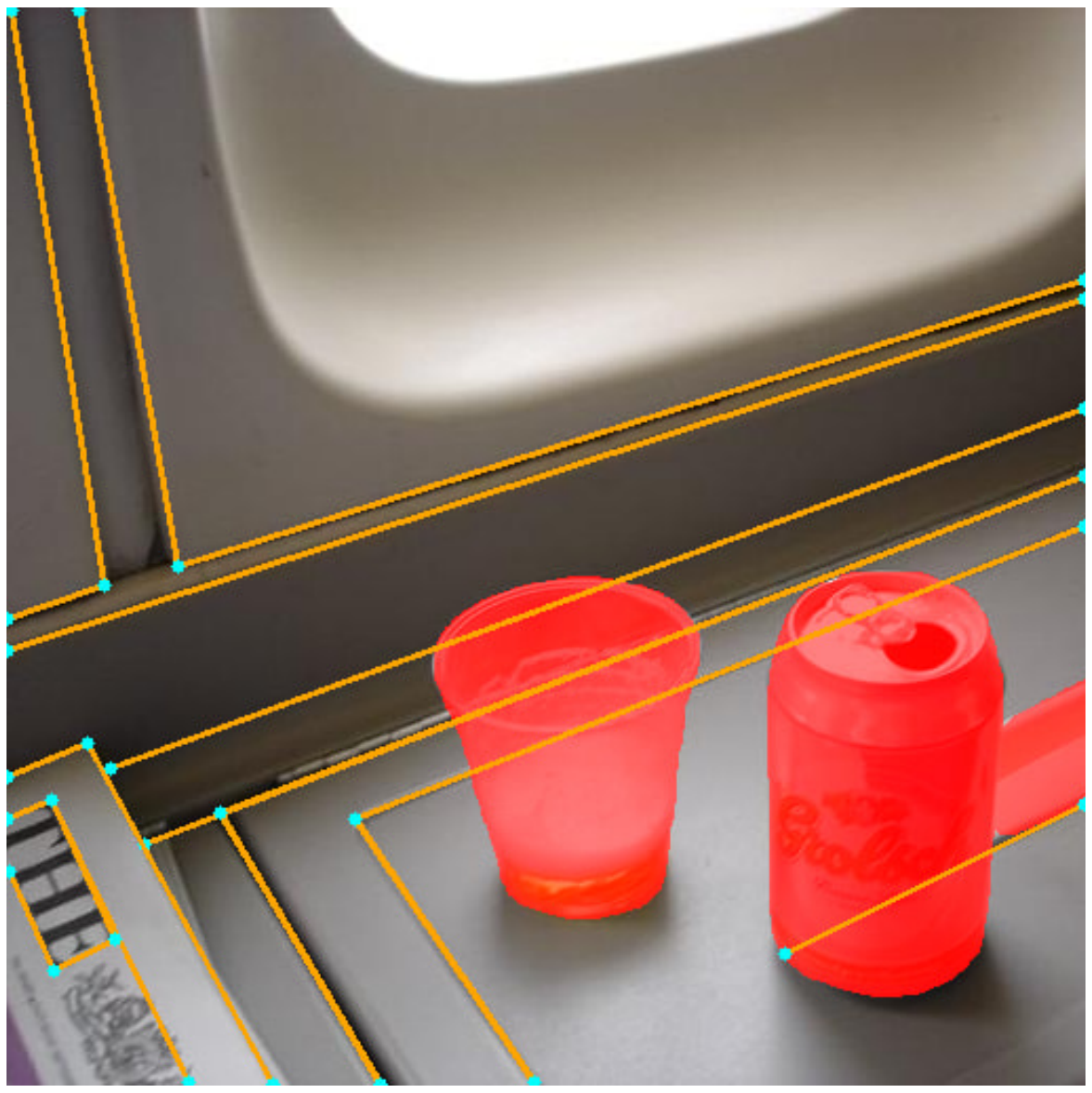

(d) Our full approach using HT-HAWP [12, 7]

Figure 22: Results on real examples with occlusion by foreground objects (cup and beer can). (a)-(c) do not well detect wireframe vectors behind the foreground objects. (d) The foreground object regions from panoptic segmentation [10] is shown in transparent red color. Using these segmented regions, we provided a pair of an image with hole and a mask map (not shown above) to our full model. The hidden structures are well detected as shown in the visualized image.

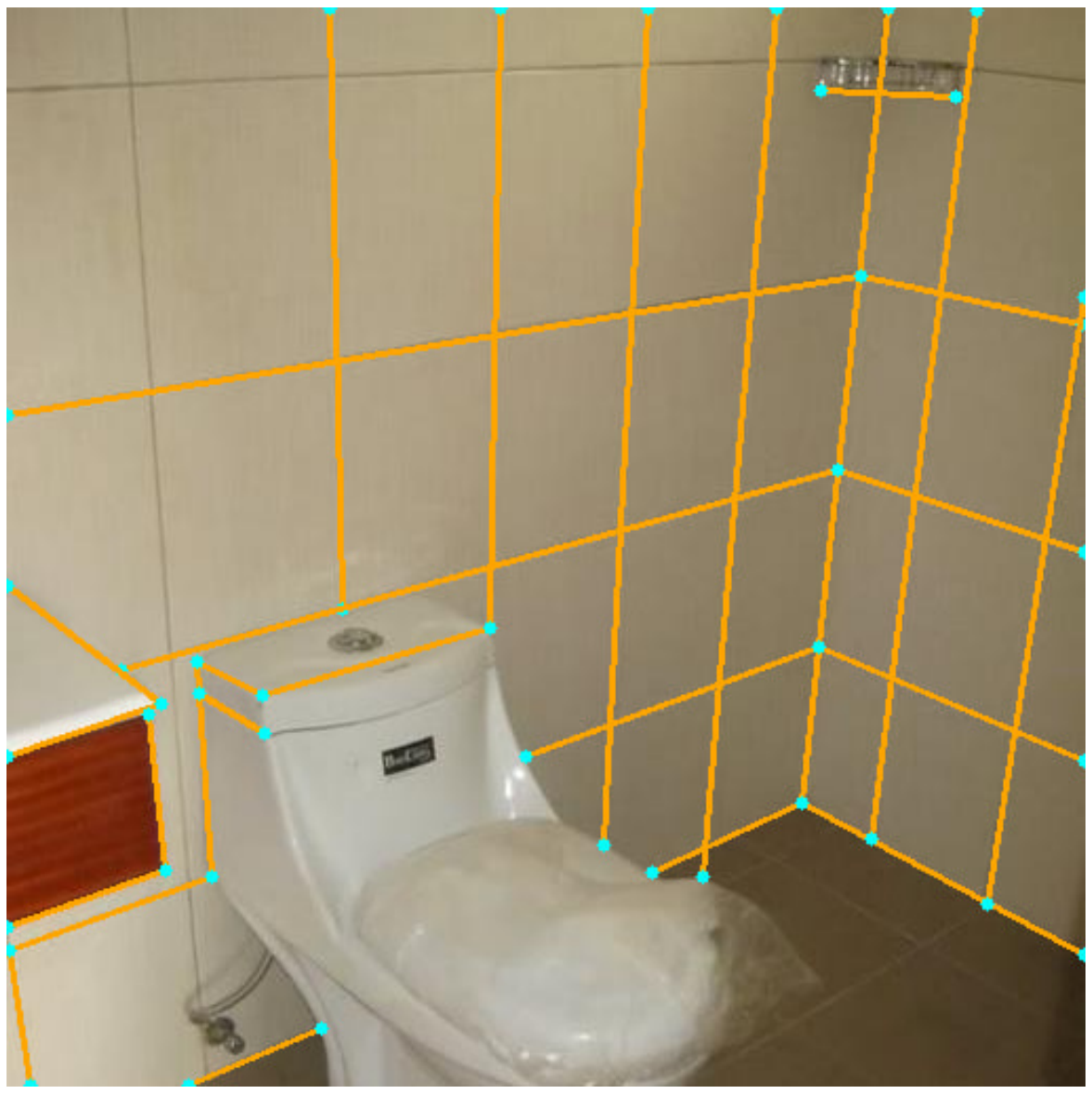

(a) HT-LCNN [15, 7]

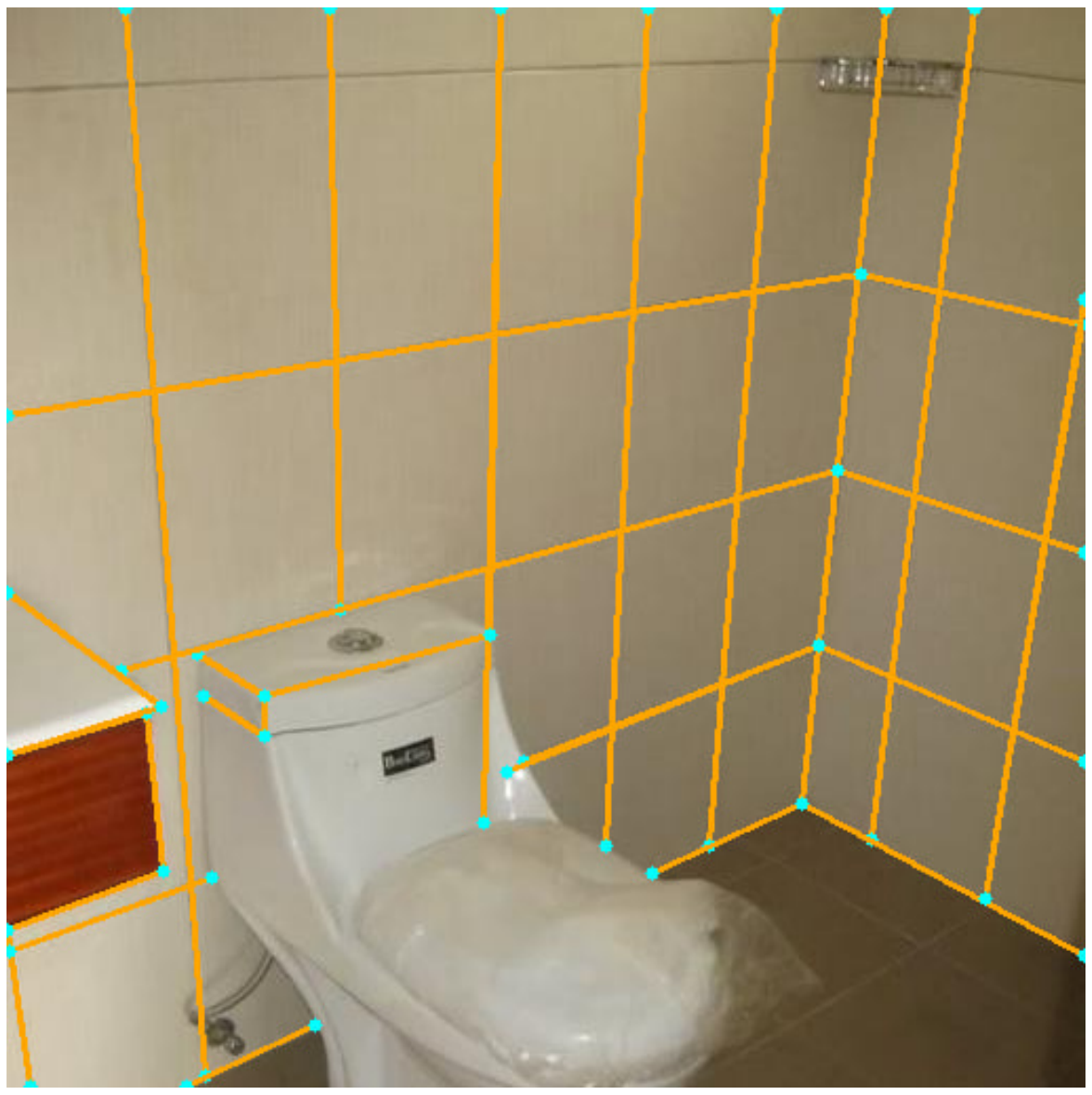

(b) HT-HAWP [12, 7]

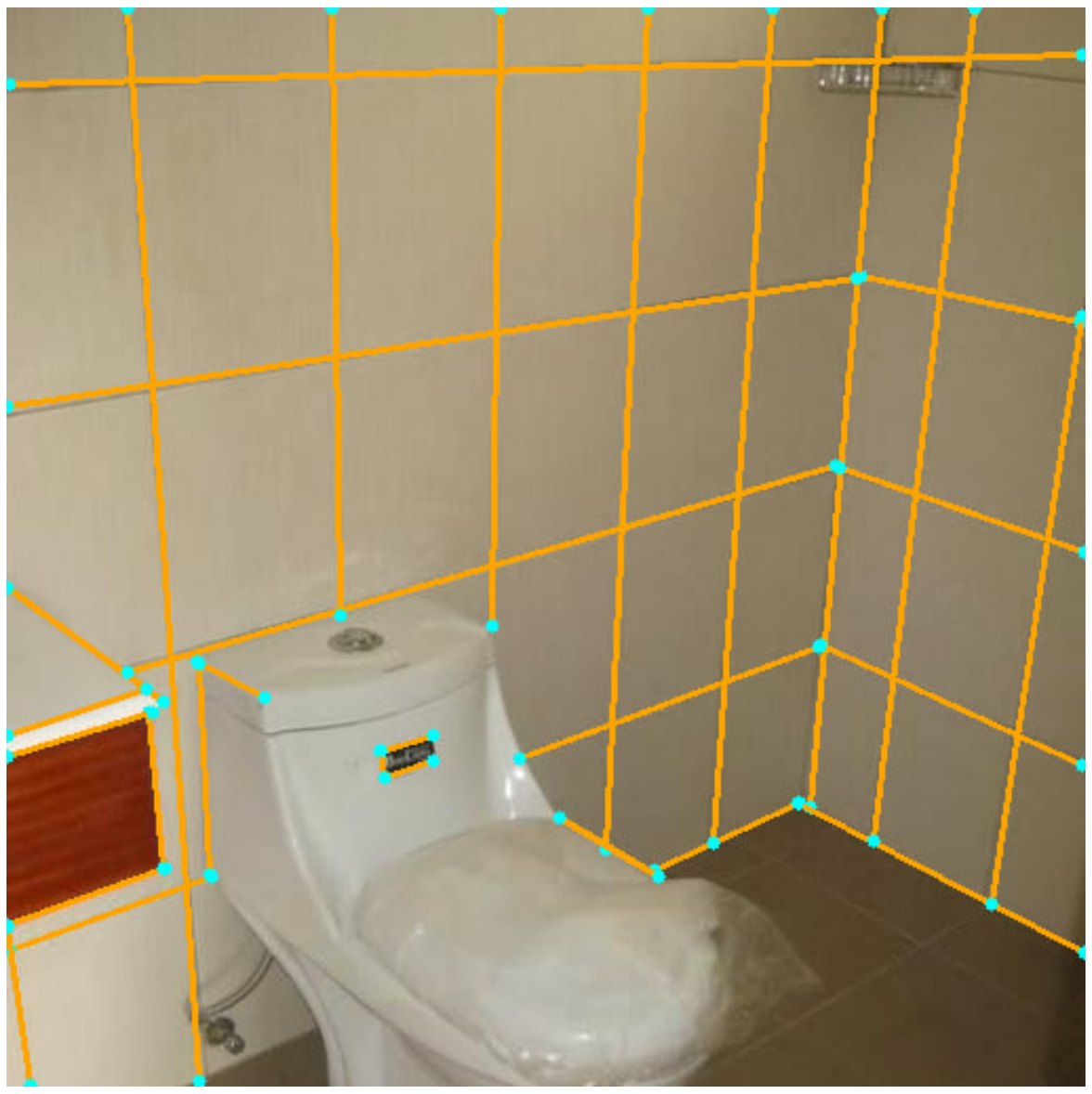

(c) F-Clip(HR) [1]

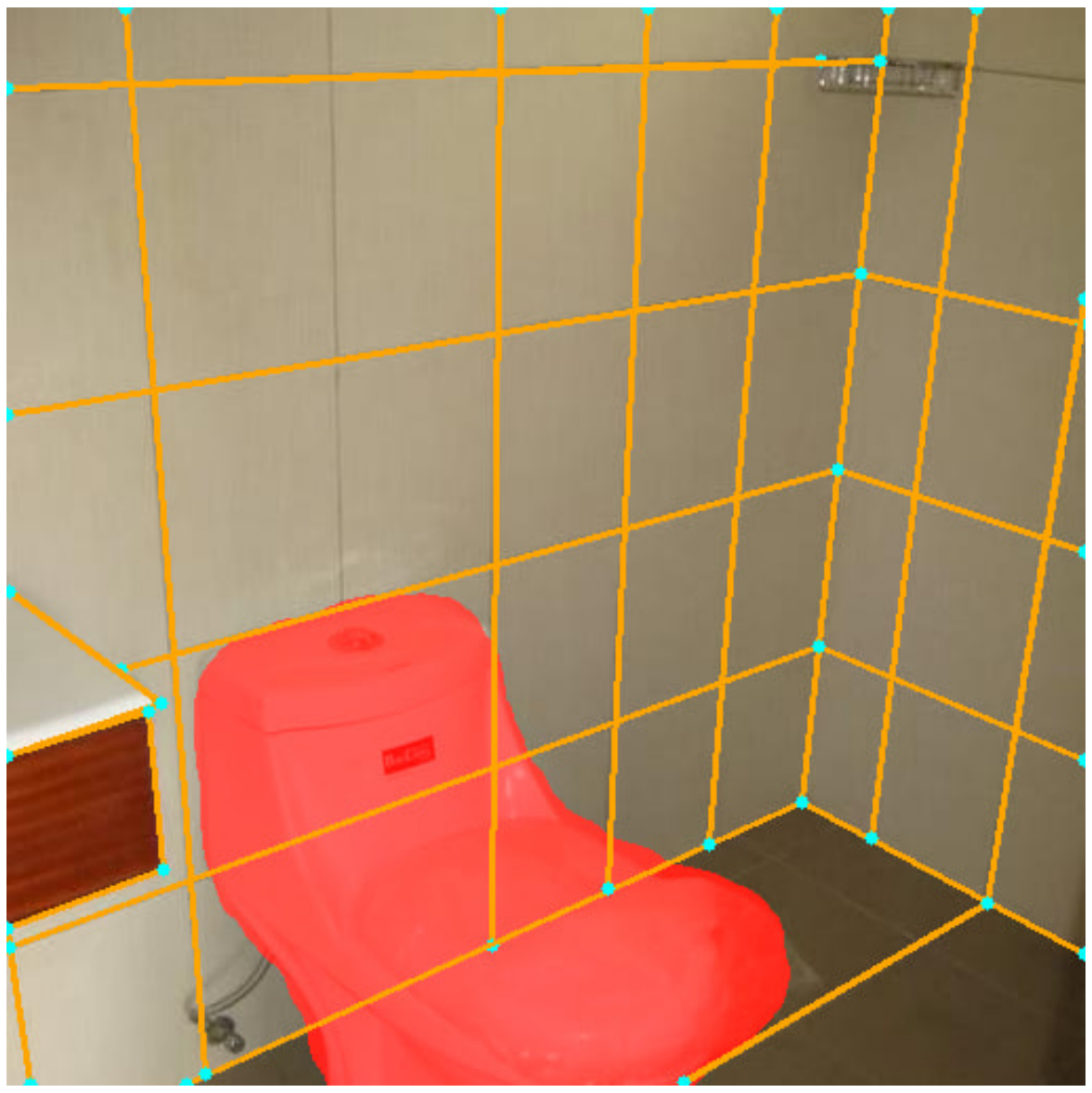

(d) Our full approach using HT-HAWP [12, 7]

Figure 23: Results on real examples with occlusion by a large foreground object (toilet). (a)-(c) failed to detect wireframe vectors behind the foreground object. (d) The foreground object region from panoptic segmentation [10] is indicated in transparent red color. Using this segmented region, we provided a pair of an image with hole and a mask map (not shown above) to our full model. The hidden structures are well detected as shown in the visualized image.

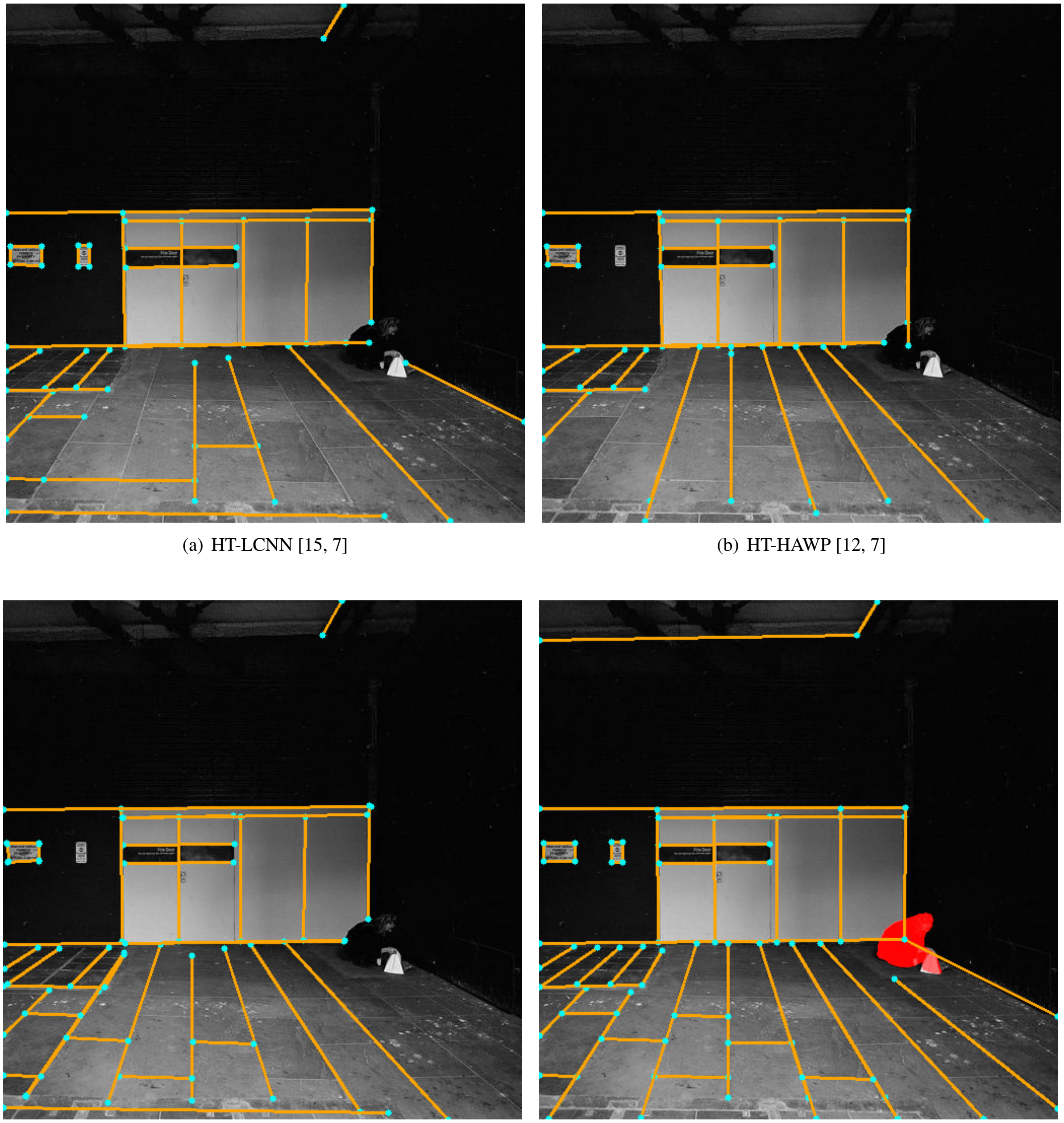

(a) HT-LCNN [15, 7]

(b) HT-HAWP [12, 7]

(c) F-Clip(HR) [1]

(d) Our full approach using HT-HAWP [12, 7]

Figure 24: Results on real examples with occlusion by a foreground object (crouching person in black clothes, quite invisible due to the black background). (a)-(c) do not well detect wireframe vectors behind the foreground object. (d) The foreground object region from panoptic segmentation [10] is indicated in transparent red color. Using this segmented region, we provided a pair of an image with hole and a mask map (not shown above) to our full model. The hidden structures are well detected as shown in the visualized image.